\documentclass{article}
\usepackage[T1]{fontenc}
\usepackage{iclr2027_conference,times}
\usepackage{amsmath,amssymb,amsthm,mathtools,mathrsfs}
\usepackage{booktabs,longtable,array}
\usepackage{graphicx}
\usepackage{float}
\usepackage{enumitem}
\usepackage{xcolor}
\usepackage{hyperref}
\usepackage{url}
\usepackage{xurl}
\definecolor{pendingcolor}{RGB}{125,58,24}

\newcommand{\E}{\mathbb{E}}

\DeclareMathOperator{\Var}{Var}

\hypersetup{
  hidelinks,
  pdftitle={Delayed Heuristic-Behavior-Target Coupling for Off-Policy Reinforcement Learning in Inventory Control},
  pdfauthor={Enhui Li, Yuan Cheng, Peiting Han},
  pdfsubject={}
}

\newtheorem{theorem}{Theorem}
\newtheorem{corollary}{Corollary}
\newlength{\priorTableWidth}
\DeclareMathOperator{\Unif}{Unif}
\DeclareMathOperator{\Bernoulli}{Bernoulli}
\DeclareMathOperator{\Geom}{Geom}

\title{Delayed Heuristic-Behavior-Target Coupling\\
for Off-Policy Reinforcement Learning\\ in Inventory Control}

\author{Enhui Li \quad Yuan Cheng \quad Peiting Han\\
Xi'an Jiaotong University}
\iclrfinalcopy

\begin{document}
\maketitle
\lhead{}
\begin{abstract}
Many reinforcement learning methods tie data collection to the task policy, while coordinating independently optimized behavior and target policies remains challenging. Policies learned by these methods can perform unreliably in long-horizon tasks under uncertainty, such as inventory control, where failures can incur extreme costs. We introduce Heuristic Behavior Policy Optimization (HBPO), an off-policy framework that routes heuristic guidance through persistent behavior learning, coupled with target-policy, updating separately with residual-based advantage estimation and emphatic weighting. Under matched environment interaction budgets on DSIS and its risk-augmented variant DSIS-RB, HBPO recovers many high-cost initial policies and improves most non-collapse comparison groups. Cold starts achieve competitive costs without final cost collapse across all tested seeds. We identify a transmission chain in the delayed heuristic–behavior–target coupling as a key driver of recovery and improvement. Generation-level diagnostics trace heuristic uptake and subsequent selective target responses, while component ablations support the contribution of this transmission mechanism to policy recovery and improvement.
\end{abstract}

\section{Introduction}
\label{sec:introduction}

Learning a policy for long-horizon planning requires choosing how to collect the experience used to improve it \citep{whitney2021deep}. In inventory control, an order changes the pipeline of future deliveries, and its consequences depend on subsequent demand \citep{boettcher2023dual}. Decisions therefore shape not only future costs but also the states encountered during learning. This makes the behavior policy that collects data an important intervention point: if the behavior policy is itself learned, what should it be optimized to do for the learning process as a whole? Concretely, can heuristic guidance applied to it and behavior policy persistently influence the target policy, which is the policy finally want to optimize, to achieve a better performance?

Existing designs give reason to doubt that this mechanism happens in most RL frameworks. Importantly, failures are observed in long-horizon planning tasks. In a studied multi-echelon inventory network, PPO converged consistently only after inventory transitions were constrained during training \citep{geevers2024inventory}, and our comparisons show high-cost runs for PPO, Discrete SAC, and adapted GPS (Section~\ref{sec:experiments}). These methods share a default: experience is collected by the task policy or a recent snapshot of it \citep{schulman2017ppo,haarnoja2018sac}. Off-policy learning permits a distinct behavior policy \citep{degris2012offpolicy}, but leaves open how that policy should improve and how its deviations should inform target updates. Behavior--target mismatch can impair approximate off-policy updates, and fitting a policy to a teacher's actions need not preserve the teacher's long-horizon performance. Our focus is therefore to construct explicit coupling updating rules for the transmission path itself: which behavioral deviations should a separately updated target reinforce or counteract, and when should planning-derived guidance enter its updates?

We introduce \textbf{Heuristic Behavior Policy Optimization (HBPO)}, an off-policy coupled framework that routes heuristic guidance through a persistent behavior policy to a separately and selectively optimized target policy. HBPO forms an explicit delayed heuristic–behavior–target transmission: heuristic guidance, constructed from the current generation target policy, first shapes the successor behavior policy then reaches two-next target updates through four combined optimizations. This transmission is based on data collected by behavior, residual-based advantage estimation supporting both policies and emphatic weighting carrying heuristic confidence forward. Under same interaction budgets on DSIS and DSIS-RB, HBPO recovers many high-cost initial policies, lowers mean costs across most initialization sources, and improves most non-collapse comparison groups. Cold starts achieve competitive costs without cost collapse across all tested seeds. We identify the transmission of delayed heuristic–behavior–target coupling as a key driver of recovery and improvement. An anonymized implementation is included in the supplementary material and will be made publicly available upon acceptance.
\section{Related Work}
\label{sec:related}

\paragraph{Behavior learning and selective transfer.}
Off-policy learning permits the collector to differ from the target \citep{degris2012offpolicy}. On-policy methods remove the difference by collecting only from the current policy (policy-gradient, natural-gradient, trust-region, and asynchronous families) \citep{williams1992simple,sutton1999policy,schulman2016gae,kakade2001natural,schulman2015trpo,schulman2017ppo,mnih2016asynchronous}, so the data source cannot outgrow the target.  Common algorithms derive their evolving behavior distributions from learned value functions or actors and exploration rules \citep{mnih2015human,vanhasselt2016deep,schaul2016prioritized,lillicrap2016continuous,fujimoto2018addressing,haarnoja2018sac}. Correction schemes persistently track the behavior (truncated importance sampling, Retrace, V-trace) reconcile it with target updates without shaping it \citep{wang2017acer,munos2016safe,espeholt2018impala,gruslys2018reactor}, so correction grows with the gap. Heuristic shaping (optimism, count-based bonuses, curiosity, learned novelty policies, variance-optimized behavior) can improve data collection, but these gains need not translate into consistent target-policy improvement \citep{auer2002finite,bellemare2016unifying,pathak2017curiosity, burda2019exploration,whitney2021deep,xu2018explore, liu2020adac,goodall2026behaviour}. HBPO couples three in one loop: a temporary heuristic controller guides a persistent behavior learner, and behavior data plus an advantage-filtered transfer term weighting signed target-advantage scores by absolute behavior--target disagreement feed target updates. ACER uses a one-sided residual to correct importance-weight truncation bias \citep{wang2017acer}; HBPO uses absolute behavior--target disagreement to weight an additional advantage-based transfer objective. Our experiments support the contribution of this heuristic--behavior--target coupling (Section~\ref{sec:experiments}).

\paragraph{Coupled updates and transmission.}
With the heuristic, behavior, and target roles fixed, the key question is how improvement travels through their coupled updates. GPS fits local controllers directly into the global policy \citep{levine2013gps}; conservative and regularized policy iteration fit an improved action distribution into the actor \citep{kakade2002cpi,montgomery2016mirror,abdolmaleki2018mpo}; TRPO, PPO, and GRPO make each improved policy the immediate data collector \citep{schulman2015trpo,schulman2017ppo,shao2024deepseekmath}; and TPO fits the policy to a softmax-improved target distribution \citep{kaddour2026tpo}. HBPO instead claims a delayed coupling: information and policy impacts enter the next generation separately, rather than the current one, making the transmission $\pi_g\to q_{H,g}\to\mu_{g+1}\to\pi_{g+2}$ explicit; Section~\ref{sec:experiments} examines how this transmission operates during recovery and improvement.

\paragraph{Emphatic state weighting.}
Emphatic temporal-difference and actor--critic methods relate behavior sampling to the state weighting of an off-policy learning objective \citep{sutton2016emphatic,imani2018emphatic,graves2023offpolicy,zhang2019generalized}, and expected emphatic traces and adaptive interest extend this framework to learned state emphasis \citep{jiang2022expected,klissarov2022interest}. HBPO adapts this perspective through separate learned masses for target-return learning and behavior transfer, with heuristic-improvement confidence supplying interest for the transfer mass.

\paragraph{Inventory control as a testbed.}
Deep RL has proven effective across lost-sales, dual-sourcing, and multi-echelon inventory problems \citep{gijsbrechts2022inventory}, though training conditions materially affect learning \citep{geevers2024inventory}.  \citet{boettcher2023dual} optimize dual-sourcing policies by embedding inventory dynamics in recurrent neural networks. We adapt their problem formulation for our testbed. We choose inventory control as a testbed because poor decisions carry severe downside risk, and the systems admit a compact Markov formulation whose known or learnable dynamics support planning-derived guidance. Section~\ref{sec:problem} defines our inventory control problem.

\section{Problem Settings}
\label{sec:problem}

We adapt the single-item dual-sourcing model of \citet{boettcher2023dual} into DSIS, with regular deliveries after four periods and costlier, immediate expedited deliveries. DSIS has IID demand and uninterrupted supply; our extension, DSIS-RB, adds persistent demand and supply disruptions. 

We formulate both benchmarks as underlying Markov decision
processes $(\mathcal X,\mathcal A,\mathcal P,r,\gamma)$. At period $t$, DSIS observations $s_t=(I_t,\mathbf p_t,B_t)$ contain net inventory, a four-stage order pipeline, and current supply availability. DSIS-RB adds previous demand $d_{t-1}$, increasing the dimension from six to seven. Actions $a_t=(q_t^r,q_t^e)\in\{0,\ldots,8\}^2$ specify regular and expedited orders. DSIS-RB training pools trajectories across hidden disruption-risk levels \(\rho\), fixed within each trajectory and withheld from all learners and simulation workers. All learned networks in HBPO use only the current observation $s_t$, with fixed preprocessing. The full environment state is \(x_t=(s_t,\rho)\) for DSIS-RB and \(x_t=s_t\) for DSIS. Risk variables and their averaging are omitted for DSIS throughout. The environment kernel $\mathcal P$ governs order acceptance, arrivals, demand, supply availability, and inventory carryover through $x_{t+1}\sim\mathcal P(\cdot\mid x_t,a_t)$ (Appendix~\ref{app:protocol}). Period cost $c_t$ comprises expediting, holding, and backlogging. We maximize expected discounted return using $r_t=-c_t/495$ and $\gamma=0.995$, where $495$ is the per-unit, per-period backlog penalty. All methods share information access and real-interaction budgets. DSIS dynamics are public; HBPO and GPS construct DSIS-RB models from shared public simulation specification $\widehat{\mathcal P}_g$ and their observed data, without access to true stochastic parameters (Appendix~\ref{app:protocol}). Delayed deliveries, accumulated backlogs, and severe shortage penalties make these tasks a testbed for long-horizon decision-making under uncertainty, where failures can incur extreme costs.
\section{Heuristic Behavior Policy Optimization Methods}
\label{sec:methods}

\begin{figure}[H]
    \centering
    \includegraphics[width=\linewidth]{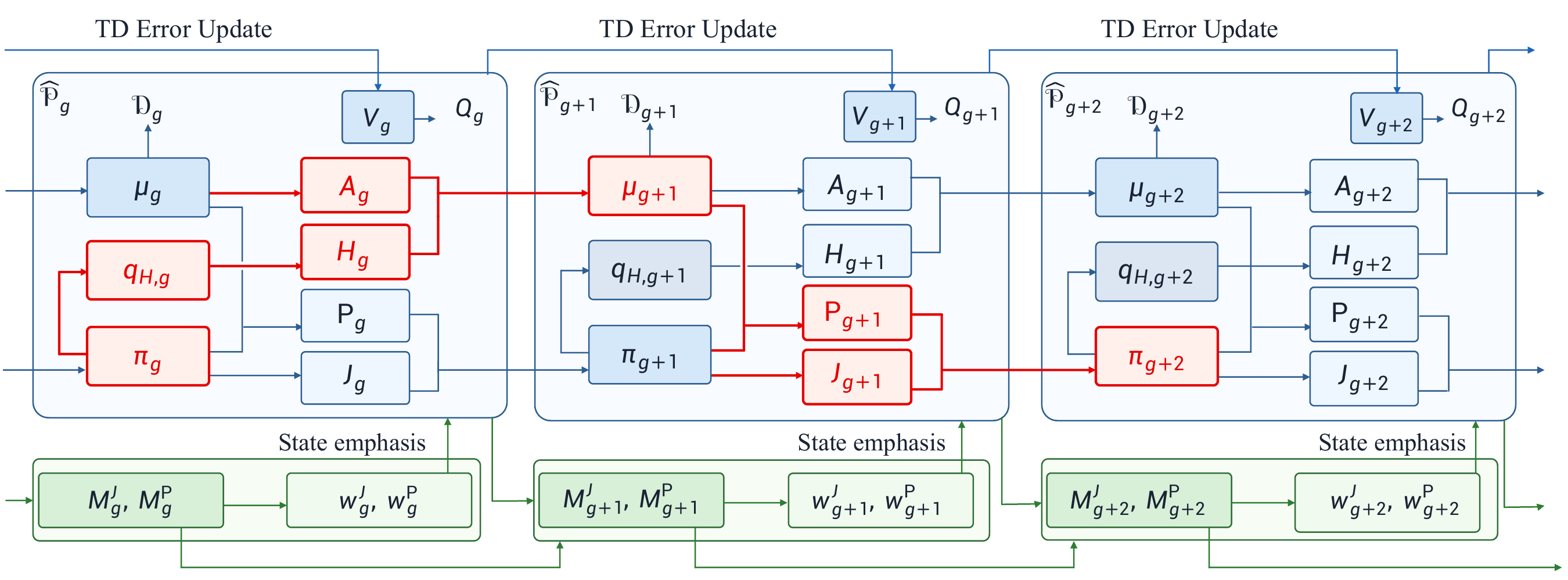}
    \caption{HBPO Learning Framework. Red highlights the delayed heuristic transmission.}
    \label{fig:hbpo-framework}
\end{figure}

HBPO maintains two separately parameterized, persistent policies, \(\mu_g=\mu_{\theta_g^\mu}\) and \(\pi_g=\pi_{\theta_g^\pi}\). The behavior policy \(\mu_g\) collects interaction data \(\mathcal D_g\), while the target policy \(\pi_g\) is evaluated for task performance and defines the continuation policy for model evaluation. Under the simulation specification \(\widehat{\mathcal P}_g\), frozen within generation \(g\), long-horizon comparisons of alternative first actions followed by \(\pi_g\) construct a temporary heuristic policy \(q_{H,g}\). A shared advantage face scores policy changes and behavior--target discrepancies using discounted target-policy advantages, defining the \(A,J,P\) objectives. Together with heuristic guidance \(H\), these yield four gradients: behavior combines target-advantage learning \(g_A\) with heuristic guidance \(g_H\), while target combines its own advantage learning \(g_J\) with selective transfer from behavior \(g_P\).

The frozen reference \(v_g(s)=\frac15\sum_{j=1}^{5}V_{g,j}(s)\), averaging five separately parameterized scalar critics, supports heuristic evaluation and residual-completed estimation of the discounted model action value \(Q_g^{\pi_g}(s,a)\); successor critics are fitted by minimizing squared TD error. Positive masses \(M_g^\chi\) provide mean-one emphatic weights \(w_g^\chi(s)\), \(\chi\in\{J,P\}\), for the state-weighted parts of learning; their construction and the critic updates are specified in Section~\ref{sec:learning-support}.

\subsection{Heuristic Objectives Guidance}
\label{sec:heuristic-guidance}

To construct a temporary heuristic policy from \(\pi_g\), anchor states are sampled from \(\mathcal D_g\), with each anchor state \(s\) defining local time \(k=0\). We compare the long-horizon consequences of candidate first actions \(a\) over \(T=1000\) periods, with all subsequent actions drawn from the frozen parent target \(\pi_g\). Let \(S_k\), \(\mathsf A_k\), and \(R_k\) denote the observation, action, and normalized reward at local decision stage \(k\). Writing \(\mathbb E_H\) for expectation under \(\widehat{\mathcal P}_g\), including the synthetic-risk draw for each DSIS-RB tape, we define the expected return \(G_{T,g}(s,a)\) and the centered estimated utility \(u_g(s,a)\) as
\[
G_{T,g}(s,a)
=\mathbb E_H\!\left[
\sum_{k=0}^{T-1}R_k
\,\middle|\,
\substack{S_0=s,\ \mathsf A_0=a\\
\mathsf A_k\sim\pi_g(\cdot\mid S_k)\;\;(1\le k<T)}
\right],
u_g(s,a)
=\widehat G_{T,g}(s,a)
-\sum_{a'\in\mathcal A}
\pi_g(a'\mid s)\widehat G_{T,g}(s,a').
\]
Here, \(\widehat G_{T,g}\) is a Monte Carlo estimate of \(G_{T,g}\), constructed using the frozen critic \(v_g\) as a control variate and \(32\) simulation tapes shared across the candidate first-action branches at each distinct query observation. All queries are performed within the model, requiring no additional environment interactions. Appendices~\ref{app:critic-model-construction} and~\ref{app:heuristic-utility} detail the simulation specification and the utility estimator.

For a candidate action distribution \(q\) at state \(s\),
\(
\langle q,u_g(s,\cdot)\rangle
=\sum_a\bigl(q_a-\pi_g(a\mid s)\bigr)\widehat G_{T,g}(s,a)
\)
measures the estimated return gain from replacing the \(\pi_g(\cdot\mid s)\) with \(q\) and following \(\pi_g\) thereafter. Maximizing this gain alone would concentrate probability on utility-maximizing actions. We therefore construct a temporary stochastic policy for heuristic guidance through Hellinger regularization
\begin{equation}
q_{H,g}(\cdot\mid s)
=\arg\max_{q\in\Delta(\mathcal A)}
\left\{
\langle q,u_g(s,\cdot)\rangle
-2\tau\sum_a
\left(\sqrt{q_a}-\sqrt{\pi_g(a\mid s)}\right)^2
\right\}.
\label{eq:h-controller}
\end{equation}
The utility term favors higher estimated returns, while the symmetric, bounded squared Hellinger penalty moderates departures from the target prior to help stabilize heuristic guidance. The coefficient \(\tau>0\) controls this trade-off and remains fixed throughout each run after calibration under the shared hyperparameter-selection protocol HPO. With a strictly positive target prior, scalar root finding yields the temporary guidance policy at each queried state. The resulting policy has no trainable actor parameters. Appendix~\ref{app:h-derivation} details the calibration, controller solution, and derivation; Appendix~\ref{app:protocol} reports the hyperparameter-selection protocol.

Denote \(\nu_g^J\) as the empirical distribution of behavior states weighted by \(w_g^J\),introduced in Section~\ref{sec:learning-support}. Starting from anchors drawn from \(\nu_g^J\), short model rollouts under \(q_{H,g}\) generate the empirical training distribution \(\xi_g\) by retaining states at stages \(0,\ldots,8\). Each retained state receives a controller query with a full \(T=1000\) horizon, so the short rollouts select training locations without shortening the utility horizon.  Using \(q_{H,g}(\cdot\mid s)\) as the fitting targets at each retained state, define the behavior-guidance gradient by
\begin{equation}
g_{H,g}= -2\left. \nabla_{\theta^\mu}
\mathbb E_{s\sim\xi_g}
\left[ \left(
\arccos\!\left[ \sum_a
\sqrt{ \mu_{\theta^\mu}(a\mid s)\,
q_{H,g}(a\mid s)}
\right]\right)^2\right]
\right|_{\theta^\mu=\theta_g^\mu},
\label{eq:h-guidance-gradient}
\end{equation}
where \(q_{H,g}\) and \(\xi_g\) are held fixed during differentiation. Detailed in Appendix~\ref{app:h-behavior-gradient}.

\subsection{Delayed Heuristic--Behavior--Target Coupling}
\label{sec:ahjp-updates}
\label{sec:delayed-properties}
Here we explicitly construct the delayed heuristic--behavior--target coupling and certain the policies updating rules. For the \(n\) behavior transitions \((s_i,a_i,s_i')\) in \(\mathcal D_g\), we now specify the advantage face used by the coupled policy updates. Following the centering used for \(u_g\), we evaluate each first action by its discounted return advantage:
\(
\mathbf a_i(a)
=
Q(s_i,a;\pi_g)
-
\sum_b\pi_g(b\mid s_i)\,Q(s_i,b;\pi_g),
\) where \(Q(s,a;\pi_g)\) denotes the discounted model action value, helping advantage estimation, introduced in Equation~\eqref{eq:critic-residual-completion}. The vector \(\widehat{\mathbf a}_i\) records these gains at sampled state \(s_i\) compared to  \(\pi_g\), and for policy change \(\ell\), define advantage face
\(\Lambda_t[\ell]
=\frac1n\sum_{i=1}^n c_i^t\langle\ell_i,\widehat{\mathbf a}_i\rangle,\ (c_i^A,c_i^J,c_i^P)=(1,w_i^J,w_i^P).
\)
The score rewards adding weight to actions with positive estimated advantage and removing weight from actions with negative estimated advantage. Here \(\chi \in\{A,J,P\}\), \(w_i^\chi=w_g^\chi(s_i)\), and \(c_i^\chi\) determines how strongly each sampled state contributes. Appendix~\ref{app:residual-derivation} details the estimator.

Suppressing generation indices, we hold the parent policies \(\mu,\pi\) fixed and use \(m,p\) to denote possible updated behavior and target policies. Both actors receive a direct advantage-driven update:

\begingroup\setlength{\parskip}{0pt}
\setlength{\abovedisplayskip}{1pt}
\setlength{\belowdisplayskip}{1pt}
\setlength{\abovedisplayshortskip}{1pt}
\setlength{\belowdisplayshortskip}{1pt}
\begin{minipage}{0.48\linewidth}
\begin{equation}
A(m)=\tau^{-1}\Lambda_A[m-\mu].
\label{eq:ahjp-A}
\end{equation}
\end{minipage}\hfill
\begin{minipage}{0.48\linewidth}
\begin{equation}
g_A=\tau^{-1}\Lambda_A
\bigl[\mu\nabla_{\theta^\mu}\log\mu\bigr].
\label{eq:ahjp-gA}
\end{equation}
\end{minipage}

\begin{minipage}{0.48\linewidth}
\begin{equation}
J(p)=\Lambda_J[p-\pi].
\label{eq:ahjp-J}
\end{equation}
\end{minipage}\hfill
\begin{minipage}{0.48\linewidth}
\begin{equation}
g_J=\tau^{-1}\Lambda_J
\bigl[\pi\nabla_{\theta^\pi}\log\pi\bigr].
\label{eq:ahjp-gJ}
\end{equation}
\end{minipage}
\endgroup
Gradients are evaluated at the frozen parent parameters, with $g_J=\nabla_{\theta^\pi}(J/\tau)$ matching the advantage scale used by $A$ and $P$. $\tau$ is the heuristic regularization coefficient in Equation~\eqref{eq:h-controller}. $A(m)$ directs behavior toward actions favored by target-advantage evidence, while $J(p)$ updates target under its own state emphasis. Together with $g_H$ from Equation~\eqref{eq:h-guidance-gradient}, behavior combines advantage-driven learning $g_A$ with attraction toward $q_H$.

Heuristic-guided behavior should also inform target action choices through an explicit policy-level signal. We define 
\begin{equation}
P(p)=\tau^{-1}\Lambda_P[|\mu-\pi|(p/\pi-1)],
\qquad
g_P=\frac1{n\tau}\sum_{i,a}w_i^P|\mu_i(a)-\pi_i(a)|\widehat a_{ia}\nabla_{\theta^\pi}\log\pi_i(a),
\label{eq:ahjp-p-gradient}
\end{equation}
where $\mu_i(a)=\mu(a\mid s_i)$ and similarly for $\pi$. This advantage-filtered transfer weights disagreements by their magnitude and uses advantage to set their direction: supported behavioral deviations are retained, and opposed deviations are reversed in the scores. $P$ can emphasize actions favored by behavior but rarely selected by target; it can increase disagreement and vanishes when $\mu=\pi$. Behavior's action distribution thus guides target optimization beyond its role as a data source. Theorem~\ref{thm:coupling-selective-transmission} shows that the filtered transfer signal admits an auxiliary distribution with nonnegative gain in the frozen advantage face.

Each actor combines its two gradients in its own Fisher geometry, which relates parameter changes to local changes in policy distributions \citep{kakade2001natural}. We use $\operatorname{Fuse}_F(g_1,g_2)=F^{-1}(\alpha_1g_1+\alpha_2g_2)$, with coefficients determined by the natural norms $n_i=\sqrt{g_i^\top F^{-1}g_i}$ to balance their contributions. The updates are

\begin{minipage}{0.48\textwidth}
\begin{equation}
\theta_{g+1}^{\mu}
= \theta_g^{\mu}
+ t_{\mu}\operatorname{Fuse}_{F_{\mu}}(g_A,g_H).
\label{eq:ahjp-update-mu}
\end{equation}
\end{minipage}\hfill
\begin{minipage}{0.48\textwidth}
\begin{equation}
\theta_{g+1}^{\pi}
= \theta_g^{\pi}
+ t_{\pi}\operatorname{Fuse}_{F_{\pi}}(g_J,g_P).
\label{eq:ahjp-update-pi}
\end{equation}
\end{minipage}
    Shared Fisher geometry makes the two gradients comparable in terms of local policy change, while norm-based fusion balances their contributions. Denote \(\nu^P\) for the empirical behavior-state distribution weighted by \(w^P\), introduced in Section~\ref{sec:learning-support}, step fractions \(t_\mu,t_\pi\in[0,1]\) then constrain endpoint movement on \((\mathcal D,\xi)\) and \((\nu^J,\nu^P)\), respectively, keeping each actor’s update within both channel budgets.  Theorem~\ref{thm:coupling-fusion} establishes bounded ascent of the frozen combined objective for the ideal fused direction and characterizes when both channels are first-order nondecreasing. Appendix~\ref{app:actor-derivation} specifies the fusion coefficients, Fisher operators, and endpoint checks.

\subsection{Critic, State Emphasis and Initialization}
\label{sec:learning-support}
\label{sec:advantage-correction}

The value estimates used above are constructed from the frozen critic reference \(v_g\) under \(\widehat{\mathcal P}_g\). Let $\widehat{\mathcal P}_{g,z}$ denote the model kernel at fixed risk $z$. For a model transition with reward $R_t$ and successor observation $S_{t+1}$, define the one-step backup and target-policy Bellman residual by
\begin{equation}
\begin{aligned}
L_{g,z}v_g(s,a)
&=\mathbb E_{\widehat{\mathcal P}_{g,z}}
  \!\left[R_t+\gamma v_g(S_{t+1})
  \,\middle|\,S_t=s,\mathsf A_t=a\right],\\
e_{g,z}(s)
&=\sum_{a'\in\mathcal A}\pi_g(a'\mid s)
  L_{g,z}v_g(s,a')-v_g(s).
\end{aligned}
\label{eq:critic-backup-residual}
\end{equation}
Appendix~\ref{app:critic-backups} gives the expanded backups. Theorem~\ref{thm:critic-residual-completion} proves that, under its stated conditions, the expected discounted residual sum exactly equals the action-value error of the one-step backup under the frozen model. Adding this correction therefore recovers the model's exact discounted action value without requiring the reference critic to have converged:
\begin{equation}
Q_g^{\pi_g}(s,a)
=\mathbb E_{Z\sim m}\!\left[L_{g,Z}v_g(s,a)\right]+\mathbb E_{Z\sim m}
  \mathbb E_{\widehat{\mathcal P}_{g,Z},\pi_g}
  \!\left[\sum_{k=1}^{\infty}\gamma^k e_{g,Z}(S_k)
  \,\middle|\,S_0=s,\mathsf A_0=a\right].
\label{eq:critic-residual-completion}
\end{equation}
where each rollout follows $\pi_g$ after the first action. To update the critics, we simulate continuations starting from observations sampled from the current batch $\mathcal D_g$. The resulting training distribution $\widetilde\zeta_g$ describes the simulated continuation states and their retained risks (Appendix~\ref{app:critic-training-bank}). Each successor critic starts from its parent and is trained to minimize an expected squared TD error of the form $\E [(V_\theta-\mathcal T V_\theta)^2]$, where the expectation is over training samples generated using
$\widetilde\zeta_g$, and $\mathcal T$ denotes the model-based Bellman backup used for critic fitting under $\pi_g$. After a finite number of optimization steps, the updated critics are averaged to form $v_{g+1}$. Appendices~\ref{app:critic-training-bank}--\ref{app:critic-successor-update} specify the sampling and update procedures.

Behavior and target policies can visit states with different frequencies, so behavior-data frequencies alone may not reflect state importance for target learning. HBPO learns two positive state $M^J$ and $M^P$ (raw masses) from $\mathcal D_g$. $M^J$ carries state importance through target-policy continuation from behavior states. For selective transfer from behavior, $M^P$ combines confidence in heuristic improvement with existing state importance propagated along behavior-policy transitions, so state emphasis in later target updates reflects both where the heuristic appears helpful and how the behavior policy moves between states. On behavior data \(\mathcal D_g\) collected by \(\mu_g\), full-data normalization and shrinkage of the parent masses give positive, mean-one weights:
\begin{equation}
w_g^\chi(s)
=1+\frac{M_g^\chi(s)-\mathbb E_{D_g}[M_g^\chi]}
{\sqrt{\mathbb E_{D_g}[(M_g^\chi)^2]}},
\qquad \chi\in\{J,P\}.
\label{eq:emphatic-weights}
\end{equation}
where $\mathbb E_{D_g}$ averages over all current-state occurrences in the data. These weights induce the empirical state distributions $\nu_g^J$ and $\nu_g^P$ used above for target-return learning and selective transfer from behavior respectively. Appendices~\ref{app:emphatic-mass-learning}--\ref{app:emphatic-consumer-weights} specify the mass fitting and weight construction. 

To obtain the confidence used by $M^P$, paired simulations share random inputs to compare \(T\)-step undiscounted returns: \(q_{H,g}\) for nine stages then \(\pi_g\), versus \(\pi_g\) throughout. The paired simulations return differences yield confidence scores for heuristic improvement. These scores are interpolated using fixed state features to the successor states in \(\mathcal D_g\) and centered around one to form the interest signal for fitting \(M^P_{g+1}\). Appendix~\ref{app:emphatic-confidence} specifies the confidence and interest construction.

The recursion starts from cold initialization or a warm handoff. Both produce separately stored policies with $\mu_0=\pi_0$, five prefitted scalar critics, and neutral emphatic masses $M_0^J=M_0^P\equiv1$, yielding $w_0^J=w_0^P\equiv1$ (Appendix~\ref{app:emphatic-initialization}). Cold initialization uses a near-uniform policy and fits critics on simulated states. In DSIS-RB, the model is fitted to the same first-generation real-data batch, included in the common 2M provider-transition budget. Warm handoff copies or distills a source policy and fits fresh scalar critics using centered action information from temporary Q networks trained to evaluate $\pi_0$. Both routes finish with constant-offset calibration (Appendix~\ref{app:critic-initialization}).
\section{Experiments}
\label{sec:experiments}

\subsection{Experimental Setup}
\label{sec:protocol}

We compare HBPO with nine native baselines on DSIS and DSIS-RB: PPO, A2C, Discrete SAC, ACE, PPO-81, TRPO, GRPO, TPO, and GPS. Warm comparisons branch from each baseline's 0.5M-interaction checkpoint into native continuation and HBPO, each using another 1.5M real interactions to assess recovery and improvement across policy sources and initializations. Cold uses 2M real interactions without native pretraining to assess learning without a source policy. Baselines use individually selected configurations; HBPO shares one configuration across all ten initializations per environment, selected on three tuning seeds to balance relative cost reductions from high-cost and non-collapsed starts among seven configuration sets. Final policies from five separate training seeds are evaluated with sampled actions; HBPO reports its target policy $\pi$. Appendix~\ref{app:protocol} details baseline adaptations, initialization, tuning, evaluation, and separately accounted simulation costs. 

\subsection{Main Results}
\label{sec:results}

Table~\ref{tab:main-results} classifies final costs as Non-collapse ($c\le1010$), Collapse ($1010<c\le3980$), or Severe Collapse ($c>3980$). Thresholds 1010 and 3980 represent one and four periods of backlog-cost accumulation, respectively, indicating economic severity rather than delivery delays (Appendix~\ref{app:cost_thre}). Appendix~\ref{app:results_main} provides high-precision results and cost-grade calculations.

\begin{table}[!htb]
\centering
\begingroup
\definecolor{hbpocollapse}{RGB}{165,115,0}
\definecolor{hbposevere}{RGB}{175,34,39}
\setlength{\tabcolsep}{1.5pt}
\renewcommand{\arraystretch}{1.0}
\fontsize{8.2}{9.4}\selectfont
\caption{Final policy costs at a matched budget of 2M environment interactions across five training seeds. H denotes warm-start HBPO; Cold denotes HBPO without a native training prefix. Seed costs are black for Non-collapse, yellow for Collapse, and red for Severe Collapse. Std is the sample standard deviation across seeds; bold marks the lower paired mean or median, computed from unrounded values. Underlining marks HBPO endpoints with costs $\le1010$ whose seed-matched native endpoints exceed 1010. k/M denote $10^3/10^6$.}
\label{tab:main-results}
\par\vspace{4pt}
\resizebox{\linewidth}{!}{%
\begin{tabular}{@{}l*{8}{r}@{\hspace{7pt}}*{8}{r}@{}}
\toprule
& \multicolumn{8}{c}{\textbf{DSIS}: Mean cost per period $\downarrow$} 
& \multicolumn{8}{c}{\textbf{DSIS-RB}: Normalized risk-AUC cost $\downarrow$} \\
\cmidrule(lr){2-9}\cmidrule(l){10-17}
& \multicolumn{5}{c}{Seed cost} & \multicolumn{3}{c}{Across seeds} & \multicolumn{5}{c}{Seed cost} & \multicolumn{3}{c}{Across seeds} \\
\cmidrule(lr){2-6}\cmidrule(lr){7-9}\cmidrule(lr){10-14}\cmidrule(lr){15-17}
Algorithm & 0 & 1 & 2 & 3 & 4 & Mean & Median & Std & 0 & 1 & 2 & 3 & 4 & Mean & Median & Std \\
\midrule
PPO & 21.0 & 21.2 & 56.9 & \textcolor{hbposevere}{13.8k} & 21.3 & 2.79k & 21.3 & 6.17k & 70.1 & 75.5 & \textcolor{hbposevere}{67.0k} & 75.1 & 74.9 & 13.5k & \textbf{75.1} & 29.9k \\
H-PPO & 20.7 & 21.3 & 51.4 & \underline{159} & 20.8 & \textbf{54.6} & \textbf{21.3} & 59.8 & 52.7 & 81.9 & \textcolor{hbposevere}{5.89k} & 55.3 & 112 & \textbf{1.24k} & 81.9 & 2.60k \\
\addlinespace[1.5pt]
A2C & 23.7 & 28.7 & \textcolor{hbposevere}{23.0k} & 30.4 & 23.4 & 4.62k & 28.7 & 10.3k & 74.5 & 69.6 & \textcolor{hbposevere}{8.04k} & 807 & 69.6 & 1.81k & \textbf{74.5} & 3.49k \\
H-A2C & 32.3 & 20.8 & \textcolor{hbposevere}{6.64k} & 20.9 & 20.7 & \textbf{1.35k} & \textbf{20.9} & 2.96k & 51.6 & 51.3 & \textcolor{hbposevere}{7.40k} & 126 & 654 & \textbf{1.66k} & 126 & 3.22k \\
\addlinespace[1.5pt]
Discrete SAC & 27.8 & 36.7 & 38.6 & 27.5 & 345 & 95.1 & 36.7 & 140 & \textcolor{hbpocollapse}{1.18k} & \textcolor{hbposevere}{11.0k} & 71.2 & 69.7 & 102 & \textbf{2.48k} & 102 & 4.77k \\
H-Discrete SAC & 26.6 & 34.0 & 33.1 & 32.4 & 55.4 & \textbf{36.3} & \textbf{33.1} & 11.1 & \textcolor{hbposevere}{5.36k} & \textcolor{hbposevere}{7.03k} & 61.8 & 58.9 & 66.4 & 2.51k & \textbf{66.4} & 3.41k \\
\addlinespace[1.5pt]
ACE & \textcolor{hbpocollapse}{1.68k} & 22.6 & 22.3 & \textcolor{hbposevere}{7.50k} & \textcolor{hbpocollapse}{1.12k} & 2.07k & 1.12k & 3.12k & \textcolor{hbposevere}{9.96k} & \textcolor{hbposevere}{12.2k} & \textcolor{hbposevere}{22.3k} & \textcolor{hbposevere}{13.8k} & \textcolor{hbposevere}{15.5k} & 14.8k & 13.8k & 4.70k \\
H-ACE & \textcolor{hbpocollapse}{1.21k} & 320 & 21.4 & \underline{77.8} & \underline{37.5} & \textbf{333} & \textbf{77.8} & 504 & \textcolor{hbposevere}{4.01k} & \underline{110} & \underline{798} & \underline{121} & \underline{85.1} & \textbf{1.03k} & \textbf{121} & 1.70k \\
\addlinespace[1.5pt]
PPO-81 & 22.4 & 22.7 & 21.7 & 23.0 & 21.5 & 22.2 & 22.4 & 0.633 & 191 & 103 & 96.1 & 95.3 & 95.3 & 116 & 96.1 & 42.1 \\
H-PPO-81 & 21.4 & 22.0 & 21.2 & 20.9 & 21.5 & \textbf{21.4} & \textbf{21.4} & 0.418 & 125 & 80.0 & 86.6 & 84.8 & 84.1 & \textbf{92.1} & \textbf{84.8} & 18.5 \\
\addlinespace[1.5pt]
TRPO & \textcolor{hbposevere}{5.03k} & \textcolor{hbposevere}{12.5k} & 26.0 & 29.7 & 25.4 & 3.52k & \textbf{29.7} & 5.46k & 428 & 366 & 318 & \textcolor{hbpocollapse}{2.00k} & 293 & 680 & 366 & 738 \\
H-TRPO & \underline{103} & \underline{219} & 25.6 & 30.8 & 25.1 & \textbf{80.7} & 30.8 & 84.1 & 317 & 290 & 303 & \underline{235} & 166 & \textbf{262} & \textbf{290} & 61.9 \\
\addlinespace[1.5pt]
GRPO & 37.6 & 32.9 & 32.3 & 50.7 & 30.9 & 36.9 & 32.9 & 8.11 & 537 & \textcolor{hbpocollapse}{1.24k} & \textcolor{hbpocollapse}{1.10k} & \textcolor{hbpocollapse}{3.23k} & \textcolor{hbpocollapse}{1.22k} & 1.47k & 1.22k & 1.03k \\
H-GRPO & 22.5 & 26.8 & 21.0 & 23.2 & 21.9 & \textbf{23.1} & \textbf{22.5} & 2.24 & 80.8 & \underline{92.3} & \underline{107} & \underline{93.2} & \underline{222} & \textbf{119} & \textbf{93.2} & 58.3 \\
\addlinespace[1.5pt]
TPO & \textcolor{hbposevere}{4.21k} & \textcolor{hbposevere}{4.80k} & \textcolor{hbposevere}{4.56k} & \textcolor{hbposevere}{4.17k} & \textcolor{hbposevere}{4.40k} & 4.43k & 4.40k & 262 & \textcolor{hbposevere}{11.2k} & \textcolor{hbpocollapse}{1.62k} & \textcolor{hbpocollapse}{2.86k} & \textcolor{hbpocollapse}{1.99k} & \textcolor{hbpocollapse}{3.39k} & 4.22k & 2.86k & 3.98k \\
H-TPO & \underline{45.1} & \underline{44.8} & \underline{45.9} & \underline{44.6} & \underline{45.7} & \textbf{45.2} & \textbf{45.1} & 0.552 & \underline{238} & \underline{199} & \underline{220} & \underline{215} & \underline{163} & \textbf{207} & \textbf{215} & 28.1 \\
\addlinespace[1.5pt]
GPS & 123 & 120 & 107 & 118 & 124 & 118 & 120 & 6.93 & \textcolor{hbpocollapse}{1.77k} & \textcolor{hbposevere}{6.22k} & \textcolor{hbpocollapse}{2.41k} & \textcolor{hbpocollapse}{1.92k} & \textcolor{hbpocollapse}{2.01k} & 2.87k & 2.01k & 1.89k \\
H-GPS & 39.8 & 41.7 & 41.2 & 44.4 & 35.5 & \textbf{40.5} & \textbf{41.2} & 3.28 & \underline{234} & \underline{200} & \underline{188} & \underline{240} & \underline{216} & \textbf{216} & \textbf{216} & 21.8 \\
\midrule
Cold & 44.6 & 47.5 & 45.1 & 45.7 & 42.7 & 45.1 & 45.1 & 1.75 & 69.7 & 73.1 & 74.9 & 70.0 & 72.8 & 72.1 & 72.8 & 2.21 \\
\bottomrule
\end{tabular}%
}
\endgroup
\end{table}
HBPO lowers mean final cost in 17 of 18 warm source--environment combinations, median cost in 15, and sample standard deviation across seeds in all 18. Discrete SAC on DSIS-RB is the mean-cost exception (1.45\% higher), despite a lower median and standard deviation. For one seed, its native endpoint is Collapse while HBPO's is Severe Collapse. \textbf{High-cost native endpoints.} Among 36 seed-matched comparisons with native final costs above 1010, HBPO achieves lower costs in 35 and costs $\le1010$ in 29 (DSIS: 10/12; DSIS-RB: 19/24). \textbf{Low-cost native endpoints.} All 54 native endpoints with costs $\le1010$ have seed-matched HBPO endpoints within this range, with 46 achieving lower costs. \textbf{Cold starts.} Cold avoids final cost collapse across all five seeds in both environments, with mean costs of 45.1 on DSIS and 72.1 on DSIS-RB. By median cost over all five seeds, Cold ranks seventh among Cold and the nine native baselines on DSIS and first on DSIS-RB.

\subsection{Tracing Heuristic–Behavior–Target Transmission}
\label{sec:transmission recog}

\begin{figure}[!htb]
    \centering
    \includegraphics[width=\linewidth]{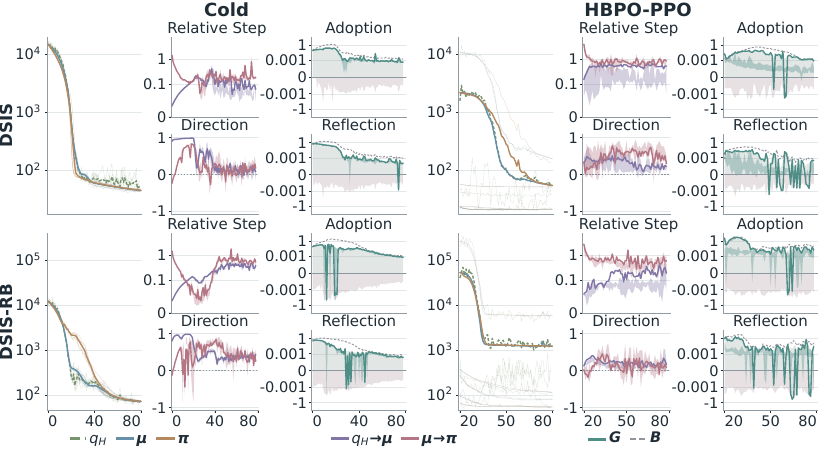}
    \caption{
    Transmission analysis. Left: standard evaluation costs of $\mu,\pi$ and a $q_H$ reference averaging $H$-weighted immediate costs over the original 9-stages rollouts (lower is better). Middle: relative step $\varrho$ (top) and direction cosine $c$ (bottom) for $q_H\to\mu$ and $\mu\to\pi$. Right: adoption and reflection, with net responses $G_\pm$, dashed opportunities $B_\pm$, and shaded favorable/adverse contributions above/below zero.
}
    \label{fig:p-transmission-analysis}
\end{figure}

These diagnostics characterize heuristic uptake and subsequent advantage-aligned target responses during policy recovery and stabilization. Figure~\ref{fig:p-transmission-analysis} examines the delayed transmission chain $\pi_g\to q_{H,g}\to\mu_{g+1}\to\pi_{g+2}$ (red in Section~\ref{fig:hbpo-framework}). Relative steps measure update size relative to each guidance gap
\(
\varrho_{H\mu}
=\frac{\|\mu_{g+1}-\mu_g\|_H}
       {\|q_{H,g}-\mu_g\|_H},\,
\varrho_{\mu\pi}
=\frac{\|\pi_{g+2}-\pi_{g+1}\|_J}
       {\|\mu_{g+1}-\pi_{g+1}\|_J}
\)
Here, $0$ denotes no update and $1$ equal update and gap norms. Larger ratios need not be better: the two updates balance $A$ with $H$ and $J$ with $P$ respectively. Weighted cosines measure alignment with the same gaps:
\(
c_{H\mu}
=\cos_H\!\left(
  \mu_{g+1}-\mu_g,\,
  q_{H,g}-\mu_g
\right),\,
c_{\mu\pi}
=\cos_J\!\left(
  \pi_{g+2}-\pi_{g+1},\,
  \mu_{g+1}-\pi_{g+1}
\right)
\)
Values $-1$, $0$, and $1$ indicate opposite, orthogonal, and fully aligned directions. Under the frozen target advantage $\widehat a_{g+1}$,
$S_\pm=\mathbf1\{
\pm(\mu_{g+1}-\pi_{g+1})\widehat a_{g+1}>0
\}$
identify advantageous and disadvantageous behavioral deviations. Their opportunity magnitudes and net target responses are
\(
B_\pm
=\left\langle
  \left[\pm(\mu_{g+1}-\pi_{g+1})
  \widehat a_{g+1}\right]_+
\right\rangle_P,
G_\pm
=\left\langle
  (\pi_{g+2}-\pi_{g+1})
  \widehat a_{g+1}S_\pm
\right\rangle_P
\).
Positive $G_+$ and $G_-$ indicate adoption of advantageous deviations and correction of disadvantageous ones respectively. Norms, cosines, and averages use the corresponding diagnostic weights.

Together, these observations provide evidence of effective heuristic–behavior–target transmission, whose corrective dynamics help explain recovery from policy collapse, improved average performance, and stable cold-start learning. \textbf{Existence.} In Fig.~\ref{fig:p-transmission-analysis}, broadly corresponding trends in the short-horizon $q_H$ reference and the behavior and target costs accompany nonzero relative steps and predominantly positive alignment along $q_{H,g}\to\mu_{g+1}\to\pi_{g+2}$. Together, these indicate sustained guidance, with target improvement sometimes following behavior improvement. \textbf{Corrective effectiveness.} During the pronounced DSIS--PPO cost reduction, $\mu\to\pi$ exhibits relatively large steps and closer alignment with nonzero $q_H\to\mu$ steps, alongside mostly positive adoption and reflection responses. This connects continued heuristic uptake with target adjustments that, on balance, retain advantageous behavioral deviations and oppose disadvantageous ones. Positive reflection also explains how useful transfer can coexist with weak geometric following. \textbf{Interpretability.} In DSIS--Cold, strong early alignment and substantial selection responses give way to weaker following and smaller $B,G$ as costs stabilize. DSIS-RB--Cold shows a complementary pattern: the target lags behind behavior when relative transfer steps are small, then catches up as these steps increase under positive alignment. Changes in transmission thus correspond to identifiable phases of cost improvement. The appendix shows related correspondences: DSIS--TPO exhibits diminishing opportunities and responses after rapid recovery, whereas DSIS--ACE combines positive adoption and reflection with weak or negative alignment and falling costs. In DSIS--TRPO, an interval of negative reflection instead accompanies subsequent cost increases, helping interpret temporary setbacks. These patterns characterize how selective transmission operates during learning. The component interventions below test its contribution to final policy performance.

\subsection{Ablation Results}
\label{sec:ablations}

\begin{figure}[!htb]
    \centering
    \includegraphics[width=\linewidth]{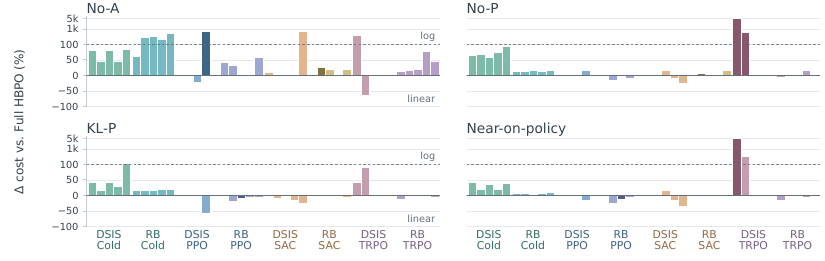}
    \caption{Final target-policy cost changes (\%) relative to matched Full HBPO; positive values favor Full. Each color represents one environment--source, with seeds 0--4 from left to right. Dark bars indicate ablation costs above 1010. Shared axes are linear up to 100\% and logarithmic above.}
    \label{fig:p-transmission-ablation}
\end{figure}
\textbf{No-A} removes target-advantage feedback from behavior learning; \textbf{No-P} removes explicit $\mu\to\pi$ transfer while retaining behavior-generated data. \textbf{KL-P} replaces $P$ with KL imitation at a matched natural-gradient norm. \textbf{Near-on-policy} uses the newly updated behavior policy for immediate KL transfer, then sets $\mu=\pi$. In Figure~\ref{fig:p-transmission-ablation}, Full consistently outperforms all four ablations from Cold and avoids the high-cost failures under No-P and Near-on-policy in DSIS--TRPO. Together with broad degradation under No-A, these results support target-informed behavior learning and explicit transfer beyond data collection. KL-P and Near-on-policy often yield slightly lower costs from PPO and SAC starts, indicating a trade-off with protection against high-cost failures. Mean paired relative-cost changes favor Full for all four ablations.

\begin{figure}[!htb]
    \centering
    \includegraphics[width=\linewidth]{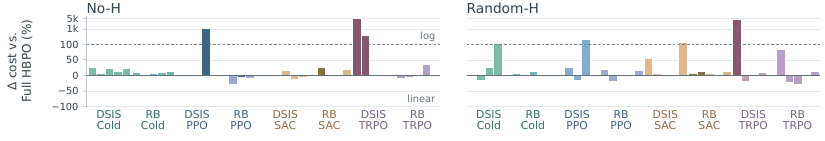}
    \caption{Heuristic-guidance ablations, No-H and Random-H. Plotting conventions follow Figure~\ref{fig:p-transmission-ablation}.}
    \label{fig:h-guidance-ablation}
\end{figure}
\textbf{No-H} removes $H$ and its weighting and geometry from the $\operatorname{Fuse}$ behavior update while retaining behavior--target transmission. \textbf{Random-H} generates random guidance targets around $\mu$, matching the statewise distance of the simulation-derived $q_H$ from $\mu$. Both increase mean final cost in most tested settings. In these DSIS ablations, No-H and Random-H produce three and one collapsed endpoints, respectively, versus none for Full. Random-H occasionally outperforms Full but costs more in 32 of 40 matched runs, supporting the value of informative guidance directions.
\section{Discussion and Limitations}
\label{sec:discussion}

First, generation-level diagnostics and component ablations together support heuristic–behavior–target transmission as a mechanism contributing to policy recovery and improvement. The delayed path follows from the frozen-parent update schedule; the experiments assess the coupled learning mechanism and its component channels. Emphatic weighting adapts the principle of propagating state interest through policy-dependent transitions to HBPO’s two learning roles, providing separate state emphasis for target-return learning and heuristic-informed transfer. The ablations in Appendix~\ref{app:emphatic-results} show mixed endpoint effects, so a consistent final-cost benefit is not established in the evaluated settings. We adopt this pathway considering its theoretical advantages, which not obviously show in experiment endpoint. Achieving sustained state-distribution correction remains an opportunity to strengthen this framework. Second, HBPO improves 35 of 36 comparisons with high-cost native endpoints. For DSIS-RB SAC seed 0, cost falls from 7.76k at handoff to 5.36k but exceeds native continuation’s 1.18k, highlighting limits to recovery across initializations. Third, the coupling rules are not inventory-specific, though HBPO uses structural priors in \(\widehat{\mathcal P}\). Shared information access and comparison with model-informed GPS support the architectural comparison under common modeling assumptions. Isolating the contribution of structural priors and testing other domains remain future directions.
\section{Conclusion}
\label{sec:conclusion}

We introduced HBPO, built on an explicit delayed heuristic–behavior–target coupling mechanism. Across DSIS and DSIS-RB, HBPO recovers many high-cost policies, improves overall performance and avoids final collapse in all tested cold starts. Transmission diagnostics trace how heuristic guidance shapes behavior learning and subsequent selective target updates, helping explain these gains. Ablations further support the contributions of informative heuristic guidance, target-informed behavior learning, and explicit behavior–target transfer to performance and robustness.
\label{main-text-end}

\section*{AI Use Statement}
We used OpenAI Codex with GPT-5.5 Extra high, GPT-5.6 Sol Ultra and GPT-6 Astra Ultra, primarily in regular mode and occasionally in fast mode. These were the only generative AI assistants used in this work. Codex assisted formalizing the mathematics of the authors' method, developing and checking proofs, and preparing and inspecting the appendix. It also generated the method implementation and ran and debugged experiments under the authors' supervision. The authors polished the main section, identified and reviewed related work, and checked references, with assistance from Codex. The authors conceived the method and reinforcement-learning framework, designed the experiments, analyzed the results, and produced the figures.

The authors carefully drafted the main section, reviewed the manuscript, read the AI-assisted proofs, and manually worked through selected derivations. Their code review consisted of inspecting key implementation components. The authors take full responsibility for the final manuscript and accompanying code, including the mathematical claims, references, and experimental results.

\section*{Reproducibility Statement}
Appendix~\ref{app:notation} summarizes the notation. Appendix~\ref{app:algorithm} specifies the algorithm, benchmark dynamics, information access, initialization procedures, and mathematical assumptions and derivations. Appendix~\ref{app:protocol} describes hyperparameter selection, interaction budgets, random seeds, and evaluation. Appendix~\ref{app:results} reports extended results, and Appendix~\ref{app:ablations} details the ablation interventions and diagnostic procedures. An anonymized code package is included in the supplementary material, with the implementation, experiment configurations, dependency specifications, and instructions for running training and evaluation. The code will be made publicly available upon acceptance.

\bibliography{references}
\bibliographystyle{iclr2027_conference}

\clearpage
\appendix
\section{Notation}
\label{app:notation}
Table~\ref{tab:notation} summarizes the benchmark, model, critic, and
policy-update notation. Its main entries follow Appendix~\ref{app:algorithm};
the final groups record notation used in the main text and the other
appendices. Scope-qualified entries distinguish local uses of the same
letter rather than identifying different mathematical objects.

\begingroup
\small
\setlength{\tabcolsep}{4pt}
\renewcommand{\arraystretch}{1.18}
\begin{longtable}{@{}>{\raggedright\arraybackslash}p{0.35\linewidth}>{\raggedright\arraybackslash}p{\dimexpr0.65\linewidth-2\tabcolsep\relax}@{}}
\caption{Notation and scope. The generation-$g$ model, policies, reference critic, and sampling measures are frozen when constructing that generation's learning signals.}
\label{tab:notation}\\
\toprule
Symbol & Meaning and convention \\
\midrule
\endfirsthead
\multicolumn{2}{@{}l}{\tablename~\thetable\ (continued)}\\
\toprule
Symbol & Meaning and convention \\
\midrule
\endhead
\midrule
\multicolumn{2}{r@{}}{Continued on next page}\\
\endfoot
\bottomrule
\endlastfoot

\multicolumn{2}{@{}l}{\textbf{Indices, observations, actions, and physical dynamics}}\\*[2pt]
$g;\allowbreak\ t;\allowbreak\ k$
& Learning generation; physical period; simulated period, optimization iteration, or sampled-root index, as specified locally. \\
$j;\allowbreak\ b;\allowbreak\ \chi$
& Critic member $j\in\{1,\ldots,5\}$; simulation-tape index; weighting role ($J$ for target return, $P$ for transmission, and $0$ for the unweighted behavior-return role). The risk grid separately uses $j=0,\ldots,5$. \\
$\mathcal S;\allowbreak\ s,S;\allowbreak\ s',S'$
& Observation space; an observation and its random counterpart; successor observation and its random counterpart. In the residual theorem, $\mathcal S$ is countable and closed under the frozen model's transitions. \\
$I_t;\allowbreak\ \mathbf p_t=(p_{t,1},\ldots,p_{t,4})$
& Net inventory (negative values represent backlog) and the four-stage regular-order pipeline. \\
$B_t;\allowbreak\ B(s);\allowbreak\ u,U$
& Current binary regular-supply availability, its observation-coordinate accessor, and realized/random next availability. Zero means unavailable and one means available. \\
$\mathsf D_t,d_t;\allowbreak\ d^-(s)$
& Random demand and its realization; the previous-demand coordinate of an RB observation. The demand coordinate is omitted in DSIS. \\
$\mathcal A;\allowbreak\ a,\mathsf A;\allowbreak\ a_t=(q_t^r,q_t^e)$
& The $81$-action set $\{0,\ldots,8\}^2$; a realized/random action; regular and expedited order quantities. The symbols $q^r,q^e$ denote quantities only. \\
$F(s,a,d,u)$
& Deterministic observation-transition map from the public inventory balance, pipeline shift, current-availability order gate, and next availability $u$. \\
$c(s,a,d),c_t;\allowbreak\ r(s,a,d),r_t;\allowbreak\ R,R_k$
& Period cost and its realization; normalized reward and its realization; random one-step or step-$k$ model reward. Here $r=-c/495$, the common unit for values and utilities. \\
$[x]_+;\allowbreak\ \gamma$
& Positive part $\max\{x,0\}$; critic discount factor, equal to $0.995$ in the implementation. \\
$\delta_x;\allowbreak\ \mathbf 1\{E\};\allowbreak\ \otimes$
& Unit point mass at $x$; indicator of statement $E$; product of probability measures. \\

\multicolumn{2}{@{}l}{\textbf{Benchmark laws and risk stratification}}\\*[2pt]
$\rho;\allowbreak\ X_t=(S_t,\rho)$
& Actual environment risk, fixed within a trajectory; the corresponding augmented physical state. Risk is not observed by the policy or critic. \\
$\mathcal D_{\mathrm{RB}};\allowbreak\ \omega_D$
& True RB demand support $\{0,1,2,3,8\}$ and stationary demand distribution $(0.3,0.2,0.2,0.2,0.1)$ in that displayed order. \\
$K_D(x,d);\allowbreak\ K_{B,\rho}(b_{\mathrm{cur}},u)$
& True demand-transition probability given previous demand $x$; true supply-transition probability given current availability $b_{\mathrm{cur}}$ and actual risk $\rho$. \\
$\mathcal P_\rho(dr,ds'\mid s,a)$
& True joint reward--observation transition kernel conditional on environment risk $\rho$; it excludes scheduled episode resets. \\
$H_{\mathrm{env}};\allowbreak\ \rho_j$
& Physical episode length, $1000$ periods; registered risk grid $\rho_j=0.1j$, $j=0,\ldots,5$. \\
$\omega_{\rho,j};\allowbreak\ \omega_\rho;\allowbreak\ \nu_{\mathrm{env}}$
& Risk-grid weight, the weight vector $(0.1,0.2,0.2,0.2,0.2,0.1)$, and the target risk-composition measure $\sum_j\omega_{\rho,j}\delta_{\rho_j}$. \\
$\mathbf n^{(0)},\mathbf n^{(1)}$
& Alternating $25$-trajectory risk-stratum count vectors, $(3,5,5,5,5,2)$ and $(2,5,5,5,5,3)$; two blocks have exactly the target risk proportions. \\
$J_\pi(\rho_j);\allowbreak\ \mathcal R_{\mathrm{trap}}(\pi)$
& Expected episode mean cost under fixed policy $\pi$ at risk $\rho_j$; normalized trapezoidal risk score $\sum_j\omega_{\rho,j}J_\pi(\rho_j)$. \\
$Y_\ell$
& Realized mean cost of trajectory $\ell$ in the fixed-policy, $50$-trajectory stratification calculation. \\

\multicolumn{2}{@{}l}{\textbf{Observed-data model and persistent synthetic risk}}\\*[2pt]
$\mathscr D_{\le g}^{\mathrm{real}};\allowbreak\ N_g^{\mathrm{real}}$
& Multiset of all counted real transitions through generation $g$, including the initialization prefix when applicable; its size. \\
$\widehat{\mathcal D}_g;\allowbreak\ d_{\max}$
& Demand support discovered from current and successor previous-demand coordinates; its maximum, used in the inventory-growth bound. \\
$\mathcal C_g(x,y);\allowbreak\ \mathcal C_g(x)$
& Number of observed demand transitions $x\to y$; their row total $\sum_y\mathcal C_g(x,y)$. \\
$\widehat\omega_g;\allowbreak\ \alpha_D$
& Pooled empirical next-demand distribution; total pseudocount mass per fitted demand row, equal to one. \\
$\widehat K_{D,g}(x,y)$
& Fitted demand-transition probability, $(\mathcal C_g(x,y)+\alpha_D\widehat\omega_g(y))/(\mathcal C_g(x)+\alpha_D)$, for $x,y\in\widehat{\mathcal D}_g$. \\
$\mathcal C_g^\downarrow;\allowbreak\ \mathcal C_g^{\mathrm{rec}}$
& Counts of observed transitions starting in an outage and of those that recover to availability. \\
$\kappa^{\mathrm{rec}};\allowbreak\ \widehat\kappa_g^{\mathrm{rec}}$
& Common recovery probability in the fitted supply family; its smoothed estimate $(\mathcal C_g^{\mathrm{rec}}+1)/(\mathcal C_g^\downarrow+2)$. The superscript distinguishes it from heuristic hyperparameters. \\
$Z,z;\allowbreak\ m$
& Synthetic risk and its realization; its prior $m=\operatorname{Unif}[0,1/2]$. Each tape draws risk independently at its query root and retains that risk throughout the tape. \\
$\widehat K_{B,g,z}(b_{\mathrm{cur}},u)$
& Fitted supply-transition probability conditional on synthetic risk $z$; recovery probability is $\widehat\kappa_g^{\mathrm{rec}}$ and available-to-outage probability is $\widehat\kappa_g^{\mathrm{rec}}z/(1-z)$. \\
$\widehat p_g(d,u\mid s,a,z)$
& Modeled joint demand/next-availability probability mass function, factoring into the fitted demand and supply kernels; one-variable forms denote its marginals. \\
$\widehat{\mathcal P}_{g,z}(dr,ds'\mid s,a)$
& Frozen fitted reward--observation kernel conditional on one retained synthetic risk $z$. \\
$\widehat{\mathcal P}_g$
& Full simulation specification: a query observation, an independent root draw $Z\sim m$, and subsequent transitions retaining $Z$. It does not denote an observation-only Markov kernel with risk redrawn at each step. \\
$\widetilde{\mathcal P}_g(dr,ds',dz'\mid s,z,a)$
& Augmented reward--observation--risk kernel $\widehat{\mathcal P}_{g,z}(dr,ds'\mid s,a)\delta_z(dz')$, preserving risk. \\

\multicolumn{2}{@{}l}{\textbf{Policies, scalar critics, and Bellman operators}}\\*[2pt]
$\mu_g;\allowbreak\ \pi_g;\allowbreak\ \pi_0;\allowbreak\ \sigma$
& Frozen parent behavior policy; frozen parent target policy; initial categorical policy; a generic fixed policy used to define an augmented transition kernel. \\
$\phi(s);\allowbreak\ d_\phi$
& Fixed normalized network feature vector; its dimension, $7$ in RB and $6$ in DSIS. \\
$V_\theta;\allowbreak\ \theta$
& A live scalar critic and its parameter vector $(W_1,b_1,W_2,b_2,w_3,b_3)$: two width-$128$ Tanh layers followed by a scalar linear output. \\
$W_1,W_2;\allowbreak\ b_1,b_2;\allowbreak\ w_3,b_3$
& Hidden-layer weight matrices and biases; output weight vector and scalar bias of $V_\theta$. Their dimensions are specified with the network definition. \\
$V_{g,j},\theta_{g,j};\allowbreak\ v_g$
& Generation-$g$ critic member and its parameters; frozen reference $v_g=\tfrac15\sum_{j=1}^5V_{g,j}$, averaging outputs rather than parameters. \\
$L_g^uV(s,a)$
& Discounted action backup integrating model demand at fixed next availability $u$: expected reward plus $\gamma$ times successor value. \\
$L_{g,z}V(s,a)$
& Discounted action backup obtained from $L_g^u$ by also integrating next availability under $\widehat K_{B,g,z}$. \\
$\mathcal T_g^u V;\allowbreak\ \mathcal T_{g,z}^{\pi_g}V$
& Target-policy averages of $L_g^uV$ and $L_{g,z}V$, respectively; the former holds next availability fixed, whereas the latter integrates it conditional on risk $z$. \\
$\widehat P_{g,z}^{a};\allowbreak\ \widehat P_{g,z}^{\pi_g}$
& One-step state-transition operators acting on scalar functions: first action forced to $a$, or sampled from $\pi_g$, under fixed risk $z$. They contain neither a reward term nor a discount factor. \\
$\mathrm{Id};\allowbreak\ (\widehat P_{g,z}^{\pi_g})^\ell$
& Identity operator on scalar functions; $\ell$ successive target-policy transition operators, with zeroth power equal to $\mathrm{Id}$. \\
$\mathbb E_z^{s,a};\allowbreak\ \mathbb E_z^s$
& Frozen-model rollout expectation at fixed risk $z$ from observation $s$, with first action forced to $a$, or also sampled from $\pi_g$; all later actions follow $\pi_g$. \\

\multicolumn{2}{@{}l}{\textbf{Conditional values, residual completion, and sampled advantages}}\\*[2pt]
$V_{g,z}^{\pi_g};\allowbreak\ \Delta_{g,z}$
& Exact infinite-horizon discounted model value at fixed risk $z$; conditional value error $V_{g,z}^{\pi_g}-v_g$. \\
$e_{g,z}$
& Frozen-reference Bellman residual $\mathcal T_{g,z}^{\pi_g}v_g-v_g$ (backup minus prediction), integrating the target action and model disturbances conditional on $z$. \\
$C_v$
& Finite uniform bound on $|v_g(s)|$ in the residual-completion theorem; it does not bound exact values or rewards uniformly over all initial states. \\
$\bar V_g;\allowbreak\ c_{\mathrm{val}}$
& Root-risk-averaged exact conditional value $\mathbb E_{Z\sim m}[V_{g,Z}^{\pi_g}]$; a scalar prediction used in the conditional-value variance calculation. \\
$Q_g^{\pi_g};\allowbreak\ Q_g^0$
& Exact discounted model action value with an independent persistent root risk; one-step base $\mathbb E_{Z\sim m}[L_{g,Z}v_g]$. \\
$n_{\mathrm{tr}}$
& Deterministic residual-continuation truncation index used in the proof; the remaining evaluation error is at step $n_{\mathrm{tr}}+1$. \\
$N,N_b$
& Geometric residual-continuation length, with $\Pr(N=\ell)=(1-\gamma)\gamma^\ell$ for $\ell\ge0$; its draw on tape $b$. Geometric survival supplies the discount. \\
$M;\allowbreak\ Z_b;\allowbreak\ S_{b,k}^{a}$
& Number of residual tapes per sampled query, $M=32$; tape-$b$ risk; its step-$k$ observation after forced first action $a$ and target-policy continuation. \\
$\widehat Q_g$
& Ideal Monte Carlo action-value estimator: the dense base plus the average geometrically stopped sum of unweighted residuals. The implementation may retain only its centered correction. \\
$A_g^{\mathrm{adv}};\allowbreak\ A_g^{\mathrm{base}}$
& Exact-model and one-step-base advantages, each centered over actions using $\pi_g(\cdot\mid s)$. \\
$\widehat A_g$
& Sampled advantage $\widehat Q_g(s,a)-\sum_{a'}\pi_g(a'\mid s)\widehat Q_g(s,a')$. \\
$\Delta_g^Q(s,a)$
& Actionwise evaluation error $Q_g^{\pi_g}(s,a)-Q_g^0(s,a)$; centering removes it exactly only when it is constant across actions at the given state. \\

\multicolumn{2}{@{}l}{\textbf{Emphatic root measures and successor-critic fitting}}\\*[2pt]
$n;\allowbreak\ s_i;\allowbreak\ D_g$
& Number of generation-batch occurrences, $n=25{,}000$; occurrence-$i$ observation; empirical measure $D_g=n^{-1}\sum_i\delta_{s_i}$, retaining repetitions. \\
$w_i^\chi;\allowbreak\ \nu_g^\chi$
& Frozen normalized occurrence weight and weighted root measure $n^{-1}\sum_iw_i^\chi\delta_{s_i}$, for $\chi\in\{J,P\}$; $n^{-1}\sum_iw_i^\chi=1$. \\
$\widetilde D_g;\allowbreak\ \widetilde\nu_g^\chi$
& Augmented root measures $D_g\otimes m$ and $\nu_g^\chi\otimes m$, initially independent in observation and synthetic risk. \\
$\widetilde P_{g,\sigma};\allowbreak\ \lambda$
& Policy-$\sigma$ transition kernel on observation--risk pairs, preserving risk; a generic augmented-state distribution. The measure $\lambda\widetilde P_{g,\sigma}$ is its one-step successor law. \\
$\widetilde\eta_g$
& Augmented distribution after one consumer transition: the equal mixture of roots/first-action policies $(D_g,\mu_g)$, $(\nu_g^J,\pi_g)$, and $(\nu_g^P,\mu_g)$. \\
$\widetilde\zeta_g$
& Critic-training state--risk law $(1-\gamma)\sum_{\ell\ge0}\gamma^\ell\widetilde\eta_g\widetilde P_{g,\pi_g}^{\,\ell}$; its zero-index term is already after the first transition. \\
$G;\allowbreak\ n_\zeta;\allowbreak\ (S_\ell,U_\ell)$
& Additional geometric target-policy transition count when sampling a training-bank row; bank size $n_\zeta=25{,}000$; retained observation and a newly sampled next availability under that row's retained risk. \\
$\varepsilon_g(\theta;s,u)$
& Live fitting error $V_\theta(s)-(\mathcal T_g^uV_\theta)(s)$ (prediction minus backup), distinct from the frozen-reference residual $e_{g,z}$. \\
$\mathcal J_{V,g};\allowbreak\ \widehat{\mathcal J}_{V,g}$
& Population and empirical means of $\varepsilon_g^2$ for one critic member. Actions and demand are integrated before squaring; paired next availability is sampled. \\
$\theta_j^{(k)};\allowbreak\ V_{g+1,j};\allowbreak\ v_{g+1}$
& Live scalar parameters after $k$ updates from $\theta_{g,j}$; member after $1000$ updates; successor output mean, available to consumers in the next generation. \\
$\beta_1,\beta_2;\allowbreak\ \epsilon_{\mathrm{Adam}}$
& Adam first/second-moment decay coefficients $(0.9,0.999)$ and denominator stabilizer $10^{-8}$. \\

\multicolumn{2}{@{}l}{\textbf{Warm/cold initialization and temporary fitted Q evaluation}}\\*[2pt]
$n_0;\allowbreak\ (s_i,u_i)$
& Number of initialization pairs, $500{,}000$; observation/next-availability pair, recorded in warm initialization and model-generated in cold initialization. \\
$d_i;\allowbreak\ s'_{i,a},r_{i,a}$
& Demand realized in recorded source transition $i$; all-action factual counterfactuals $F(s_i,a,d_i,u_i)$ and $r(s_i,a,d_i)$. \\
$Q_j^{(k)},Q_j^{-,(k)};\allowbreak\ Q_j$
& Temporary online and target FQE networks at iteration $k$; final online network $Q_j=Q_j^{(2000)}$. These networks are discarded after warm initialization. \\
$\psi_j^{(k)},\psi_j^{-,(k)}$
& Online/target FQE parameter vectors. Target parameters are updated after each online step; these are distinct from scalar-critic parameters $\theta$. \\
$y_{j,i,a}^{(k)}$
& Detached FQE reward-plus-bootstrap target using the frozen initial categorical policy $\pi_0$ and target network $Q_j^{-,(k)}$. \\
$\mathcal I^{(k)};\allowbreak\ n_{\mathrm b};\allowbreak\ h$
& Ordered list of uniformly sampled source-occurrence indices, with replacement, at FQE iteration $k$; its size $256$; a position within that list. All five FQE members share the indices. \\
$\xi_{j,h,a}^{(k)}$
& Independent $\operatorname{Bernoulli}(0.8)$ FQE mask for member $j$, minibatch position $h$, action $a$, and iteration $k$. Repeated source indices may receive different masks. \\
$\mathcal L_j^{\mathrm{FQE},(k)}$
& Masked FQE squared-error loss, normalized by $n_{\mathrm b}|\mathcal A|$, including the masked-out pairs in the denominator. \\
$A_j^{\mathrm{init}}$
& Frozen centered action label from final online FQE member $j$: $Q_j(s,a)-\sum_{a'}\pi_0(a'\mid s)Q_j(s,a')$. \\
$L_{\mathrm{init}}^u;\allowbreak\ \mathcal T_{\mathrm{init}}^u$
& Supply-conditioned action backup using the initialization model; its average over the initial policy $\pi_0$. \\
$\mathcal J_j^{\mathrm{warm}};\allowbreak\ \mathcal J_j^{\mathrm{cold}}$
& Full-gradient scalar initialization objectives: actionwise fitting to $A_j^{\mathrm{init}}$ for warm initialization; squared policy-averaged Bellman discrepancy for cold initialization. \\
$\widetilde V_j;\allowbreak\ \overline e_j;\allowbreak\ c_{\mathrm{off}}$
& Scalar member before offset calibration; its full-bank mean backup-minus-prediction residual; a constant output shift, chosen as $\overline e_j/(1-\gamma)$. \\
$V_{0,j}$
& Calibrated initial critic member $\widetilde V_j+\overline e_j/(1-\gamma)$; its empirical mean residual vanishes in exact arithmetic, not necessarily pointwise. \\

\multicolumn{2}{@{}l}{\textbf{Heuristic utility and control variates}}\\*[2pt]
$H;\allowbreak\ G_{H,g}(s,a)$
& Heuristic horizon, $H=1000$; expected undiscounted $H$-period normalized reward after first action $a$, averaging independent root risk and target-policy continuation. \\
$v_{g,k}$
& Heuristic control function, equal to $v_g$ for $k<H$ and zero at $k=H$. \\
$L_{g,z,k}^{\mathrm H}(s,a)$
& Undiscounted control backup $\mathbb E[R+v_{g,k+1}(S')\mid s,a,z]$. Unlike $L_{g,z}V$, it is the evaluated backup rather than an operator with a free value-function argument. \\
$e_{g,z,k}^{\mathrm H}(s)$
& Heuristic control residual $\sum_{a'}\pi_g(a'\mid s)L_{g,z,k}^{\mathrm H}(s,a')-v_{g,k}(s)$; its successor-control coefficient is one, not $\gamma$. \\
$\widetilde G_{H,g,z};\allowbreak\ \widehat G_{H,g}$
& One-tape residual-control utility estimator at fixed risk; average over $32$ independently drawn tape risks and their coupled action continuations. \\
$h_g;\allowbreak\ a_\star$
& Estimated heuristic utility centered over actions using $\pi_g$; fixed reference action used when accumulating equivalent action differences. \\

\multicolumn{2}{@{}l}{\textbf{Actor surrogate and shared critic evidence}}\\*[2pt]
$\vartheta,\vartheta_g;\allowbreak\ \pi_\vartheta$
& Actor parameters and frozen parent parameters; the parameterized target policy, with $\pi_g=\pi_{\vartheta_g}$. \\
$\upsilon$
& Fixed state-sampling distribution for the local actor surrogate. \\
$\mathcal J_g^{\mathrm{actor}}(\vartheta;f,\upsilon)$
& Frozen-state surrogate $\mathbb E_{S\sim\upsilon}[\sum_a\pi_\vartheta(a\mid S)f(S,a)]$, for a frozen scalar action-evaluation function $f$. \\
$\operatorname{sc}_g(s,a)$
& Parent-policy score vector $\left.\nabla_\vartheta\log\pi_\vartheta(a\mid s)\right|_{\vartheta=\vartheta_g}$. \\
$K_{\mathrm{res}};\allowbreak\ \varpi_i;\allowbreak\ \mathsf I_k$
& Number of sampled residual roots, $128$; root probability $(1+w_i^J+w_i^P)/(3n)$; independently sampled occurrence index for root $k$, with replacement. \\
$w_i^0;\allowbreak\ \operatorname{coef}_i(a)$
& Unit weight for the unweighted behavior-return role; prescribed scalar coefficient for occurrence $i$ and action $a$, independent of the shared sampling when unbiasedness is asserted. \\
$\widehat A_g^{(k)}$
& Sampled centered advantage from the independent continuation bank attached to sampled root occurrence $k$; repeated selections receive distinct banks. \\
$\widehat\Lambda_g^\chi[\operatorname{coef}]$
& Shared linear critic-evidence functional for role $\chi\in\{0,J,P\}$: a dense weighted base plus an importance-weighted sampled centered correction. \\

\multicolumn{2}{@{}l}{\textbf{Main-text actors, weighting, and coupled updates}}\\*[2pt]
$\mathcal X;\allowbreak\ x_t;\allowbreak\ \mathcal P$
& Full environment state space, state, and transition kernel in the problem formulation. In RB, $x_t=(s_t,\rho)$; in DSIS, $x_t=s_t$. These differ from the observation space $\mathcal S$ and fitted model specification $\widehat{\mathcal P}_g$. \\
$\mathcal Y_g$
& Persistent learner tuple $(\mu_g,\pi_g,V_{g,1:5},M_g^J,M_g^P)$ in Methods; it is not the physical state $X_t$. \\
$\theta_g^\mu,\theta_g^\pi$
& Behavior/target actor parameters in Methods. The target parameter corresponds to $\vartheta_g$ in Appendix~\ref{app:algorithm}; unqualified $\theta$ in the critic construction denotes scalar-critic parameters. \\
$\mathcal D_g;\allowbreak\ w_g^t(s)$
& Methods notation for the behavior-data panel and normalized state weight, $t\in\{J,P\}$. Empirical observation averaging corresponds to $D_g$ in Appendix~\ref{app:algorithm}, and occurrence $i$ has weight $w_i^t=w_g^t(s_i)$. \\
$M_g^J,M_g^P$
& Positive learned raw emphatic masses for target-return and transmission weighting; the normalized consumer weights $w_g^J,w_g^P$ are derived from these masses. \\
$T;\allowbreak\ G_{T,g};\allowbreak\ \widehat G_{T,g};\allowbreak\ u_g$
& Methods notation for the heuristic horizon, exact utility, estimated utility, and centered estimated utility. These correspond to $H,G_{H,g},\widehat G_{H,g},h_g$ in Appendix~\ref{app:algorithm}, respectively. \\
$\Delta(\mathcal A);\allowbreak\ q;\allowbreak\ q_{H,g}$
& Probability simplex over actions; a candidate distribution in the Methods heuristic optimization; its Hellinger-regularized controller. These local uses of $q$ are distributions, whereas $q^r,q^e$ are order quantities. \\
$\tau;\allowbreak\ \kappa$
& Frozen run-level heuristic utility/regularization scale; the separately tuned heuristic scale hyperparameter in the protocol. Neither is the fitted supply recovery probability $\widehat\kappa_g^{\mathrm{rec}}$. \\
$\xi_g$
& Empirical distribution of states from short heuristic-controller rollouts in Methods. It is distinct from the indexed binary FQE mask $\xi_{j,h,a}^{(k)}$. \\
$\widehat P_{g,z}\ \text{(Methods)}$
& Reward--observation kernel in the Methods backup formulas, written $\widehat{\mathcal P}_{g,z}$ in Appendix~\ref{app:algorithm}; it is distinct from the appendix's state-transition operators with superscripts $a$ or $\pi_g$. \\
$Q(s,a;\pi_g);\allowbreak\ \mathbf a_i(a),\widehat{\mathbf a}_i$
& Methods notation for $Q_g^{\pi_g}$ and for its exact/estimated centered action evidence at observation occurrence $s_i$. Appendix~\ref{app:algorithm} spells out $A_g^{\mathrm{adv}}$ and its sampled correction. \\
$\Lambda_A,\Lambda_J,\Lambda_P;\allowbreak\ c_i^t;\allowbreak\ \ell_i$
& Methods advantage functionals, role weights $(c_i^A,c_i^J,c_i^P)=(1,w_i^J,w_i^P)$, and actionwise coefficients. The Methods role $A$ corresponds to role $0$ in the shared functional $\widehat\Lambda_g^\chi$. \\
$A(m),J(p),P(p);\allowbreak\ H$
& Local behavior-advantage, target-advantage, and discrepancy-weighted transmission surfaces; heuristic guidance channel in Methods. The bare channel label $H$ differs from the numeric heuristic horizon $H$ in Appendix~\ref{app:algorithm}. \\
$m,p\ \text{(Methods)}$
& Candidate updated behavior/target probability vectors in the local surfaces. This local $m$ is distinct from the synthetic-risk probability measure $m$ in Appendix~\ref{app:algorithm}. \\
$g_A,g_H,g_J,g_P$
& Frozen-parent channel gradients. The target-return channel uses $g_J=\nabla(J/\tau)$; the $A$ and $P$ surfaces already include the factor $1/\tau$. \\
$F_\mu,F_\pi;\allowbreak\ \operatorname{Fuse}_F$
& Actor-specific damped Fisher matrices; fusion map $F^{-1}(\alpha_1g_1+\alpha_2g_2)$. These matrices are distinct from the physical transition map $F(s,a,d,u)$. \\
$n_i;\allowbreak\ \alpha_1,\alpha_2;\allowbreak\ t_\mu,t_\pi$
& Channel natural norms $n_i=(g_i^\top F^{-1}g_i)^{1/2}$; fusion coefficients; accepted endpoint step fractions. The norms $n_i$ are not sample counts, and the coefficients need not sum to one. \\

\multicolumn{2}{@{}l}{\textbf{Evaluation and protocol-specific notation}}\\*[2pt]
$C_s(\rho),C_s^{\mathrm{RB}};\allowbreak\ \bar C,S_C^2$
& Evaluation mean cost at a risk and its risk-weighted aggregate for training seed $s$; mean and sample variance across training seeds. Here subscript $s$ indexes seeds rather than observations. \\
$X_0\ \text{cost}$
& Evaluated cost of the initialized target policy in the result/protocol discussion; it is not the augmented physical state $X_t$ or learner tuple $\mathcal Y_g$. \\
$C,NC;\allowbreak\ n_C,n_{NC};\allowbreak\ \Delta_C,\Delta_{NC}$
& Retrospective paired-endpoint cost groups, their sizes, and mean per-seed percentage changes from native to HBPO. Group thresholds follow the corresponding result table. \\
$Q_t$
& $Q_t$ here denotes an order quantity, not an action-value function. \\
$N,T\ \text{(evaluation)}$
& Number of evaluation tapes and periods per tape in the protocol. This deterministic sample count $N$ is distinct from the geometric residual length in Appendix~\ref{app:algorithm}. \\
$P_1,\ldots,P_4;\allowbreak\ q_r,q_e\ \text{(protocol)}$
& Protocol names for pipeline quantities and regular/expedited orders, written $p_1,\ldots,p_4$ and $q^r,q^e$ in Appendix~\ref{app:algorithm}. \\
$D;\allowbreak\ A;\allowbreak\ K_A(\rho)\ \text{(protocol)}$
& Protocol demand, availability, and true supply kernel, written $\mathsf D_t$, $B_t$, and $K_{B,\rho}$ in Appendix~\ref{app:algorithm}. In Methods model rollouts, $A_k$ instead denotes an action, written $\mathsf A_k$ in the appendix. \\
$\mathcal D_{\mathrm{demand}};\allowbreak\ p;\allowbreak\ \omega_k\ \text{(protocol)}$
& Protocol true-demand support, stationary demand probabilities, and risk-evaluation weights; these correspond to $\mathcal D_{\mathrm{RB}},\omega_D,\omega_{\rho,k}$, respectively. \\
\end{longtable}
\endgroup

\section{Algorithmic Details and Mathematical Derivations}
\label{app:algorithm}

\subsection{Benchmark and Fitted Simulation Model}
\label{app:benchmark-model}

\paragraph{Scope and the DSIS special case.}
\label{par:critic-scope}
We first specify DSIS-RB and its fitted simulation model; the critic
construction follows in Section~\ref{app:critic-details}.
DSIS is recovered as a special case of this framework by making regular
supply permanently available and replacing the demand process with its
original public i.i.d.\ distribution. Its previous-demand coordinate and
latent-risk variable can then be omitted. Thus, recovering DSIS requires
specializing both supply and demand; setting the RB risk to zero alone
does not recover its original demand process.

Throughout, $\rho$ denotes the actual trajectory-specific risk in the
environment, whereas $z$ is a realization of the synthetic risk variable
$Z$ drawn by the learning algorithm. Neither risk is an input to the
policy or value networks. Model-based computations use a declared
simulation distribution for $Z$, independently of the unobserved risk
of an observed trajectory. The nonnegative integer $g$ indexes learning
generations; model snapshot $g$ includes the real batch collected for
that generation. A synthetic tape is one simulated continuation with
its own random draws.

\subsubsection{Public supply-chain dynamics and rewards.}
\label{app:critic-physics}
\label{par:critic-physics}
At period $t\in\{0,1,\ldots\}$, let $s_t$ denote the observed inventory
configuration. Its net inventory is $I_t\in\mathbb Z$; its regular-order
pipeline is $\mathbf p_t=(p_{t,1},p_{t,2},p_{t,3},p_{t,4})$, where
$p_{t,\ell}\in\{0,\ldots,8\}$ is the outstanding quantity at stage
$\ell\in\{1,2,3,4\}$; and its current availability is
$B_t\in\{0,1\}$, with $0$ meaning unavailable and $1$ available.
Negative $I_t$ represents backlog. DSIS-RB additionally observes the
previous demand $d_{t-1}$. The six- and seven-dimensional observations are
\begin{equation}
  s_t^{\mathrm{DSIS}}=(I_t,\mathbf p_t,B_t),
  \qquad
  s_t^{\mathrm{RB}}=(I_t,\mathbf p_t,B_t,d_{t-1}).
  \label{eq:critic-observation}
\end{equation}
Hereafter, $s_t$ refers to the applicable observation above. We use $S_t$
for the corresponding random observation and $\mathsf D_t$ for random demand,
with realization $d_t$. Let $\mathcal A$ be the finite action set, and
let $q_t^r$ and $q_t^e$ denote the regular and expedited order quantities:
\begin{equation}
  a_t=(q_t^r,q_t^e)\in\mathcal A=\{0,\ldots,8\}^2.
  \label{eq:critic-action-space}
\end{equation}
There are $81$ actions. The implementation encodes $(q^r,q^e)$ by the
integer $9q^r+q^e\in\{0,\ldots,80\}$; the equations below use the pair.

Given realized demand $d_t$ and next-period availability $B_{t+1}$,
the public inventory and pipeline equations are
\begin{equation}
  \begin{aligned}
    I_{t+1}&=I_t+p_{t,1}+q_t^e-d_t,\\
    \mathbf p_{t+1}
    &=(p_{t,2},p_{t,3},p_{t,4},B_tq_t^r).
  \end{aligned}
  \label{eq:critic-inventory-pipeline}
\end{equation}
Define the deterministic RB observation-transition map $F$ by
\begin{equation}
    s_{t+1}
    =F(s_t,a_t,d_t,B_{t+1})=\bigl(I_{t+1},p_{t,2},p_{t,3},p_{t,4},
             B_tq_t^r,B_{t+1},d_t\bigr).
  \label{eq:critic-physical-transition}
\end{equation}
DSIS uses the same map with the last coordinate omitted and
$B_t=B_{t+1}=1$. Current availability $B_t$ determines whether the new
regular order $q_t^r$ is accepted. Previously accepted pipeline orders
$p_{t,1}$ still arrive during an outage. Expedited orders $q_t^e$ arrive
immediately, while accepted regular orders $B_tq_t^r$ enter the
four-period pipeline.

For a real number $x$, write $[x]_+=\max\{x,0\}$. Let $c(s,a,d)$ and
$r(s,a,d)$ denote the period cost and normalized reward functions,
respectively; their realized values are $c_t$ and $r_t$. They satisfy
\begin{equation}
  \begin{aligned}
    c_t&=c(s_t,a_t,d_t)
        =10q_t^e+5[I_{t+1}]_++495[-I_{t+1}]_+,\\
    r_t&=r(s_t,a_t,d_t)=-\frac{c_t}{495},
  \end{aligned}
  \label{eq:critic-reward}
\end{equation}
All values, Bellman residuals, and heuristic utilities below use these
normalized reward units. The critic discount factor is $\gamma=0.995$.

\subsubsection{The true benchmark transition law.}
\label{app:critic-benchmark-law}
\label{par:critic-true-model}
For the RB benchmark, let $\mathcal D_{\mathrm{RB}}$ denote the demand support:
\begin{equation}
  \mathcal D_{\mathrm{RB}}=\{0,1,2,3,8\}.
\end{equation}
The stationary demand probability row vector $\omega_D$ is
\begin{equation}
  \omega_D=(0.3,0.2,0.2,0.2,0.1)
          =\frac1{10}(3,2,2,2,1)
  \label{eq:critic-demand-stationary-law}
\end{equation}
Its entries follow the displayed order in $\mathcal D_{\mathrm{RB}}$,
and $\omega_D(d)$ denotes the entry associated with demand value $d$.
Thus $\omega_D(d)=\Pr(\mathsf D_t=d)$ under stationary initialization, where
$\Pr$ denotes probability. Let $\mathbf 1\{E\}$ equal one when statement
$E$ holds and zero otherwise. For $x,d\in\mathcal D_{\mathrm{RB}}$, define
the demand transition matrix $K_D$ by
\begin{equation}
  \Pr(\mathsf D_t=d\mid \mathsf D_{t-1}=x)
  =K_D(x,d)
  =0.75\,\mathbf 1\{d=x\}+0.25\,\omega_D(d).
  \label{eq:critic-true-demand}
\end{equation}
Thus, with probability $0.75$ demand retains its previous value;
otherwise it is independently redrawn from $\omega_D$.
This construction satisfies $\omega_DK_D=\omega_D$ and
$\sum_{d\in\mathcal D_{\mathrm{RB}}}d\,\omega_D(d)=2$.

An environment trajectory receives a risk $\rho\in[0,0.5]$, which
remains fixed throughout that trajectory. Conditional on $\rho$, let
$K_{B,\rho}(b_{\mathrm{cur}},u)=\Pr(B_{t+1}=u\mid B_t=b_{\mathrm{cur}},\rho)$ for
$b_{\mathrm{cur}},u\in\{0,1\}$. In row and column order $(0,1)$, this supply
transition matrix is
\begin{equation}
  K_{B,\rho}
  =
  \begin{pmatrix}
    \frac34 & \frac14\\[2pt]
    \frac{\rho}{4(1-\rho)}
    & 1-\frac{\rho}{4(1-\rho)}
  \end{pmatrix}.
  \label{eq:critic-true-supply}
\end{equation}
Its stationary probability row vector is $(\rho,1-\rho)$: the flows
$\rho/4$ from $0$ to $1$ and
$(1-\rho)\rho/[4(1-\rho)]$ from $1$ to $0$ agree.
Demand and supply innovations are independent conditional on the
current observation and $\rho$. Below we suppress time indices:
$s=s_t$, $a=a_t$, $s'=s_{t+1}$, and $u=B_{t+1}$.
Write $B(s)$ for the availability coordinate of $s$ and $d^-(s)$ for
its previous-demand coordinate. For any point $x$, $\delta_x$ denotes
the probability measure with unit mass at $x$; $dr$ and $ds'$ indicate
the reward and next-observation coordinates of a measure.
The joint reward--observation transition kernel $\mathcal P_\rho$ is
\begin{equation}
    \mathcal P_\rho(dr,ds'\mid s,a)=
    \sum_{d\in\mathcal D_{\mathrm{RB}}}
    \sum_{u\in\{0,1\}}
    K_D(d^-(s),d)K_{B,\rho}(B(s),u)\,
    \delta_{(r(s,a,d),F(s,a,d,u))}(dr,ds').
  \label{eq:critic-true-kernel}
\end{equation}

At each episode initialization, inventory and pipeline quantities are
zero. Writing $\operatorname{Bernoulli}(p)$ for a binary distribution
with probability $p$ of one, the initial demand and availability obey
\begin{equation}
  \mathsf D_{-1}\sim\omega_D,
  \qquad B_0\mid\rho\sim\Bernoulli(1-\rho),
  \label{eq:critic-true-initial-law}
\end{equation}
with independent demand and supply initialization conditional on $\rho$.

\emph{Risk stratification and its statistical target.}
The registered risk grid is $\rho_j=0.1j$, $j=0,\ldots,5$.
An episode contains $H_{\mathrm{env}}=1000$ physical periods. Let $\pi$
be a fixed policy assigning a probability $\pi(a\mid s)$ to each
$a\in\mathcal A$, and let $\mathbb E_{\pi,\rho_j}$ denote expectation
over its actions and the benchmark disturbances at risk $\rho_j$, with
the initialization above. Its expected episode mean cost is
\begin{equation}
 J_\pi(\rho_j)=\mathbb E_{\pi,\rho_j}\!\left[
 \frac{1}{H_{\mathrm{env}}}\sum_{t=0}^{H_{\mathrm{env}}-1}c_t\right].
 \label{eq:critic-risk-episode-cost}
\end{equation}
Define $\mathcal R_{\mathrm{trap}}(\pi)$ as the normalized trapezoidal
risk score and $\omega_{\rho,j}$ as its weight at grid point $j$;
$\omega_\rho$ collects these six weights in grid order:
\begin{equation}
  \begin{aligned}
    \mathcal R_{\mathrm{trap}}(\pi)
    &=\frac1{0.5}\sum_{j=0}^{4}\frac{0.1}{2}
          \bigl[J_\pi(\rho_j)+J_\pi(\rho_{j+1})\bigr]\\
    &=\sum_{j=0}^{5}\omega_{\rho,j}J_\pi(\rho_j),
    \qquad
    \omega_\rho=\frac1{10}(1,2,2,2,2,1).
  \end{aligned}
  \label{eq:critic-risk-trapezoid-target}
\end{equation}
The endpoints receive half the weight of an interior grid point
because each endpoint belongs to one trapezoid, whereas each interior
point belongs to two. These risk weights are distinct from the demand
probabilities $\omega_D$.

The probability measure $\nu_{\mathrm{env}}$ describing the target risk composition is
\begin{equation}
  \nu_{\mathrm{env}}
  =\sum_{j=0}^{5}\omega_{\rho,j}\delta_{\rho_j}.
  \label{eq:critic-environment-risk-law}
\end{equation}
Writing $\rho\sim\nu_{\mathrm{env}}$ specifies this target composition,
not independent sampling of every training trajectory. Training
implements proportional stratification in blocks of $25$ trajectories.
The desired counts are $25\omega_\rho=(2.5,5,5,5,5,2.5)$.
Let $\mathbf n^{(0)}$ and $\mathbf n^{(1)}$ denote the alternating
integer count vectors, with coordinates ordered by the risk grid:
\begin{equation}
  \begin{aligned}
    \mathbf n^{(0)}&=(3,5,5,5,5,2),\\
    \mathbf n^{(1)}&=(2,5,5,5,5,3),\\
    \mathbf n^{(0)}+\mathbf n^{(1)}
       &=(5,10,10,10,10,5)=50\omega_\rho.
  \end{aligned}
  \label{eq:critic-environment-risk-quotas}
\end{equation}
Each block is privately permuted before trajectories are supplied.
Alternating the two integer allocations avoids systematically rounding
one endpoint upward: every complete pair of blocks has exactly the
registered risk proportions, without random fluctuation in its stratum
counts.

For a policy held fixed across such a pair, let $Y_\ell$ be the realized
mean cost of trajectory $\ell\in\{1,\ldots,50\}$, and let $\mathbb E$
denote expectation over benchmark and policy randomness. Then
\begin{equation}
  \mathbb E\!\left[\frac1{50}\sum_{\ell=1}^{50}Y_\ell\right]
  =\sum_{j=0}^{5}\omega_{\rho,j}J_\pi(\rho_j)
  =\mathcal R_{\mathrm{trap}}(\pi).
  \label{eq:critic-fixed-policy-stratified-unbiasedness}
\end{equation}
This is unbiasedness for the registered discrete trapezoidal target.
For the actual changing training policies, the applicable guarantee is
exact two-block risk balance; it does not assert unbiasedness of a
training update for a single fixed-policy objective. Likewise, the
trapezoidal score is a quadrature approximation to a continuous-risk
integral, not an exactly unbiased estimator of that integral. Formal
evaluation holds the policy fixed, evaluates each stratum separately,
and combines its sample mean with $\omega_\rho$.

Within an episode, the augmented variable $X_t=(S_t,\rho)$ is a
controlled Markov state for the physical dynamics. The observation
$S_t$ alone need not be Markov after a persistent hidden risk is
marginalized. The displayed physical kernel excludes the scheduled
reset after $H_{\mathrm{env}}$ periods.
The preceding probabilities specify the benchmark for reproducibility;
the RB learner does not receive its true stochastic kernels, trajectory
risks, risk assignments, or private random tapes.

For DSIS, the same physical and reward equations apply on the reachable
set $B_t=B_{t+1}=1$. Here $\operatorname{Unif}$ denotes the uniform
distribution, and demand is independent and identically distributed (i.i.d.):
\begin{equation}
  \mathsf D_t\overset{\mathrm{iid}}{\sim}\operatorname{Unif}\{0,1,2,3,4\}.
  \label{eq:critic-dsis-demand}
\end{equation}
Its observation is the six-dimensional vector
\begin{equation}
  s_t=(I_t,\mathbf p_t,B_t).
  \label{eq:critic-dsis-observation}
\end{equation}

\subsubsection{Constructing the fitted simulation model.}
\label{app:critic-model-construction}
\label{par:critic-structural-prior}
\label{par:critic-fitted-demand}
For DSIS-RB, the generation-$g$ model is fitted only to counted real observations.
We define its components before summarizing all structural and
estimation assumptions in Table~\ref{tab:critic-structural-priors}.

\emph{Observed-data demand model.}
Let $\mathscr D_{\le g}^{\mathrm{real}}$ be the multiset of counted,
observed transitions available after collecting the generation-$g$
batch, including any counted initialization prefix. Every observed
occurrence is counted once, including repeated transitions. Set
$N_g^{\mathrm{real}}=|\mathscr D_{\le g}^{\mathrm{real}}|>0$;
no fitted RB model is formed without real data. Recall that $d^-(s)$
selects the previous-demand coordinate. Let $\widehat{\mathcal D}_g$
be the union of the values of $d^-(s)$ and $d^-(s')$ in this multiset.
For $x,y\in\widehat{\mathcal D}_g$, define the pair count
$\mathcal C_g(x,y)$, its row total $\mathcal C_g(x)$, and the total
count as follows:
\begin{equation}
  \mathcal C_g(x,y)
  =
  \sum_{(s,a,r,s')\in\mathscr D_{\le g}^{\mathrm{real}}}
  \mathbf 1\{d^-(s)=x,\ d^-(s')=y\},
  \label{eq:critic-demand-counts}
\end{equation}
The row and total counts satisfy
\begin{equation}
  \mathcal C_g(x)=\sum_{y\in\widehat{\mathcal D}_g}\mathcal C_g(x,y),
  \qquad
  N_g^{\mathrm{real}}=\sum_{x,y\in\widehat{\mathcal D}_g}\mathcal C_g(x,y).
  \label{eq:critic-demand-row-counts}
\end{equation}
The pooled empirical next-demand probability mass function
$\widehat\omega_g$ is
\begin{equation}
  \widehat\omega_g(y)
  =\frac{\sum_{x\in\widehat{\mathcal D}_g}\mathcal C_g(x,y)}{N_g^{\mathrm{real}}}.
  \label{eq:critic-pooled-demand}
\end{equation}
With pseudocount mass $\alpha_D=1$, define the fitted demand transition
matrix $\widehat K_{D,g}$ on $\widehat{\mathcal D}_g$ by
\begin{equation}
  \widehat K_{D,g}(x,y)
  =\frac{\mathcal C_g(x,y)+\alpha_D\widehat\omega_g(y)}
         {\mathcal C_g(x)+\alpha_D},
  \qquad \alpha_D=1.
  \label{eq:critic-fitted-demand}
\end{equation}
Each conditional row is smoothed toward the pooled empirical
distribution using one total pseudocount. Its entries are nonnegative
and sum to one. If $\mathcal C_g(x)=0$, the row equals
$\widehat\omega_g$; this includes a value observed only as a successor.
Queries require $d^-(s)\in\widehat{\mathcal D}_g$; the implementation
rejects an unseen conditioning value rather than inventing a row.

\emph{Supply model and persistent synthetic risk.}
\phantomsection
\label{par:critic-fitted-supply}
Recall that $B(s)$ is the availability coordinate. Let
$\mathcal C_g^\downarrow$ count observed transitions starting in an
outage and $\mathcal C_g^{\mathrm{rec}}$ count those that recover:
\begin{equation}
  \begin{aligned}
    \mathcal C_g^\downarrow
    &=\sum_{(s,a,r,s')\in\mathscr D_{\le g}^{\mathrm{real}}}
      \mathbf 1\{B(s)=0\},\\
    \mathcal C_g^{\mathrm{rec}}
    &=\sum_{(s,a,r,s')\in\mathscr D_{\le g}^{\mathrm{real}}}
      \mathbf 1\{B(s)=0,B(s')=1\}.
  \end{aligned}
\end{equation}
Let $\kappa^{\mathrm{rec}}\in(0,1)$ be the common recovery-probability parameter in
the fitted supply family, and let $\widehat\kappa_g^{\mathrm{rec}}$ be its estimate:
\begin{equation}
  \widehat\kappa_g^{\mathrm{rec}}
  =\frac{\mathcal C_g^{\mathrm{rec}}+1}{\mathcal C_g^\downarrow+2}.
  \label{eq:critic-fitted-recovery}
\end{equation}
Under a $\operatorname{Beta}(1,1)$ prior (the uniform prior on
$(0,1)$) for $\kappa^{\mathrm{rec}}$, this is the posterior mean based on the
outage-origin transitions. It remains defined when no outage has been
observed, in which case $\widehat\kappa_g^{\mathrm{rec}}=1/2$.
Let $m$ denote the synthetic-risk distribution. At the root of each
new synthetic tape, the algorithm draws $Z$ independently of the queried
observation and of the other tapes:
\begin{equation}
  Z\sim m=\operatorname{Unif}[0,0.5].
  \label{eq:critic-synthetic-risk}
\end{equation}
The tape retains this draw throughout its evolution. Conditional on
$Z=z$, let $\widehat K_{B,g,z}$ be the modeled supply transition
matrix, with rows and columns ordered as $(0,1)$:
\begin{equation}
  \widehat K_{B,g,z}
  =
  \begin{pmatrix}
    1-\widehat\kappa_g^{\mathrm{rec}} & \widehat\kappa_g^{\mathrm{rec}}\\[2pt]
    \widehat\kappa_g^{\mathrm{rec}}\frac{z}{1-z}
    & 1-\widehat\kappa_g^{\mathrm{rec}}\frac{z}{1-z}
  \end{pmatrix}.
  \label{eq:critic-fitted-supply}
\end{equation}
Because $0<\widehat\kappa_g^{\mathrm{rec}}<1$ and $0\le z/(1-z)\le1$, all
matrix entries lie in $[0,1]$ and each row sums to one. The identity
\begin{equation}
  (1-z)\widehat K_{B,g,z}(1,0)
  =z\widehat K_{B,g,z}(0,1)
\end{equation}
shows that $(z,1-z)$ is a stationary probability row vector, so the
stationary outage probability is $z$, including the boundary $z=0$.
This parameterization is a modeling assumption; knowing only the risk
interval would not imply it.

\emph{The combined model and generation-level freezing.}
Recall that $r$ and $F$ are the public reward and physical-transition
maps in Equations~\eqref{eq:critic-reward} and
\eqref{eq:critic-physical-transition}. For a query with
$d^-(s)\in\widehat{\mathcal D}_g$ and $z\in[0,0.5]$, define the
conditional reward--observation kernel $\widehat{\mathcal P}_{g,z}$ by
\begin{equation}
  \begin{aligned}
    &\widehat{\mathcal P}_{g,z}(dr,ds'\mid s,a)\\
    &\quad=
    \sum_{d\in\widehat{\mathcal D}_g}
    \sum_{u\in\{0,1\}}
    \widehat K_{D,g}(d^-(s),d)
    \widehat K_{B,g,z}(B(s),u)\,
    \delta_{(r(s,a,d),F(s,a,d,u))}(dr,ds').
  \end{aligned}
  \label{eq:critic-fitted-kernel}
\end{equation}
Here $\delta_x$ again denotes unit probability mass at $x$.
We write $\widehat{\mathcal P}_g$ for the full simulation specification:
a query observation, an independent initial risk draw from $m$, and
the conditional dynamics above with that risk retained. To state its
Markov transition explicitly, let $z'$ be the successor risk, let $dz'$
indicate its measure coordinate, and define the augmented kernel
$\widetilde{\mathcal P}_g$ on observation--risk pairs by
\begin{equation}
  \widetilde{\mathcal P}_g(dr,ds',dz'\mid s,z,a)
  =\widehat{\mathcal P}_{g,z}(dr,ds'\mid s,a)\delta_z(dz').
  \label{eq:critic-augmented-model}
\end{equation}
The model snapshot is frozen before generation-$g$ model-based
computations begin. For DSIS, $\widehat{\mathcal P}_{g,z}$ reduces to
the public nominal kernel independently of $g$ and $z$; no
stochastic-law estimation or latent-risk simulation is required.

\emph{Summary of assumptions.}
Table~\ref{tab:critic-structural-priors} lists all structural and
estimation assumptions; HBPO uses every applicable row. The last
column reports explicit baseline use, separately from access to public
information. In the table, $\widehat p_g$ denotes the conditional joint
probability mass function of modeled demand $d$ and next availability
$u$, with its one-variable versions denoting the corresponding marginals.
For a synthetic tape, $k\in\{0,1,\ldots\}$ indexes simulated periods,
$S_0$ is its query observation, and $Z_k$ is its retained risk at period
$k$; $\perp$ means independence. In rows marked Both, DSIS omits the
unused previous-demand and risk arguments: its demand is uniform on
$\{0,\ldots,4\}$ and its next availability equals one.

\begingroup
\footnotesize
\setlength{\tabcolsep}{3pt}
\renewcommand{\arraystretch}{1.25}
\setlength{\priorTableWidth}{\dimexpr\linewidth-8\tabcolsep\relax}
\begin{longtable}{@{}
  >{\raggedright\arraybackslash}p{0.33\priorTableWidth}
  >{\raggedright\arraybackslash}p{0.16\priorTableWidth}
  >{\raggedright\arraybackslash}p{0.23\priorTableWidth}
  >{\raggedright\arraybackslash}p{0.10\priorTableWidth}
  >{\raggedright\arraybackslash}p{0.18\priorTableWidth}@{}}
\caption{Declared structural and estimation priors. Both means DSIS
and DSIS-RB. Baseline use refers to explicit use in the learner;
all methods interact with the same physical benchmark.}
\label{tab:critic-structural-priors}\\
\toprule
Formula & Description & Physical or modeling role & DSIS / DSIS-RB & Baseline use \\
\midrule
\endfirsthead
\multicolumn{5}{@{}l}{\tablename~\thetable\ (continued)}\\
\toprule
Formula & Description & Physical or modeling role & DSIS / DSIS-RB & Baseline use \\
\midrule
\endhead
\midrule
\multicolumn{5}{r@{}}{Continued on next page}\\
\endfoot
\bottomrule
\endlastfoot
\multicolumn{5}{@{}l}{\emph{Physical rules and stochastic model structure}}\\
\addlinespace[2pt]
$\begin{gathered}
 a_t=(q_t^r,q_t^e),\\
 q_t^r,q_t^e\in\{0,\ldots,8\}
\end{gathered}$
& Finite action domain.
& Two ordering channels, each accepting quantities from $0$ to $8$.
& Both
& All: action interface; GPS: model branches. \\
\addlinespace[3pt]
$I_{t+1}=I_t+p_{t,1}+q_t^e-d_t$
& Deterministic inventory update.
& Inventory conservation, scheduled arrivals, and immediate, reliable expediting.
& Both
& GPS: explicit state model. \\
\addlinespace[3pt]
$\begin{gathered}
 p_{t+1,\ell}=p_{t,\ell+1},\quad \ell=1,2,3,\\
 p_{t+1,4}=B_tq_t^r
\end{gathered}$
& Pipeline shift and order gate.
& Four-period regular lead time; outages reject new regular orders without cancelling existing orders.
& Both
& GPS: explicit state model. \\
\addlinespace[3pt]
$\begin{gathered}
 c_t=10q_t^e+5[I_{t+1}]_+\\
 \phantom{c_t={}}+495[-I_{t+1}]_+,\\
 r_t=-c_t/495
\end{gathered}$
& Known cost and reward map.
& Expediting, holding, and backlog costs; a common reward scale.
& Both
& GPS: model rewards; all: observed rewards. \\
\addlinespace[3pt]
$\begin{gathered}
 \mathsf D_t\overset{\mathrm{iid}}{\sim}\Unif\{0,\ldots,4\},\\
 B_t=B_{t+1}=1
\end{gathered}$
& Public nominal stochastic law.
& Memoryless demand and uninterrupted regular supply.
& DSIS
& GPS: nominal model. \\
\addlinespace[3pt]
$\begin{gathered}
 \widehat p_g(d\mid s_t,a_t,z)\\
 {}=\widehat K_{D,g}(d^-(s_t),d),\\
 \widehat p_g(u\mid s_t,a_t,z)\\
 {}=\widehat K_{B,g,z}(B_t,u)
\end{gathered}$
& First-order, exogenous disturbance models.
& Previous demand and current availability carry temporal dependence; orders do not control these laws.
& DSIS-RB
& GPS: same conditional model structure. \\
\addlinespace[3pt]
$\begin{gathered}
 \widehat p_g(d,u\mid s_t,a_t,z)\\
 {}=\widehat K_{D,g}(d^-(s_t),d)\\
 \quad\cdot\widehat K_{B,g,z}(B_t,u)
\end{gathered}$
& Conditional factorization.
& Separate demand and supply innovations; DSIS has i.i.d.\ demand and degenerate supply.
& Both
& GPS: factorized disturbance model. \\
\addlinespace[3pt]
$\begin{gathered}
 \widehat K_{B,g,z}(0,1)=\widehat\kappa_g^{\mathrm{rec}},\\
 \widehat K_{B,g,z}(1,0)=\widehat\kappa_g^{\mathrm{rec}}\frac{z}{1-z}
\end{gathered}$
& Common recovery and stationary-risk family.
& Binary interruptions with modeled stationary outage share $z$.
& DSIS-RB
& GPS: same supply family. \\
\addlinespace[3pt]
$\begin{gathered}
 Z_0\sim\Unif[0,0.5],\quad Z_0\perp S_0,\\
 Z_k=Z_0\quad\text{for all }k
\end{gathered}$
& Independent query prior; persistent tape risk.
& Models sustained risk heterogeneity on the public interval; not the actual risk-assignment law.
& DSIS-RB
& GPS: same synthetic-risk prior. \\
\midrule
\multicolumn{5}{@{}l}{\emph{Estimation priors (regularization, not benchmark laws)}}\\
\addlinespace[2pt]
$\begin{gathered}
 \widehat K_{D,g}(x,y)\\
 {}=\dfrac{\mathcal C_g(x,y)+\widehat\omega_g(y)}
              {\mathcal C_g(x)+1}
\end{gathered}$
& One pooled pseudocount per row.
& Regularizes demand transitions with sparse observed counts.
& DSIS-RB
& GPS: same demand estimator. \\
\addlinespace[3pt]
$\begin{gathered}
 \kappa^{\mathrm{rec}}\sim\operatorname{Beta}(1,1),\\
 \widehat\kappa_g^{\mathrm{rec}}=\dfrac{\mathcal C_g^{\mathrm{rec}}+1}
                        {\mathcal C_g^\downarrow+2}
\end{gathered}$
& Smoothed recovery estimate.
& Regularizes recovery counts from observed outages.
& DSIS-RB
& GPS: same recovery estimator. \\
\end{longtable}
\endgroup

\emph{Reading the baseline column.}
All refers to PPO, A2C, Discrete SAC, ACE, PPO-81, TRPO, GRPO, TPO,
and GPS. Every method has access to the same public inventory rules
and cost function, and to the complete nominal stochastic law in DSIS.
The declared model-class assumptions are available to all comparison
methods. An entry naming GPS identifies explicit use in its predictive
model, rather than exclusive access to the corresponding information.
The other baselines use their native learning rules on observed
transitions; receiving a reward does not itself constitute evaluating
the analytical reward map on a hypothetical transition. GPS and HBPO
fit unknown RB probabilities to their respective counted data, so
sharing an estimator does not imply sharing fitted parameters.

\emph{Information boundary.}
The functional forms in the table are specified in advance, rather
than learned solely from data. In particular, the supply parameterization
$\widehat\kappa_g^{\mathrm{rec}} z/(1-z)$ does not follow from knowing the risk interval
$[0,0.5]$ alone: it also assumes a common recovery probability and a
stationary outage share $z$. These assumptions do not give the learner
the true RB recovery probability $0.25$, the true demand transition
matrix, the actual trajectory risk $\rho$, or future random tapes.
The RB demand support is discovered from observations, and unknown
probabilities are estimated only from the learner's own counted
observations. The synthetic distribution of $Z$ is neither the private
benchmark risk-assignment law nor a posterior conditioned on the queried
state. The DSIS fallback separately grants every method its complete
public nominal law.

\subsubsection{Use of structural priors and the information-fair comparison.}
\label{app:critic-prior-comparison}
\label{par:critic-prior-use}
The priors in Table~\ref{tab:critic-structural-priors} enter HBPO through
initialization, including fitted Q evaluation (FQE) on counterfactuals
using observed disturbances; supply-conditioned replay; action-value
backups and residual continuations; successor-critic state generation
and fitting; and heuristic utility, occupancy, calibration, and
confidence/interest calculations. These computations use simulated
transitions and conditional expectations for learning, with enumeration
over finite action and disturbance branches rather than the inventory
state space. They provide no additional real-environment observations.

All methods share public information and a $2{,}000{,}000$
real-interaction budget per formal learning run, with synthetic
computation recorded separately. GPS uses the same model class,
estimation rules, and tape-fixed uniform synthetic-risk prior as HBPO,
fitted to its own observations. Neither learner receives unrestricted
environment resets. Fairness therefore concerns information access and
real interactions, not equal computation or identical fitted models.

The structural prior is part of HBPO. Comparison with model-informed
GPS assesses the overall learning architecture beyond shared model
access; it does not isolate an individual transmission component.
Mechanism analyses and ablations examine guidance and transmission
under common modeling and estimation rules, while their collected data
and fitted probabilities may differ. Using models and inventory
structure to support policy learning has established precedents
\citep{sutton1991dyna,levine2013gps,boettcher2023dual}.

\subsection{Critic Construction and Model-Based Residual Completion}
\label{app:critic-details}

\subsubsection{The five scalar critics and the frozen value reference.}
\label{app:critic-reference}
\label{par:critic-ensemble}
At generation $g$, HBPO maintains five separately parameterized scalar
networks. For member index $j\in\{1,\ldots,5\}$, let $\theta_{g,j}$
denote its trainable parameters and write
\begin{equation}
  V_{g,j}(s)=V_{\theta_{g,j}}(s).
\end{equation}
The frozen ensemble reference is the mean of their scalar outputs,
\begin{equation}
  v_g(s)=\frac15\sum_{j=1}^{5}V_{g,j}(s).
  \label{eq:critic-ensemble-mean}
\end{equation}
The networks have no shared trainable layers.

Let $d_\phi$ denote the feature dimension: $d_\phi=7$ in RB and
$d_\phi=6$ in DSIS. Recall that $I$ is inventory, $p_1,\ldots,p_4$
are pipeline coordinates, $B$ is current availability, and $d^-$ is
previous demand. The fixed RB feature map, written as a column vector, is
\begin{equation}
  \phi(s)=
  \left(
    \operatorname{asinh}(I/4),
    p_1/8,p_2/8,p_3/8,p_4/8,
    B,d^-/8
  \right)^\top\in\mathbb R^7,
  \label{eq:critic-features}
\end{equation}
where $\operatorname{asinh}$ is the inverse hyperbolic sine. DSIS omits
the last coordinate.

Each network has two Tanh hidden layers of width $128$ and a scalar
linear output. For a generic parameter vector
$\theta=(W_1,b_1,W_2,b_2,w_3,b_3)$, the dimensions are
$W_1\in\mathbb R^{128\times d_\phi}$,
$W_2\in\mathbb R^{128\times128}$,
$b_1,b_2,w_3\in\mathbb R^{128}$, and $b_3\in\mathbb R$; $\tanh$
acts coordinatewise. The resulting scalar function is
\begin{equation}
  V_\theta(s)
  =w_3^\top\tanh\!\left(
    W_2\tanh(W_1\phi(s)+b_1)+b_2
  \right)+b_3.
  \label{eq:critic-network}
\end{equation}

During generation $g$, $v_g$ is a frozen continuation-value reference.
It need not equal the exact value under either the current model or the
environment. For $g\ge1$, its parameters are inherited from the preceding
generation even when the newly fitted model has changed; $v_0$ is supplied
by the initialization procedure described below.

\subsubsection{One-step action backups and the target-policy residual.}
\label{app:critic-backups}
\label{par:critic-backup}
Recall that $\widehat{\mathcal P}_{g,z}$ is the frozen model conditional
on synthetic risk $z$, $\widehat{\mathcal D}_g$ is its finite demand
support, and $\widehat K_{D,g}$ and $\widehat K_{B,g,z}$ are its demand
and availability kernels. Also recall the deterministic successor map
$F$, normalized reward $r$, and discount $\gamma=0.995$.
Here $d^-(s)$ and $B(s)$ denote the previous-demand and current-availability
coordinates of $s$. For a real-valued function $V$ with finite backups
and a prescribed next availability $u\in\{0,1\}$, define the supply-conditioned action backup
by integrating demand:
\begin{equation}
  L_g^uV(s,a)
  =\sum_{d\in\widehat{\mathcal D}_g}
  \widehat K_{D,g}(d^-(s),d)
  \left[r(s,a,d)+\gamma V(F(s,a,d,u))\right].
  \label{eq:critic-supply-conditioned-backup}
\end{equation}
Integrating next availability conditional on the fixed tape risk $z$ gives
\begin{equation}
  L_{g,z}V(s,a)
  =\sum_{u\in\{0,1\}}
  \widehat K_{B,g,z}(B(s),u)L_g^uV(s,a).
  \label{eq:critic-risk-conditioned-backup}
\end{equation}
Let $R$ and $S'$ denote the reward and successor from one transition of
this model. Then
\begin{equation}
  L_{g,z}V(s,a)
  =\E_{\widehat{\mathcal P}_{g,z}}
  [R+\gamma V(S')\mid s,a].
\end{equation}

The frozen parent target policy $\pi_g$ induces the discounted Bellman
operator $\mathcal T_{g,z}^{\pi_g}$,
\begin{equation}
  \mathcal T_{g,z}^{\pi_g}V(s)
  =\sum_{a\in\mathcal A}\pi_g(a\mid s)L_{g,z}V(s,a).
  \label{eq:critic-bellman-operator}
\end{equation}
Its residual at the frozen reference $v_g$ is
\begin{equation}
  \begin{aligned}
    e_{g,z}(s)
    &=(\mathcal T_{g,z}^{\pi_g}v_g)(s)-v_g(s)\\
    &=\sum_{a'\in\mathcal A}\pi_g(a'\mid s)
      \sum_{d\in\widehat{\mathcal D}_g}\sum_{u\in\{0,1\}}
      \widehat K_{D,g}(d^-(s),d)\widehat K_{B,g,z}(B(s),u)\\
    &\qquad\qquad \cdot\bigl[r(s,a',d)+\gamma v_g(F(s,a',d,u))\bigr]-v_g(s).
  \end{aligned}
  \label{eq:app-critic-backup-residual}
\end{equation}
This residual integrates the target action, demand, and next availability
conditional on $z$. It is the conditional mean of a temporal-difference
error, rather than a sampled reward increment. Its sign indicates whether
the model's one-step evaluation exceeds or falls below the reference
value; it does not indicate that the policy itself has improved.
Theorem~\ref{thm:critic-residual-completion} shows why propagating this
signed inconsistency recovers the remaining continuation-value error.

Recall the synthetic-risk prior $m=\operatorname{Unif}[0,1/2]$ in RB.
The risk-averaged one-step base is
\begin{equation}
  Q_g^0(s,a)=\E_{Z\sim m}[L_{g,Z}v_g(s,a)].
  \label{eq:critic-dense-base}
\end{equation}
Let $\widehat\kappa_g^{\mathrm{rec}}$ denote the fitted recovery probability, as in the
supply model above. Since $v_g$ has no risk input, the RB root backup
uses the analytic mean supply kernel
\begin{equation}
  \E_{Z\sim m}[\widehat K_{B,g,Z}(B,0)]
  =\begin{cases}
    1-\widehat\kappa_g^{\mathrm{rec}},&B=0,\\[2pt]
    \widehat\kappa_g^{\mathrm{rec}}(2\log 2-1),&B=1,
  \end{cases}
  \label{eq:critic-prior-mean-supply}
\end{equation}
where $\log$ is the natural logarithm and
$\E_{Z\sim m}[Z/(1-Z)]=2\int_0^{1/2}z/(1-z)\,dz=2\log2-1$.
Only the one-step root backup is averaged in this way. Multistep
simulation retains the sampled $z$; replacing its kernel at every step
by the prior mean would in general define a different process.

\subsubsection{Residual completion of the discounted action value.}
\label{app:critic-completion}
\label{app:residual-derivation}
\label{par:critic-residual-completion}
Recall that $m$ is the synthetic-risk law and that
$\widehat{\mathcal P}_{g,z}$ is the frozen reward--observation kernel
at generation $g$ (Equation~\eqref{eq:critic-fitted-kernel}).
Let $S_k$, $\mathsf A_k$, and $R_k$ denote the observation, action,
and normalized reward at simulated step $k$. Define the model
\emph{action-value function} queried by HBPO as
\begin{equation}
  Q_g^{\pi_g}(s,a)
  =
  \E_{Z\sim m}
  \E_{\widehat{\mathcal P}_{g,Z},\pi_g}
  \left[
    \sum_{k=0}^{\infty}\gamma^kR_k
    \,\middle|\,
    S_0=s,\mathsf A_0=a
  \right],
  \label{eq:critic-model-action-value}
\end{equation}
where the first action is fixed to $a$, and
$\mathsf A_k\sim\pi_g(\cdot\mid S_k)$ for $k\ge1$.
The risk is sampled independently at the query root and remains fixed
thereafter. This defines a model action-value function under the
declared risk prior; $z$ is a simulated scenario, not an estimate of
the queried trajectory's true risk $\rho$.

For the following derivation, fix generation $g$ and write
$\E_z^{s,a}$ for expectation under the rollout law with fixed risk $z$, initial
observation $s$, and forced first action $a$.
Write $\E_z^s$ when the initial action also follows $\pi_g$.
Define the exact conditional model value $V_{g,z}^{\pi_g}$ and
its error $\Delta_{g,z}$ relative to $v_g$ by
\begin{equation}
  V_{g,z}^{\pi_g}(s)=\E_z^s\!\left[\sum_{k=0}^{\infty}\gamma^kR_k\right],
  \qquad
  \Delta_{g,z}(s)=V_{g,z}^{\pi_g}(s)-v_g(s).
  \label{eq:critic-conditional-value-error}
\end{equation}
Here $V_{g,z}^{\pi_g}$ is the exact conditional \emph{model} value,
whereas $v_g$ is the frozen network reference. For a scalar function $f$,
let the model's one-step state-transition operators be
$\widehat P_{g,z}^{a}$ (forced first action $a$) and
$\widehat P_{g,z}^{\pi_g}$ (first action sampled from $\pi_g$):
\begin{equation}
  (\widehat P_{g,z}^{a}f)(s)=\E_z^{s,a}[f(S_1)],
  \qquad
  (\widehat P_{g,z}^{\pi_g}f)(s)
  =\sum_{a'\in\mathcal A}\pi_g(a'\mid s)(\widehat P_{g,z}^{a'}f)(s).
  \label{eq:critic-state-transition-operators}
\end{equation}

\begin{theorem}[Residual completion and the omitted continuation error]
\label{thm:critic-residual-completion}
Let $\mathcal S$ be the countable observation space of the frozen
inventory model, closed under its transitions. Fix the model, the
observation-based policy $\pi_g$, a discount $0<\gamma<1$, and a
reference $v_g$ satisfying $|v_g(s)|\le C_v<\infty$ for all
$s\in\mathcal S$, where $C_v$ is a constant. Assume that, for every
$s\in\mathcal S$ and $a\in\mathcal A$,
\begin{equation}
  \int \E_z^{s,a}\!\left[
    \sum_{k=0}^{\infty}\gamma^k|R_k|\right]m(dz)<\infty.
  \label{eq:critic-reward-integrability}
\end{equation}
The action set is finite, so this also implies the corresponding
condition when the first action follows $\pi_g$.
Recall that $\mathcal T_{g,z}^{\pi_g}$ is the Bellman operator
in Equation~\eqref{eq:critic-bellman-operator}, and that
$e_{g,z}=\mathcal T_{g,z}^{\pi_g}v_g-v_g$.
Write $\mathrm{Id}$ for the identity operator on scalar functions.
For an integer $\ell\ge0$, $(\widehat P_{g,z}^{\pi_g})^\ell$
denotes $\ell$-fold operator composition, with zeroth power
$\mathrm{Id}$.
Then the Bellman residual and conditional value error satisfy
\begin{equation}
  e_{g,z}=(\mathrm{Id}-\gamma\widehat P_{g,z}^{\pi_g})\Delta_{g,z},
  \qquad
  \Delta_{g,z}
  =\sum_{\ell=0}^{\infty}\gamma^\ell
    (\widehat P_{g,z}^{\pi_g})^\ell e_{g,z}.
  \label{eq:critic-error-propagation}
\end{equation}
These identities hold at every $s\in\mathcal S$ on a common set
of risks of full $m$-measure. The series converges absolutely
pointwise; no bounded-value-space inverse is assumed. In particular, the exact
error of the one-step base is
\begin{equation}
  Q_g^{\pi_g}(s,a)-Q_g^0(s,a)
  =\gamma\E_{Z\sim m}[(\widehat P_{g,Z}^{a}\Delta_{g,Z})(s)].
  \label{eq:critic-omitted-continuation-error}
\end{equation}
It is recovered by the discounted residual sum:
\begin{equation}
  Q_g^{\pi_g}(s,a)
  =Q_g^0(s,a)
   +\int\E_z^{s,a}\!\left[
      \sum_{k=1}^{\infty}\gamma^ke_{g,z}(S_k)
    \right]m(dz).
  \label{eq:app-critic-residual-completion}
\end{equation}
\end{theorem}

\begin{proof}
Because $\mathcal S$ is countable and $\mathcal A$ is finite,
Equation~\eqref{eq:critic-reward-integrability} provides a common
set of risks of full $m$-measure on which absolute discounted
reward integrability holds for every initial state and action.
Fix a risk $z$ in this set. Conditional on $z$, the observation
process is Markov. Bellman consistency of the exact conditional
value gives
\begin{equation}
  \Delta_{g,z}
  =e_{g,z}+\gamma\widehat P_{g,z}^{\pi_g}\Delta_{g,z}.
  \label{eq:critic-error-recursion}
\end{equation}
For any integer $n_{\mathrm{tr}}\ge0$, iterating this relation yields
\begin{equation}
  \Delta_{g,z}(s)
  =\sum_{\ell=0}^{n_{\mathrm{tr}}}\gamma^\ell
    ((\widehat P_{g,z}^{\pi_g})^\ell e_{g,z})(s)
   +\gamma^{n_{\mathrm{tr}}+1}\E_z^s[\Delta_{g,z}(S_{n_{\mathrm{tr}}+1})].
  \label{eq:critic-error-finite-expansion}
\end{equation}
The absolute terminal term is bounded by
\begin{equation}
  \gamma^{n_{\mathrm{tr}}+1}\E_z^s[|\Delta_{g,z}(S_{n_{\mathrm{tr}}+1})|]
  \le \E_z^s\!\left[\sum_{k=n_{\mathrm{tr}}+1}^{\infty}\gamma^k|R_k|\right]
      +\gamma^{n_{\mathrm{tr}}+1}C_v \longrightarrow 0.
  \label{eq:critic-error-tail-bound}
\end{equation}
Here the reward-tail bound follows from the Markov property and
conditional Jensen's inequality. Absolute integrability makes its
expectation tend to zero. Moreover, conditional Jensen's inequality
applied to the one-step residual gives
\begin{equation}
  \sum_{\ell=0}^{\infty}\gamma^\ell
    \E_z^s[|e_{g,z}(S_\ell)|]
  \le \E_z^s\!\left[\sum_{\ell=0}^{\infty}\gamma^\ell|R_\ell|\right]
      +\frac{(1+\gamma)C_v}{1-\gamma}<\infty.
  \label{eq:critic-absolute-residual-bound}
\end{equation}
This proves both identities in
Equation~\eqref{eq:critic-error-propagation}, including absolute
convergence of its series. The same bounds can be integrated against
$m(dz)$ by the stated mixture integrability assumption.

Conditioning the discounted return on its first transition yields
\begin{equation}
  Q_g^{\pi_g}(s,a)
  =\int\left\{L_{g,z}v_g(s,a)
       +\gamma(\widehat P_{g,z}^{a}\Delta_{g,z})(s)\right\}m(dz).
  \label{eq:critic-first-step-decomposition}
\end{equation}
Subtracting $Q_g^0(s,a)=\int L_{g,z}v_g(s,a)m(dz)$ proves
Equation~\eqref{eq:critic-omitted-continuation-error}.

For the residual-completion identity, at every continuation step
$k\ge1$ the tower property gives
\begin{equation}
  \E_z^{s,a}[e_{g,z}(S_k)]
  =\E_z^{s,a}[R_k+\gamma v_g(S_{k+1})-v_g(S_k)].
  \label{eq:critic-expected-td-error}
\end{equation}
The restriction $k\ge1$ ensures that the action follows $\pi_g$;
the forced root action is handled by $L_{g,z}v_g(s,a)$.
Multiplying by $\gamma^k$ and summing gives, for each integer $n_{\mathrm{tr}}\ge0$,
\begin{equation}
    L_{g,z}v_g(s,a)
     +\E_z^{s,a}\!\left[\sum_{k=1}^{n_{\mathrm{tr}}}\gamma^k e_{g,z}(S_k)\right]=\E_z^{s,a}\!\left[
       \sum_{k=0}^{n_{\mathrm{tr}}}\gamma^k R_k+\gamma^{n_{\mathrm{tr}}+1}v_g(S_{n_{\mathrm{tr}}+1})\right].
  \label{eq:critic-finite-telescope}
\end{equation}
The sum is empty when $n_{\mathrm{tr}}=0$. After integration over $m(dz)$, the
remaining reference-value term satisfies
\begin{equation}
  \gamma^{n_{\mathrm{tr}}+1}\int\E_z^{s,a}[|v_g(S_{n_{\mathrm{tr}}+1})|]m(dz)
  \le\gamma^{n_{\mathrm{tr}}+1}C_v\longrightarrow0.
  \label{eq:critic-vanishing-tail}
\end{equation}
For the forced-action rollout, the analogue of
Equation~\eqref{eq:critic-absolute-residual-bound} starts at $k=1$;
its right-hand side remains integrable over the root risk.
Consequently the infinite residual sum and its expectation can be
interchanged, and taking $n_{\mathrm{tr}}\to\infty$ proves
Equation~\eqref{eq:app-critic-residual-completion}.
\end{proof}

The theorem identifies the correction exactly: it is the part of
future policy value missing from the one-step bootstrap. The factor
$\gamma^k$ is imposed by the original discounted return and the
telescoping cancellation; it is not a separately chosen weight for
risk or optimism. At a finite deterministic continuation length $n_{\mathrm{tr}}$,
the remaining error is exactly
\begin{equation}
    Q_g^{\pi_g}(s,a)
    -\left\{Q_g^0(s,a)+\int\E_z^{s,a}\!\left[
        \sum_{k=1}^{n_{\mathrm{tr}}}\gamma^ke_{g,z}(S_k)\right]m(dz)\right\} =\gamma^{n_{\mathrm{tr}}+1}\int\E_z^{s,a}[\Delta_{g,z}(S_{n_{\mathrm{tr}}+1})]m(dz).
  \label{eq:critic-truncation-error}
\end{equation}
Thus a fixed truncation retains a terminal evaluation error.
\hyperref[par:critic-geometric-estimator]{The geometrically stopped estimator below} implements the complete
series with an almost surely finite random continuation.

\emph{Persistent risk and an observation-only critic.}
RB introduces a further reason for conditional evaluation error:
$v_g(s)$ has no $z$ input, whereas $V_{g,z}^{\pi_g}(s)$ can depend on
the persistent risk. At any fixed state $s$, assume
$\E_{Z\sim m}[(V_{g,Z}^{\pi_g}(s))^2]<\infty$.
Writing $c_{\mathrm{val}}\in\mathbb R$ for a scalar prediction and $\Var_{Z\sim m}$
for variance over the root-risk law,
\begin{equation}
  \inf_{c_{\mathrm{val}}\in\mathbb R}
  \E_{Z\sim m}[(V_{g,Z}^{\pi_g}(s)-c_{\mathrm{val}})^2]
  =\Var_{Z\sim m}(V_{g,Z}^{\pi_g}(s)).
  \label{eq:critic-latent-value-variation}
\end{equation}
Indeed, the expected squared error equals this variance plus
$(c_{\mathrm{val}}-\E_{Z\sim m}[V_{g,Z}^{\pi_g}(s)])^2$. When the conditional values differ,
even an ideal scalar reference cannot represent all of them
simultaneously. Moreover, setting
$\bar V_g(s)=\E_{Z\sim m}[V_{g,Z}^{\pi_g}(s)]$ does not generally make
$\E_{Z\sim m}[\widehat P_{g,Z}^{a}(V_{g,Z}^{\pi_g}-\bar V_g)](s)$ zero:
the same $z$ determines both the successor distribution and the
continuation value. This is why HBPO conditions both the residual
and the continuation on one persistent $z$, then averages their
completed action values. Averaging the transition kernel anew at
every step would describe a different stochastic process.

The theorem's assumptions hold for the implemented inventory models.
Let $d_{\max}$ be the largest demand in the frozen model's finite
support, and let $(I,\mathbf p)$ be the initial inventory and pipeline.
Since each accepted regular order and each expedited order is at most
$8$, for every simulated step $k\ge0$,
\begin{equation}
  \begin{aligned}
    |I_k|&\le |I|+\sum_{\ell=1}^4p_\ell+(16+d_{\max})k,\\
    |R_k|&\le \frac{80}{495}+|I_{k+1}|.
  \end{aligned}
  \label{eq:critic-inventory-growth-bound}
\end{equation}
These bounds are uniform in $z$; geometric discounting makes their
sums finite. A frozen Tanh network with finite parameters is bounded
by $\sum_i|w_{3,i}|+|b_3|$, with $w_3,b_3$ as in
Equation~\eqref{eq:critic-network}. Thus the theorem does not require
rewards or exact values to be globally bounded across initial states.

\emph{Interpretation and scope of the guarantee.}
The theorem establishes the correction needed to make the one-step
reference exact under the frozen model. It does not make residual
simulation the only possible policy-evaluation method. Its telescoping
basis is closely related to multi-step temporal-difference
decompositions, including generalized advantage estimation with
its trace parameter set to one \citep{schulman2016gae}; here the increments are
model-integrated residuals and the RB latent persists throughout
each continuation. The reference critic can therefore be used as
a control variate without changing the population action-value target.
Finite simulated tapes still introduce Monte Carlo error, and the
identity alone does not guarantee a variance reduction. Model
misspecification and mismatch between the simulation risk prior and
the benchmark remain separate sources of error. Claims about
real-environment return improvements consequently rest on the
experiments, rather than on this model-evaluation identity alone.

\subsubsection{Finite computation through geometrically stopped continuations.}
\label{app:critic-estimator}
\label{par:critic-geometric-estimator}
We estimate the infinite residual series using a random, almost surely
finite continuation length. For a fixed query $(s,a)$, independently of the entire model path
and of its risk $Z$, draw
\begin{equation}
  N\in\{0,1,\ldots\},
  \qquad \Pr(N=\ell)=(1-\gamma)\gamma^\ell,\quad \ell=0,1,\ldots.
  \label{eq:critic-geometric-length}
\end{equation}
For any $z$ in the common full-$m$-measure set of
Theorem~\ref{thm:critic-residual-completion}, all path expectations
below use $\E_z^{s,a}$,
and the outer expectation also averages over $N$.
Since $\Pr(N\ge k)=\gamma^k$ for integers $k\ge0$, absolute
residual integrability in Theorem~\ref{thm:critic-residual-completion}
justifies
\begin{equation}
    \E\left[\sum_{k=1}^{N}e_{g,z}(S_k)\right]
    =
    \sum_{k=1}^{\infty}
    \Pr(N\ge k)\E[e_{g,z}(S_k)] =
    \E\left[
      \sum_{k=1}^{\infty}\gamma^ke_{g,z}(S_k)
    \right].
  \label{eq:critic-geometric-unbiasedness}
\end{equation}
Let $M=32$ be the number of tapes per sampled query root, and
let $b=1,\ldots,M$ index the tapes. Tape $b$ draws an independent
$Z_b\sim m$ and a length $N_b$ with the law above. Write
$S_{b,k}^{a}$ for its step-$k$ observation after forcing first action
$a$ and following $\pi_g$ thereafter. An ideal action-value estimator is
\begin{equation}
    \widehat Q_g(s,a)
    =
    Q_g^0(s,a)
    +
    \frac1M\sum_{b=1}^{M}
    \sum_{k=1}^{N_b}
    e_{g,Z_b}(S_{b,k}^{a}),
    \qquad M=32.
  \label{eq:critic-q-estimator}
\end{equation}
Each tape samples its own $Z_b$, which is shared across the $81$
alternative first-action branches and held fixed throughout their
continuations. The branches also share the addressed random numbers
used for geometric stopping, exogenous disturbances, and subsequent
policy-action sampling. These shared numbers couple action branches
without changing any one branch's marginal rollout law.

Geometric survival already supplies the discount, so the accumulated
residuals receive no additional factor $\gamma^k$. The $N_b=0$ sum is
empty. These continuations have no fixed 1000-step cutoff or terminal
bootstrap. In RB, the first modeled transition also uses
$\widehat{\mathcal P}_{g,Z_b}$; it is not conditioned on the real
transition's next availability.

Under the stated sampling laws and exact arithmetic,
$\E[\widehat Q_g(s,a)]=Q_g^{\pi_g}(s,a)$.
A finite collection of tapes still produces Monte Carlo error.
For the policy-learning interface, define the sampled centered
advantage by
\begin{equation}
  \widehat A_g(s,a)=\widehat Q_g(s,a)
    -\sum_{a'\in\mathcal A}\pi_g(a'\mid s)\widehat Q_g(s,a').
  \label{eq:critic-sampled-advantage}
\end{equation}
The implementation stores the centered residual correction directly.
When all first-action branches on a tape reach the same observation,
their remaining common tail cancels under this centering and need
not be simulated. Thus it implements the centered estimator without
materializing every absolute $\widehat Q_g(s,a)$. The selection of
query roots and their use by policy updates are described in
Section~\ref{app:policy-updates}.

\subsubsection{The state distribution used to train successor critics.}
\label{app:critic-training-bank}
\label{par:critic-continuation-bank}
The next critics are fitted at states reached when the current behavior
and target policies use the model. Let $\mu_g$ and $\pi_g$ denote the
frozen parent behavior and target policies, respectively. We first specify how starting
observations are selected, then how their continuations determine the
training distribution. Let $s_1,\ldots,s_n$ be the observations in the
generation-$g$ batch, with $n=25{,}000$. These are transition occurrences:
repeated observations retain their multiplicity. Write $\delta_s$ for
the Dirac probability measure concentrated at $s$, so that
$\delta_s(E)=\mathbf 1\{s\in E\}$ for any measurable observation set $E$.

For the training-bank construction, let $D_g=n^{-1}\sum_{i=1}^{n}\delta_{s_i}$ be the empirical
observation distribution, and let $\nu_g^J$ and $\nu_g^P$ be the
normalized empirical distributions obtained by the frozen target-return
and transmission emphatic weights. Each selects an occurrence from this
same batch; their weights are nonnegative and have empirical mean one.
Their exact definitions and occurrence-sampling interpretation are in
Section~\ref{app:emphatic-interface}. The critic treats these three
root distributions as fixed inputs.

For RB, recall the independent synthetic risk in
Equation~\eqref{eq:critic-synthetic-risk}. Let $m=\Unif[0,1/2]$
denote its probability law, with density
$m(dz)=2\mathbf 1_{[0,1/2]}(z)\,dz$. The symbol $\otimes$ denotes
the product of probability measures: for observation and risk sets
$E$ and $C$, $(D_g\otimes m)(E\times C)=D_g(E)m(C)$.
For either weighting role $\chi\in\{J,P\}$, define
\begin{equation}
  \widetilde D_g=D_g\otimes m,
  \qquad
  \widetilde\nu_g^\chi=\nu_g^\chi\otimes m,
  \quad \chi\in\{J,P\},
  \label{eq:critic-augmented-root-measures}
\end{equation}
These product measures correspond to sampling a starting observation
from the indicated distribution and an independent $Z\sim m$.

Let $\widetilde P_{g,\sigma}$ be the transition kernel on $(S,Z)$ that
samples an action from a fixed policy $\sigma$, advances the observation using
$\widehat{\mathcal P}_{g,Z}$, and preserves $Z$. For an augmented-state
distribution $\lambda$, the measure $\lambda\widetilde P_{g,\sigma}$
is the distribution after one such transition. For any measurable set
$E$ of augmented states,
\begin{equation}
  (\lambda\widetilde P_{g,\sigma})(E)
  =\int\widetilde P_{g,\sigma}((s,z),E)\,\lambda(ds,dz).
  \label{eq:critic-measure-transition}
\end{equation}
For a nonnegative integer $\ell$, the power
$\widetilde P_{g,\sigma}^{\,\ell}$ denotes $\ell$ successive
transitions; its zeroth power is the identity kernel.

The three uses of the critic select roots and first actions from
$(D_g,\mu_g)$, $(\nu_g^J,\pi_g)$, and $(\nu_g^P,\mu_g)$,
respectively. Mixing these uses equally gives the distribution after
the first transition,
\begin{equation}
  \widetilde\eta_g
  =
  \frac13\left(
    \widetilde D_g\widetilde P_{g,\mu_g}
    +
    \widetilde\nu_g^J\widetilde P_{g,\pi_g}
    +
    \widetilde\nu_g^P\widetilde P_{g,\mu_g}
  \right),
  \label{eq:critic-first-consumer-transition}
\end{equation}
After this first transition, all three routes follow the frozen target
policy $\pi_g$. We weight their subsequent states geometrically, giving
the normalized critic-training distribution
\begin{equation}
  \widetilde\zeta_g
  =
  (1-\gamma)\sum_{\ell=0}^{\infty}
  \gamma^\ell
  \widetilde\eta_g
  \widetilde P_{g,\pi_g}^{\,\ell}.
  \label{eq:critic-zeta-distribution}
\end{equation}
The coefficients sum to one. In particular, the term $\ell=0$
describes the state immediately after the first transition; it does
not describe the original root observation.

Equation~\eqref{eq:critic-zeta-distribution} directly specifies a
sampling procedure for a finite training bank. Choose one of the three
root-distribution/policy pairs above with probability $1/3$. Draw $S_0$
from that root distribution and an independent $Z\sim m$, sample the first action $\mathsf A_0$
from the selected policy, and simulate the first successor $S_1$.
The pair $(S_1,Z)$ then has distribution $\widetilde\eta_g$.
Independently draw the number of additional target-policy transitions,
\begin{equation}
  G\sim\Geom_0(1-\gamma),
  \qquad \Pr(G=\ell)=(1-\gamma)\gamma^\ell,
  \quad \ell=0,1,\ldots.
  \label{eq:critic-bank-geometric-length}
\end{equation}
Starting from $S_1$, simulate $G$ further transitions under $\pi_g$
and retain $(S_{1+G},Z)$. Conditional on $G=\ell$, its distribution
is $\widetilde\eta_g\widetilde P_{g,\pi_g}^{\,\ell}$; averaging over
$G$ therefore gives exactly $\widetilde\zeta_g$. When $G=0$, retain
$S_1$ without an additional transition. The same sampled $Z$ is used
for every transition on this path.

For each retained pair $(S,z)$, sample one additional next availability
\begin{equation}
  U\sim\widehat K_{B,g,z}(B(S),\cdot).
  \label{eq:critic-paired-supply}
\end{equation}
This draw uses the same tape risk, rather than a newly sampled risk or
the availability from the first transition. The generation contains
$n_\zeta=25{,}000$ such training pairs $(S,U)$, shared by all five
critics; $n_\zeta$ denotes the training-bank size.

The augmented formulation is essential: although $Z$ is independent of
the root observation, it generally becomes statistically dependent on
later observations. Replacing the augmented kernel by powers of a
risk-averaged observation kernel would change $\widetilde\zeta_g$.
For DSIS, the latent coordinate disappears and $U=1$.

\subsubsection{Learning the successor critics by minimizing Bellman-consistency error.}
\label{app:critic-fitting}
\label{par:critic-fitting-objective}
Let $V_\theta$ denote one live scalar critic, with trainable parameter
vector $\theta$; the following objective is optimized separately for
each of the five members. We fit it using a model-based form of the
one-step temporal-difference (TD) error. As in TD-based critic
learning, this error compares the current value prediction with
the immediate reward plus a discounted bootstrap value at the
successor observation. Here, we average the TD error over actions
under the frozen target policy $\pi_g$ and over demand under the
frozen model, conditional on the sampled next availability $U$.
We then minimize the square of this conditional mean.
The resulting population objective is
\begin{equation}
  \begin{aligned}
    \mathcal J_{V,g}(\theta)
    &=\E_{\substack{(S,Z)\sim\widetilde\zeta_g\\
       U\sim\widehat K_{B,g,Z}(B(S),\cdot)}}
    \Biggl[\Biggl(
      V_\theta(S)
      -\sum_{a\in\mathcal A}\pi_g(a\mid S)
       \sum_{d\in\widehat{\mathcal D}_g}
       \widehat K_{D,g}(d^-(S),d)\\
    &\hspace{13em}\cdot
       \bigl[r(S,a,d)+\gamma V_\theta(F(S,a,d,U))\bigr]
    \Biggr)^2\Biggr].
  \end{aligned}
  \label{eq:critic-population-fitting-objective}
\end{equation}
The expression inside the square is the negative of the
conditional expected TD error; this sign convention does not
affect the squared loss. We optimize this objective
with full gradients through both the current and successor value
predictions, as detailed below.

For the frozen bank of $n_\zeta$ pairs sampled above,
$\{(S_\ell,U_\ell)\}_{\ell=1}^{n_\zeta}$, define the empirical objective
\begin{equation}
  \begin{aligned}
    \widehat{\mathcal J}_{V,g}(\theta)
    &=
    \frac1{n_\zeta}\sum_{\ell=1}^{n_\zeta}
    \Biggl[
      V_\theta(S_\ell)
      -
      \sum_{a\in\mathcal A}\pi_g(a\mid S_\ell)
      \sum_{d\in\widehat{\mathcal D}_g}\widehat K_{D,g}(d^-(S_\ell),d)\\
    &\hspace{8em}\cdot
      \left\{
        r(S_\ell,a,d)
        +\gamma V_\theta(F(S_\ell,a,d,U_\ell))
      \right\}
    \Biggr]^2.
  \end{aligned}
  \label{eq:critic-empirical-fitting-objective}
\end{equation}

Recall that $L_g^u$ in
Equation~\eqref{eq:critic-supply-conditioned-backup} averages the
one-step reward and discounted next value over model demand, holding
next availability $u$ fixed. Define the supply-conditioned
target-policy Bellman backup $\mathcal T_g^u$ and its signed error
$\varepsilon_g$ by
\begin{align}
  \mathcal T_g^uV_\theta(s)
  &=\sum_{a\in\mathcal A}\pi_g(a\mid s)L_g^uV_\theta(s,a),
    \label{eq:critic-live-backup}\\
  \varepsilon_g(\theta;s,u)
  &=V_\theta(s)-\mathcal T_g^uV_\theta(s).
    \label{eq:critic-training-residual}
\end{align}
The order of operations is part of the objective: actions and demand are
integrated before squaring, while next availability is sampled and
retained in the training pair. Recall that
$\mathcal T_{g,z}^{\pi_g}$ in
Equation~\eqref{eq:critic-bellman-operator} additionally averages
over next availability. For fixed $(s,z)$, let $\E_U$ and
$\Var_U$ denote expectation and variance under
$U\sim\widehat K_{B,g,z}(B(s),\cdot)$. Then
\begin{equation}
    \E_U\left[
      \bigl(V_\theta(s)-\mathcal T_g^UV_\theta(s)\bigr)^2
    \right] =
    \left[
      V_\theta(s)-(\mathcal T_{g,z}^{\pi_g}V_\theta)(s)
    \right]^2
    +
    \Var_U\left[\mathcal T_g^UV_\theta(s)\right].
  \label{eq:critic-supply-variance-decomposition}
\end{equation}
Therefore, the RB fitting objective is generally different from the
square of the fully supply-integrated Bellman residual. The distinction
vanishes in DSIS, where next availability is deterministic.

Both value terms belong to the same live member and remain
differentiable. Write $\nabla_\theta$ for the gradient with
respect to critic parameters $\theta$. In particular,
\begin{equation}
    \nabla_\theta\varepsilon_g(\theta;s,u)
    =
    \nabla_\theta V_\theta(s) -
    \gamma\sum_{a\in\mathcal A}\pi_g(a\mid s)
    \sum_{d\in\widehat{\mathcal D}_g}\widehat K_{D,g}(d^-(s),d)
    \nabla_\theta V_\theta(F(s,a,d,u)),
  \label{eq:critic-full-residual-gradient}
\end{equation}
and
\begin{equation}
  \nabla_\theta\widehat{\mathcal J}_{V,g}(\theta)
  =
  \frac{2}{n_\zeta}
  \sum_{\ell=1}^{n_\zeta}
  \varepsilon_g(\theta;S_\ell,U_\ell)
  \nabla_\theta\varepsilon_g(\theta;S_\ell,U_\ell).
  \label{eq:critic-full-loss-gradient}
\end{equation}
The model, policy, and sampled bank are fixed during this optimization.
There is no stop-gradient or slowly updated target critic on the
successor-value side of this loss.

\subsubsection{Finite update and use of the successor.}
\label{app:critic-successor-update}
\label{par:critic-finite-update}
For member $j\in\{1,\ldots,5\}$, write $\theta_{g,j}$ for its
frozen parent parameter vector and $\theta_j^{(k)}$ for its live
parameters after $k$ optimizer updates. It starts from
$\theta_j^{(0)}=\theta_{g,j}$, and performs $1000$ Adam updates with
minibatch size $256$ and the configured value learning rate. Adam
uses first- and second-moment decay coefficients
$(\beta_1,\beta_2)=(0.9,0.999)$, denominator stabilizer
$\epsilon_{\mathrm{Adam}}=10^{-8}$, zero weight decay, and fresh
optimizer state; gradients are not clipped. Members share the frozen training bank but
use separately addressed minibatch schedules. The successor is the
finite optimization output,
\begin{equation}
  V_{g+1,j}=V_{\theta_j^{(1000)}},
  \qquad
  v_{g+1}=\frac15\sum_{j=1}^{5}V_{g+1,j}.
  \label{eq:critic-successor}
\end{equation}
This definition does not presume an exact minimizer or Bellman fixed
point.

The subscript $g+1$ records when the fitted critic becomes available;
the policy in its fitting objective is still the frozen parent $\pi_g$.
Throughout this fit, current-generation residual estimation, heuristic
queries, and actor updates continue to use $v_g$. The fitted mean
$v_{g+1}$ becomes their value reference in the next generation. This
separates the live network being optimized from the frozen reference
used to construct the current generation's learning signals.

\subsubsection{Warm and cold critic initialization.}
\label{app:critic-initialization}
Both initialization routes produce five scalar critics
$V_{0,j}$, $j\in\{1,\ldots,5\}$. The warm route first evaluates
the initial categorical policy $\pi_0$ with temporary Q networks and transfers their centered action information; the cold
route directly fits scalar critics on simulated states. Both finish
with the same constant-offset calibration.

\paragraph{Warm initialization and temporary Q evaluation.}
\label{par:critic-warm-evaluation}
\label{par:critic-warm-fqe}
Warm initialization uses the counted $500{,}000$-transition source
prefix. After constructing the initial categorical policy $\pi_0$,
five temporary action-value networks are initialized from the source
critic. A scalar source value is copied across action outputs; for SAC,
the five initial networks use its two online Q networks, two target Q
networks, and an additional copy of the first online Q network. These
copies have separate parameter storage.

Write $n_0=500{,}000$ for the number of initialization pairs. For
source occurrence $i\in\{1,\ldots,n_0\}$, let $s_i$ be its
observation, $d_i$ its realized demand, and $u_i$ its next
availability, all obtained from the recorded transition. In RB,
$d_i$ is recorded in the next observation; in DSIS, it is recovered
from the public inventory balance using the recorded action and
current and next observations. Public dynamics provide counterfactual
successors and rewards for every action:
\begin{equation}
  s'_{i,a}=F(s_i,a,d_i,u_i),
  \qquad
  r_{i,a}=r(s_i,a,d_i).
  \label{eq:critic-factual-counterfactuals}
\end{equation}
No unobserved demand or supply realization is needed for this construction.

For fitted Q evaluation (FQE), let $Q_j^{(k)}$ and
$Q_j^{-,(k)}$ denote member $j$'s online and target action-value
networks at iteration $k$, with parameter vectors $\psi_j^{(k)}$
and $\psi_j^{-,(k)}$, respectively. Each target network starts as a
separate copy of its online network. The initial policy $\pi_0$
remains fixed. For every recorded occurrence $i$ and action
$a\in\mathcal A$, define the detached target
\begin{equation}
  y_{j,i,a}^{(k)}
  =r_{i,a}+\gamma\sum_{a'\in\mathcal A}
    \pi_0(a'\mid s'_{i,a})Q_j^{-,(k)}(s'_{i,a},a').
  \label{eq:critic-fqe-target}
\end{equation}
Here ``detached'' means that differentiation of the FQE loss does
not propagate through $y_{j,i,a}^{(k)}$. At each iteration, draw an
ordered minibatch $\mathcal I^{(k)}=(i_1,\ldots,i_{n_{\mathrm b}})$
of $n_{\mathrm b}=256$ occurrence indices uniformly with replacement
from $\{1,\ldots,n_0\}$; the five members share this minibatch.
For each member $j$, minibatch position $h$, and action $a$, draw a
fresh binary mask $\xi_{j,h,a}^{(k)}$, independently across positions,
actions, members, and iterations, with probability $0.8$ of being
one. Thus $\xi_{j,h,a}^{(k)}\sim\mathrm{Bernoulli}(0.8)$. The loss is
\begin{equation}
  \mathcal L_j^{\mathrm{FQE},(k)}
  =\frac{1}{n_{\mathrm b}|\mathcal A|}
    \sum_{h=1}^{n_{\mathrm b}}\sum_{a\in\mathcal A}
    \xi_{j,h,a}^{(k)}
    \left[Q_j^{(k)}(s_{i_h},a)-y_{j,i_h,a}^{(k)}\right]^2.
  \label{eq:critic-fqe-loss}
\end{equation}
The denominator counts every minibatch-action pair, including masked
pairs; it is not the number of nonzero masks. After the Adam update
of the online parameters to $\psi_j^{(k+1)}$, update the target by
\begin{equation}
  \psi_j^{-,(k+1)}
  =0.995\,\psi_j^{-,(k)}+0.005\,\psi_j^{(k+1)}.
  \label{eq:critic-fqe-target-update}
\end{equation}
FQE performs $2000$ updates per member. Its Adam learning rate is
$1.2020438491997917\times10^{-4}$ for PPO,
$4.418119410401287\times10^{-5}$ for ACE, and
$5.669429427494781\times10^{-4}$ for the other source methods.
These target Q networks belong only to temporary initialization;
generation-level scalar-critic fitting uses the full gradient through
both value predictions, as specified above. Write $Q_j=Q_j^{(2000)}$
for the final online Q network used in the next step.

\paragraph{Scalar fitting for the two initialization routes.}
\label{par:critic-initial-scalar-fit}
\label{par:critic-scalar-initialization}
Five fresh scalar critics are initialized with separately randomized
orthogonal hidden weights, zero hidden biases, and zero output weights
and biases. In the warm case, let the fitted temporary Q networks define
\begin{equation}
  A_j^{\mathrm{init}}(s,a)
  =
  Q_j(s,a)-\sum_{a'\in\mathcal A}\pi_0(a'\mid s)Q_j(s,a').
  \label{eq:critic-initial-advantage}
\end{equation}
The advantage labels $A_j^{\mathrm{init}}$ and the policy $\pi_0$
are held fixed during scalar fitting. The operator
$L_{\mathrm{init}}^u$ is the supply-conditioned backup in
Equation~\eqref{eq:critic-supply-conditioned-backup}, evaluated with
the frozen initialization model. The warm scalar-fitting objective is
\begin{equation}
    \mathcal J_j^{\mathrm{warm}}(\theta) =
    \frac1{n_0}\sum_{i=1}^{n_0}
    \sum_{a\in\mathcal A}\pi_0(a\mid s_i)\cdot
    \left[
      L_{\mathrm{init}}^{u_i}V_\theta(s_i,a) - V_\theta(s_i)- A_j^{\mathrm{init}}(s_i,a)
    \right]^2.
  \label{eq:critic-warm-c1}
\end{equation}
Here $u_i$ is the observed next availability in the source prefix, while
$L_{\mathrm{init}}^{u_i}$ integrates demand using the frozen
initialization model. Both occurrences of $V_\theta$ are differentiated.
This objective transfers centered action information from the temporary
Q critics into the scalar value representation.

Cold initialization instead creates $500$ synthetic trajectories of
length $1000$ under $\pi_0$. Their $n_0=500{,}000$ current-observation/
next-availability pairs are again denoted by $(s_i,u_i)$; here both
coordinates come from model simulation. Each member minimizes
\begin{equation}
    \mathcal J_j^{\mathrm{cold}}(\theta)
    =
    \frac1{n_0}\sum_{i=1}^{n_0}
    \left[
      V_\theta(s_i)
      -
      \sum_{a\in\mathcal A}\pi_0(a\mid s_i)
      L_{\mathrm{init}}^{u_i}V_\theta(s_i,a)
    \right]^2.
  \label{eq:critic-cold-initialization}
\end{equation}
For RB, this simulation starts only after collecting the initial counted
$25{,}000$ transitions and fitting their model. Initial observations
are sampled uniformly, with multiplicity, from the observed joint
reset-state occurrences. Every synthetic trajectory receives an
independent $Z\sim m=\Unif[0,1/2]$, fixed for its entire length. For DSIS, simulation
uses the public initial state and known model. The cold bank does not
reuse private environment tapes.

\paragraph{Calibrating the constant value offset.}
\label{par:critic-offset-calibration}
For either route, each scalar critic is optimized for two complete
epochs with minibatch size $256$ and Adam learning rate
$5.669429427494781\times10^{-4}$. Each epoch visits all $n_0$ pairs
once in a member-specific permutation; the last minibatch contains
$32$ pairs. Adam state is initialized once for this fit and retained
across its two epochs, with the same moment coefficients and
denominator stabilizer as in generation-level fitting. Let $\widetilde V_j$ be the resulting
network. To describe the final calibration, specialize the
supply-conditioned Bellman backup in
Equation~\eqref{eq:critic-live-backup} to the initial policy and model:
\begin{equation}
  \mathcal T_{\mathrm{init}}^u V(s)
  =\sum_{a\in\mathcal A}\pi_0(a\mid s)
    L_{\mathrm{init}}^u V(s,a).
  \label{eq:critic-initial-bellman-backup}
\end{equation}
This operator returns the expected immediate reward plus discounted
next value, averaging over the action and model demand while holding
the next availability $u$ fixed. Its conditioning matches the
initialization pairs $(s_i,u_i)$: $u_i$ is observed in the warm bank
and generated by the model in the cold bank. It is the conditional
counterpart of $\mathcal T_{g,z}^{\pi_g}$ in
Equation~\eqref{eq:critic-bellman-operator}, which additionally
averages over next availability.

Compute the full-bank mean backup-minus-prediction residual,
\begin{equation}
    \overline e_j
    =\frac1{n_0}\sum_{i=1}^{n_0}
      \left[(\mathcal T_{\mathrm{init}}^{u_i}
      \widetilde V_j)(s_i)-\widetilde V_j(s_i)\right].
  \label{eq:critic-initial-mean-residual}
\end{equation}
Adding a constant $c_{\mathrm{off}}$ to every value prediction adds $\gamma c_{\mathrm{off}}$ to
the discounted next-value term, because the action and demand
probabilities each sum to one. Therefore,
\begin{equation}
  \mathcal T_{\mathrm{init}}^u(\widetilde V_j+c_{\mathrm{off}})(s)
  =\mathcal T_{\mathrm{init}}^u\widetilde V_j(s)+\gamma c_{\mathrm{off}},
  \label{eq:critic-initial-backup-shift}
\end{equation}
and hence the residual changes by $-(1-\gamma)c_{\mathrm{off}}$ at every pair:
\begin{equation}
    \mathcal T_{\mathrm{init}}^u(\widetilde V_j+c_{\mathrm{off}})(s)
       -(\widetilde V_j(s)+c_{\mathrm{off}}) =\mathcal T_{\mathrm{init}}^u\widetilde V_j(s)
       -\widetilde V_j(s)-(1-\gamma)c_{\mathrm{off}}.
  \label{eq:critic-initial-residual-shift}
\end{equation}
The final initialization chooses
\begin{equation}
  V_{0,j}(s)
  =\widetilde V_j(s)+\frac{\overline e_j}{1-\gamma},
  \label{eq:critic-initial-c3-shift}
\end{equation}
implemented by shifting the scalar output bias (up to floating-point
rounding of that bias). The new mean residual
is $\overline e_j-(1-\gamma)\overline e_j/(1-\gamma)=0$ in exact
arithmetic. This removes a constant offset in the empirical Bellman
error; it does not set every statewise residual to zero. After warm
initialization, the temporary Q networks are discarded.

\subsection{Emphatic Weighting: Mass Learning, Heuristic Interest, and Cross-Generation Use}
\label{app:emphatic-interface}

\subsubsection{Meaning of emphatic weighting and the collected data.}
\label{app:emphatic-data}
Emphatic weighting assigns state-dependent importance to learning updates
by propagating an interest signal through transitions. Here interest is
an externally specified or data-derived nonnegative function that states
which locations should receive additional learning emphasis. This separates
how frequently an observation occurs in behavior data from how strongly
it contributes to a learning objective. At learning generation $g$, HBPO
uses two weighting roles: target-return learning, denoted by $J$, and
behavior-to-target transmission, denoted by $P$. We call their positive
state scores $M_g^J$ and $M_g^P$ raw masses. They are functions of the
observation, not normalized state distributions. Normalization and
shrinkage convert these scores into weights and then into empirical
probability measures for the corresponding learning objectives.

Let $\mathcal D_g$ be the indexed multiset of $n=25{,}000$ transitions
collected along trajectories of the frozen behavior policy $\mu_g$.
The index $i=1,\ldots,n$ identifies an occurrence, including repeated
observations and transitions:
\begin{equation}
  \mathcal D_g=\bigl((s_i,a_i,s_i')\bigr)_{i=1}^{n}.
  \label{eq:emphatic-transition-data}
\end{equation}
Recall that $D_g$ averages over current observations.
Let $D_g'$ average over successor observations and let $\Gamma_g$ retain
the paired transition occurrences. Writing $\delta_x$ for unit point
mass at $x$, these empirical measures are
\begin{equation}
  D_g=\frac1n\sum_{i=1}^{n}\delta_{s_i},
  \qquad
  D_g'=\frac1n\sum_{i=1}^{n}\delta_{s_i'},
  \qquad
  \Gamma_g=\frac1n\sum_{i=1}^{n}\delta_{(s_i,a_i,s_i')}.
  \label{eq:emphatic-empirical-measures}
\end{equation}
Repeated observations retain their occurrence multiplicities. Masses are
fitted at successor observations, whereas the weights consumed in
generation $g$ are computed from the parent masses at current observations.
The symbol $\chi\in\{J,P\}$ denotes a weighting role.

\subsubsection{Emphatic weighting in prior work.}
\label{app:emphatic-prior-work}
The emphatic TD (ETD) construction of \citet{sutton2016emphatic}
propagates user-specified interest through target-policy transitions to
weight bootstrapped value updates. Its linear stability theorem concerns
the expected update under the paper's assumptions; it is not a stability
theorem for nonlinear mass networks. To summarize the follow-on part of
this construction, consider a finite Markov model, fixed behavior and
target policies $\mu$ and $\pi$, a constant discount $0<\gamma<1$, and
a fixed nonnegative interest function $\iota$. Assume that $d_\mu$ is a
strictly positive stationary state distribution under $\mu$, that
$S_0\sim d_\mu$, and that $\mu(a\mid s)>0$ whenever $\pi(a\mid s)>0$.
Generate the trajectory under $\mu$. Let $k$ index its steps, and let
$F_k^{\mathrm{fo}}$ denote the
follow-on trace. Initialize $F_0^{\mathrm{fo}}=\iota(S_0)$; for $k\ge1$,
using the target-to-behavior action ratio $\varrho_{k-1}$, set
\begin{equation}
  F_k^{\mathrm{fo}}
  =\iota(S_k)+\gamma\varrho_{k-1}F_{k-1}^{\mathrm{fo}},
  \qquad
  \varrho_{k-1}
  =\frac{\pi(\mathsf A_{k-1}\mid S_{k-1})}
         {\mu(\mathsf A_{k-1}\mid S_{k-1})}.
  \label{eq:emphatic-classical-followon}
\end{equation}
Let $P(s'\mid s,a)$ be this model's transition probability and let
$P_\pi(s,s')=\sum_a\pi(a\mid s)P(s'\mid s,a)$ be its target-policy
transition matrix. Define the expected follow-on state mass by
$f_\iota(s)=d_\mu(s)\lim_{k\to\infty}\E[F_k^{\mathrm{fo}}\mid S_k=s]$. The mass satisfies
\begin{equation}
  f_\iota=d_\mu\odot\iota+\gamma P_\pi^\top f_\iota,
  \qquad
  f_\iota=(\mathrm{Id}-\gamma P_\pi^\top)^{-1}
                  (d_\mu\odot\iota).
  \label{eq:emphatic-classical-mass}
\end{equation}
Thus the conditional mean trace and the state mass obtained by
multiplying it by $d_\mu$ are different objects. We use this follow-on
component to explain the connection, rather than identifying it with
the full emphasis of every ETD trace-parameter setting.

OffPAC \citep{degris2012offpolicy} corrects sampled actions with importance
ratios but retains behavior-state weighting in its actor semi-gradient.
For this literature comparison, let $\vartheta$ be the actor parameter
vector and let $\pi_\vartheta$ be a differentiable target policy.
For the displayed score formula, assume that both $\mu(a\mid s)$ and
$\pi_\vartheta(a\mid s)$ are strictly positive for every state and action.
Let $Q^{\pi_\vartheta}$ and $V^{\pi_\vartheta}$ be the exact discounted
action and state values in the finite Markov model, and let
$\nabla_\vartheta$ denote differentiation with respect to $\vartheta$.
Using exact action values to isolate the state-weighting issue, write
OffPAC's basic actor direction as $g_{\mathrm{OffPAC}}(\vartheta)$:
\begin{equation}
  \begin{aligned}
    g_{\mathrm{OffPAC}}(\vartheta)
    &=\E_{S\sim d_\mu,\,\mathsf A\sim\mu(\cdot\mid S)}
      \left[
        \frac{\pi_\vartheta(\mathsf A\mid S)}{\mu(\mathsf A\mid S)}
        Q^{\pi_\vartheta}(S,\mathsf A)
        \nabla_\vartheta\log\pi_\vartheta(\mathsf A\mid S)
      \right]\\
    &=\sum_s d_\mu(s)\sum_a
       \nabla_\vartheta\pi_\vartheta(a\mid s)
       Q^{\pi_\vartheta}(s,a).
  \end{aligned}
  \label{eq:emphatic-offpac-direction}
\end{equation}
Action importance sampling changes the action average to the target
policy, while the outer state average remains weighted by $d_\mu$.

The policy-gradient theorem underlying Actor--Critic with Emphatic
weightings (ACE; \citealp{imani2018emphatic}) supplies the emphatic
state weighting for the exact gradient of the interest-weighted
excursion objective $\mathcal J_\iota$:
\begin{equation}
  \mathcal J_\iota(\vartheta)
  =\sum_s d_\mu(s)\iota(s)V^{\pi_\vartheta}(s).
  \label{eq:emphatic-excursion-objective}
\end{equation}
Holding $\mu$, $d_\mu$, and $\iota$ fixed when differentiating, its
policy-gradient theorem gives
\begin{equation}
  \nabla_\vartheta\mathcal J_\iota(\vartheta)
  =\sum_s f_\iota(s)\sum_a
       \nabla_\vartheta\pi_\vartheta(a\mid s)
       Q^{\pi_\vartheta}(s,a),
  \label{eq:emphatic-ace-gradient}
\end{equation}
where $f_\iota$ is the follow-on state mass in
Equation~\eqref{eq:emphatic-classical-mass}, evaluated at
$\pi=\pi_\vartheta$. For unit interest, comparison with
Equation~\eqref{eq:emphatic-offpac-direction} isolates the missing
follow-on state weighting in the semi-gradient. This equality uses exact
values and state weights; finite trace and value estimates introduce
additional approximation. \citet[Section~5.2]{graves2023offpolicy}
also study directly learning conditional emphatic weights from
predecessor transitions. This motivates the regression viewpoint below.
HBPO specifies its own labels, anchored mass functions, shrinkage, and
generation schedule; the cited results do not certify these modifications.

\subsubsection{Why learn positive masses, and what the regression represents.}
\label{app:emphatic-mass-learning}
The purpose of the mass networks is to store a state-dependent importance
field that can be evaluated on a later generation's observations. A
recorded transition provides a predecessor carrying existing mass and a
successor at which that mass should contribute to learning. Fitting a
shared function at successor observations pools this incoming evidence
and generalizes it across the fixed observation features. The two roles
use different labels, specified below, but share the following
parameterization and fitting principle.

Recall that $\chi\in\{J,P\}$ indexes the two raw mass functions.
For each role, let $f_{\omega_g^\chi}$ be a separate scalar network with
parameter vector $\omega_g^\chi$, two width-$128$ Tanh hidden layers,
and a scalar output. Recall the fixed observation features $\phi$ from
Equation~\eqref{eq:critic-features}. Let $s_\circ$ be the reference
observation with zero inventory and pipeline and available regular
supply; its previous-demand coordinate is also zero when present.
Let $\beta_g^\chi>0$ be the prescribed mass at this reference. Define
\begin{equation}
  M_g^\chi(s)
  =\beta_g^\chi
    \exp\!\left[
      f_{\omega_g^\chi}(\phi(s))
      -f_{\omega_g^\chi}(\phi(s_\circ))
    \right],
  \qquad \beta_g^\chi>0.
  \label{eq:emphatic-mass-parameterization}
\end{equation}
Thus $M_g^\chi(s)>0$ and $M_g^\chi(s_\circ)=\beta_g^\chi$.
The reference value is imposed by the parameterization. If $s_\circ$
occurs among the observed successors, this value need not equal the
unrestricted regression target there.

For a fixed role $\chi$, let $Y_i^\chi>0$ be the regression label attached
to transition occurrence $i$; the two roles construct these labels below.
For a candidate positive mass function $M$, define the full-data loss
$\mathcal L_g^\chi(M)$ by
\begin{equation}
  \mathcal L_g^\chi(M)
  =\frac1n\sum_{i=1}^{n}
    \left[M(s_i')-Y_i^\chi\log M(s_i')\right].
  \label{eq:emphatic-poisson-loss}
\end{equation}
This is the Poisson-form loss (a quasi-likelihood objective for positive
real labels), with label-only terms omitted. It does not assume that
these labels are integer-valued Poisson observations. Recall that
$\Gamma_g$ in Equation~\eqref{eq:emphatic-empirical-measures} samples
recorded transition occurrences uniformly. More explicitly, draw an
occurrence index $I$ uniformly from $\{1,\ldots,n\}$ and set
$(S,\mathsf A,S',Y^\chi)=(s_I,a_I,s_I',Y_I^\chi)$.
This defines the joint empirical law including the label; expectations
subscripted by $\Gamma_g$ below use this attached-label law when needed.
Its current and successor marginals remain $D_g$ and $D_g'$.

For an observed successor $s'$, write $x_{\log}=\log M(s')$, with $\log$
denoting the natural logarithm. Let $\ell(x_{\log};s')$ be the conditional
loss. Its value and first two derivatives are
\begin{equation}
  \begin{aligned}
    \ell(x_{\log};s')
    &=e^{x_{\log}}-\E_{\Gamma_g}[Y^\chi\mid S'=s']\,x_{\log},\\
    \frac{\partial\ell}{\partial x_{\log}}
    &=e^{x_{\log}}-\E_{\Gamma_g}[Y^\chi\mid S'=s'],
    \qquad
    \frac{\partial^2\ell}{\partial x_{\log}^2}=e^{x_{\log}}>0.
  \end{aligned}
  \label{eq:emphatic-conditional-loss}
\end{equation}
Let $M^\star(s')$ denote the unrestricted pointwise minimizer at an
observed successor. The positive conditional mean and strict convexity give
\begin{equation}
  M^\star(s')=\E_{\Gamma_g}[Y^\chi\mid S'=s'].
  \label{eq:emphatic-conditional-minimizer}
\end{equation}
A successor receives the average label carried into it by its observed
predecessors. This calculation explains the meaning of the labels and
the choice of loss. The actual learner approximates this regression
with the anchored network and finite optimization; it does not set each
prediction equal to the unrestricted minimizer.

\subsubsection{Target-return mass $M^J$.}
\label{app:emphatic-j-mass}
The $J$ mass addresses the difference between the states represented by
behavior data and those reached when the target policy continues from
that behavior-distributed source. Action importance ratios alone change
the action distribution at a given observation; they do not change how
often that observation contributes to the update. Propagating mass
through target-policy transitions provides a state-level adjustment for
the target-return channel. Its ideal reference is specified below, so
this motivation does not presume a stationary target-occupancy ratio.

Recall that $\pi_g$ and $\mu_g$ are the frozen parent target and behavior
policies. Define their action-ratio function $\varrho_g$ and its value at
recorded occurrence $i$ by
\begin{equation}
  \varrho_g(s,a)=\frac{\pi_g(a\mid s)}{\mu_g(a\mid s)},
  \qquad
  \varrho_{g,i}=\varrho_g(s_i,a_i).
  \label{eq:emphatic-parent-action-ratio}
\end{equation}
The implementation evaluates this ratio from parent-policy log
probabilities without clipping; the categorical policies have common
positive action support. The $J$ learner has an online parameter vector
$\omega_{g,k}^J$ and a private slow copy $\overline\omega_{g,k}^J$ at
inner iteration $k$, initialized by
$\omega_{g,0}^J=\overline\omega_{g,0}^J=\omega_g^J$.
For $k=0,\ldots,999$, define the slow mass $\overline M_{g,k}^J$ and
the detached regression label $Y_{g,i,k}^J$ as
\begin{align}
  \overline M_{g,k}^J(s)
  &=\beta_g^J\exp\!\left[
      f_{\overline\omega_{g,k}^J}(\phi(s))
      -f_{\overline\omega_{g,k}^J}(\phi(s_\circ))
    \right],
    \label{eq:emphatic-j-slow-mass}\\
  Y_{g,i,k}^J
  &=(1-\gamma)+\gamma\varrho_{g,i}\overline M_{g,k}^J(s_i).
    \label{eq:emphatic-j-label}
\end{align}
The source term $1-\gamma$ injects positive mass; the continuation term
carries predecessor mass through the recorded transition with an action
correction toward $\pi_g$. Recall that $\beta_g^J$ is the parent reference
mass. At iteration $k$, the online prediction uses the parameterization
in Equation~\eqref{eq:emphatic-mass-parameterization} with parameters
$\omega_{g,k}^J$ and successor reference value $\beta_{g+1}^J=1-\gamma$.
It is fitted at $s_i'$ using Equation~\eqref{eq:emphatic-poisson-loss}
with $Y_i^\chi=Y_{g,i,k}^J$, holding the labels fixed during that
optimization step. After the step, the slow parameters are updated by
\begin{equation}
  \overline\omega_{g,k+1}^J
  =0.995\,\overline\omega_{g,k}^J
   +0.005\,\omega_{g,k+1}^J.
  \label{eq:emphatic-j-slow-update}
\end{equation}
The slow mass retains $\beta_g^J$ throughout the generation, while the
online mass uses $\beta_{g+1}^J$. Only the final online network is
committed as $M_{g+1}^J$.

For a scalar state function $h$, define the empirical reverse-transition
operator $\mathcal B_g^J$ at each observed successor by
\begin{equation}
  (\mathcal B_g^J h)(s')
  =\E_{\Gamma_g}
    [\varrho_g(S,\mathsf A)h(S)\mid S'=s'].
  \label{eq:emphatic-j-reverse-operator}
\end{equation}
For a fixed inner iteration $k$, let $M_{g,k}^{J,\star}(s')$ be the
unrestricted pointwise minimizer with the slow mass
$\overline M_{g,k}^J$ held fixed. Recalling the regression identity in
Equation~\eqref{eq:emphatic-conditional-minimizer}, we obtain
\begin{equation}
  M_{g,k}^{J,\star}(s')
  =(1-\gamma)+\gamma(\mathcal B_g^J\overline M_{g,k}^J)(s').
  \label{eq:emphatic-j-conditional-target}
\end{equation}
This empirical identity holds on the successor-data support and does
not require $D_g=D_g'$.

For an ideal population reference, now replace the empirical transition
law $\Gamma_g$ by the stationary transition law of a finite Markov model.
Fix policies $\mu$ and $\pi$, assume a strictly positive stationary
behavior distribution $d_\mu$ and target-action support covered by $\mu$,
and write $P(s'\mid s,a)$ for the model's transition probability. Define
the population joint law $\Gamma_\mu$ and target transition matrix $P_\pi$ by
\begin{equation}
  \Gamma_\mu(s,a,s')=d_\mu(s)\mu(a\mid s)P(s'\mid s,a),
  \qquad
  P_\pi(s,s')=\sum_a\pi(a\mid s)P(s'\mid s,a).
  \label{eq:emphatic-population-transition-law}
\end{equation}
Both state marginals of $\Gamma_\mu$ are $d_\mu$. This replacement is a
population idealization, not a consequence of equality of empirical
marginals. Remove the neural parameterization and its reference-anchor
constraint for this idealization. Suppress the generation and inner-iteration indices, and let
$M^J$ be the unrestricted positive fixed point of the population
conditional equation
\begin{equation}
  M^J(s')=(1-\gamma)+\gamma\E_{\Gamma_\mu}
  \left[\frac{\pi(\mathsf A\mid S)}{\mu(\mathsf A\mid S)}M^J(S)
  \,\middle|\,S'=s'\right].
  \label{eq:emphatic-j-population-fixed-point}
\end{equation}
Define its induced state mass $e^J$ by
\begin{equation}
  e^J(s)=d_\mu(s)M^J(s),
  \qquad\text{or}\qquad e^J=d_\mu\odot M^J.
  \label{eq:emphatic-j-induced-mass}
\end{equation}
This multiplication converts a per-occurrence importance multiplier
into a state mass: behavior visits $s$ with frequency $d_\mu(s)$, and
each such occurrence contributes mass $M^J(s)$. For any scalar test
function $h$, it follows that
$\E_{d_\mu}[M^J(S)h(S)]=\sum_s e^J(s)h(s)$.
Multiplying the population conditional fixed-point equation by
$d_\mu(s')$ gives
\begin{equation}
  \begin{aligned}
    e^J(s')
    &=(1-\gamma)d_\mu(s')
      +\gamma\sum_s\sum_{a:\mu(a\mid s)>0}d_\mu(s)\mu(a\mid s)
        P(s'\mid s,a)\frac{\pi(a\mid s)}{\mu(a\mid s)}M^J(s)\\
    &=(1-\gamma)d_\mu(s')
      +\gamma\sum_s P_\pi(s,s')e^J(s).
  \end{aligned}
  \label{eq:emphatic-j-flow}
\end{equation}
Here $P$ is the population transition probability just defined.
Summing over $s'$ gives $\sum_s e^J(s)=1$. Recall that state masses are
column vectors, $\top$ denotes transpose, and $\mathrm{Id}$ is the
identity matrix. Expanding the inverse in powers indexed by the
nonnegative integer $\ell$ yields
\begin{equation}
    e^J
    =(1-\gamma)(\mathrm{Id}-\gamma P_\pi^\top)^{-1}d_\mu =(1-\gamma)\sum_{\ell=0}^{\infty}
       \gamma^\ell(P_\pi^\top)^\ell d_\mu.
  \label{eq:emphatic-j-resolvent}
\end{equation}
The ideal reference is the discounted visitation distribution of
target-policy excursions beginning from $d_\mu$. The multiplier
$M^J(s)=e^J(s)/d_\mu(s)$ adjusts behavior-state frequencies toward this
reference. The reference distribution generally differs from a stationary
target distribution
$d_\pi$ satisfying $P_\pi^\top d_\pi=d_\pi$, and from a finite evaluation
trajectory's occupancy. Uniqueness of $d_\pi$ is not needed for this
reference construction.

\subsubsection{From heuristic improvement to confidence and interest.}
\label{app:emphatic-confidence}
The $P$ role receives interest from the heuristic's estimated improvement
over the parent target policy. Recall that $w_i^J$ is the positive,
mean-one weight of current observation occurrence $i$, obtained from the
parent mass $M_g^J$ by Equation~\eqref{eq:emphatic-consumer-formula}.
Draw $16$ anchor occurrences with replacement, selecting occurrence $i$
with probability $w_i^J/n$; denote them by $\bar s_{g,f}$ for
$f=1,\ldots,16$. At each anchor, run $32$ paired simulations, indexed
by $b=1,\ldots,32$, with horizon $H=1000$ and period index
$k=0,\ldots,H-1$. Recall the heuristic controller $q_{H,g}$ defined in
Equation~\eqref{eq:h-controller}. Superscripts $\mathrm H$ and $\pi$
denote the intervention and baseline arms, while
$S_{g,f,b,k}^{\mathrm H},S_{g,f,b,k}^{\pi}$ and
$\mathsf A_{g,f,b,k}^{\mathrm H},\mathsf A_{g,f,b,k}^{\pi}$ denote their
observations and actions. Both arms start at
$S_{g,f,b,0}^{\mathrm H}=S_{g,f,b,0}^{\pi}=\bar s_{g,f}$, with action laws
\begin{equation}
    \mathsf A_{g,f,b,k}^{\mathrm H}
    \sim
      \begin{cases}
        q_{H,g}(\cdot\mid S_{g,f,b,k}^{\mathrm H}),&0\le k<9,\\
        \pi_g(\cdot\mid S_{g,f,b,k}^{\mathrm H}),&9\le k<H,
      \end{cases}\quad
    \mathsf A_{g,f,b,k}^{\pi}
    \sim\pi_g(\cdot\mid S_{g,f,b,k}^{\pi}),\qquad 0\le k<H.
  \label{eq:emphatic-confidence-policies}
\end{equation}
Each arm advances from its own observation using the frozen fitted
reward--observation kernel in Equation~\eqref{eq:critic-fitted-kernel}.
Within each pair, the arms use common random inputs, including the root
risk draw in RB (Equation~\eqref{eq:critic-synthetic-risk}); that risk is
retained throughout the pair as in
Equation~\eqref{eq:critic-augmented-model}.

Let $C_{g,f,b,k}^{\mathrm H}$ and $C_{g,f,b,k}^{\pi}$ be the realized
period costs of the two arms, and let $R_{g,f,b,k}^{\mathrm H}$ and
$R_{g,f,b,k}^{\pi}$ be their normalized rewards. Using
$R=-C/495$ from Equation~\eqref{eq:critic-reward}, define the complete
paired improvement $\Delta_{g,f,b}$ by
\begin{equation}
    \Delta_{g,f,b}
    =\sum_{k=0}^{H-1}
       \bigl(C_{g,f,b,k}^{\pi}-C_{g,f,b,k}^{\mathrm H}\bigr) = 495\sum_{k=0}^{H-1}
       \bigl(R_{g,f,b,k}^{\mathrm H}-R_{g,f,b,k}^{\pi}\bigr).
  \label{eq:emphatic-paired-improvement}
\end{equation}
Positive $\Delta_{g,f,b}$ means that the finite heuristic intervention
reduced total cost over the $H$ simulated periods. These are complete
undiscounted $H$-period returns. The common factor $495$ cancels from the
standardized statistic. Let $\overline\Delta_{g,f}$ and
$\widehat\sigma_{g,f}^{\,2}$ be the paired sample mean and sample
variance:
\begin{equation}
  \overline\Delta_{g,f}
  =\frac1{32}\sum_{b=1}^{32}\Delta_{g,f,b},
  \qquad
  \widehat\sigma_{g,f}^{\,2}
  =\frac1{31}\sum_{b=1}^{32}
       (\Delta_{g,f,b}-\overline\Delta_{g,f})^2.
  \label{eq:emphatic-confidence-moments}
\end{equation}
Let $\widehat\sigma_{g,f}$ be the nonnegative square root of this sample
variance and let $\operatorname{CDF}_{t_{31}}$ denote the cumulative
distribution function of the standard Student-$t$ distribution with
$31$ degrees of freedom. For $\widehat\sigma_{g,f}>0$, define the
confidence score $c_{g,f}$ by
\begin{equation}
  c_{g,f}
  =\operatorname{CDF}_{t_{31}}\!\left(
      \frac{\sqrt{32}\,\overline\Delta_{g,f}}{\widehat\sigma_{g,f}}
    \right).
  \label{eq:emphatic-confidence-score}
\end{equation}
At zero sample standard deviation, set $c_{g,f}$ to $1$, $1/2$, or $0$
according as the mean improvement is positive, zero, or negative. This
Student-$t$ transformation supplies a bounded directional score; no
calibrated posterior-probability interpretation is required.

Recall the fixed feature vector $\phi$. Define the normalized Gaussian
interpolation coefficients $a_{g,f}(s)$ and the resulting confidence
field $\widehat c_g(s)$ below, where $\|\cdot\|_2$ is the Euclidean norm
and $h$ in the denominator is a dummy anchor index:
\begin{equation}
  \begin{aligned}
    a_{g,f}(s)
    &=\frac{
        \exp[-\|\phi(s)-\phi(\bar s_{g,f})\|_2^2/2]
       }{
        \sum_{h=1}^{16}\exp[-\|\phi(s)-\phi(\bar s_{g,h})\|_2^2/2]
       },\\
    \widehat c_g(s)&=\sum_{f=1}^{16}a_{g,f}(s)c_{g,f}.
  \end{aligned}
  \label{eq:emphatic-confidence-interpolation}
\end{equation}
Let $i_g^P$ be the transmission interest. Recall that $D_g'$ is the
empirical measure of all successor observations $s_i'$. Centering the
confidence field over this complete successor data gives
\begin{equation}
  i_g^P(s)
  =1+\widehat c_g(s)-\E_{D_g'}[\widehat c_g],
  \qquad
  \E_{D_g'}[\widehat c_g]
  =\frac1n\sum_{i=1}^{n}\widehat c_g(s_i').
  \label{eq:emphatic-p-interest}
\end{equation}
Hence $\E_{D_g'}[i_g^P]=1$. For finite features, all interpolation
coefficients are strictly positive. If the anchor scores differ,
each interpolated score lies strictly between their minimum and
maximum; if all scores coincide, the field is constant. Together with
$c_{g,f}\in[0,1]$, this proves $i_g^P(s)>0$ in exact arithmetic.
The interest emphasizes locations with confidence above the
successor-data average. Constant confidence, including uniformly high
confidence, gives $i_g^P\equiv1$.

\subsubsection{Transmission mass $M^P$.}
\label{app:emphatic-p-mass}
The transmission channel requires a different notion of state importance.
Behavior visitation describes where experience was collected, but it does
not indicate where a finite heuristic intervention has stronger evidence
of improving on the parent target. The $P$ mass converts that spatial
variation in confidence into persistent state emphasis for transmission.
Its comparison is between behavior visitation and a confidence-weighted
reference propagated under the behavior policy. It does not estimate
visitation under $q_{H,g}$ or directly measure behavior--target action
disagreement; the latter enters the separate transmission objective.

Recall that $M_g^P$ is the frozen parent transmission mass and that $D_g$
averages over current observations $s_i$. Define its normalized
multiplier $R_g^P$ by
\begin{equation}
  R_g^P(s)=\frac{M_g^P(s)}{\E_{D_g}[M_g^P]},
  \qquad \E_{D_g}[R_g^P]=1.
  \label{eq:emphatic-p-parent-normalization}
\end{equation}
Before optimization, construct an unnormalized label $\widetilde Y_{g,i}^P$
and its full-data mean-one normalization $Y_{g,i}^P$ for every transition
occurrence $i$. Freeze this complete label vector throughout the fit:
\begin{equation}
  \widetilde Y_{g,i}^P=i_g^P(s_i')+\gamma R_g^P(s_i),
  \qquad
  Y_{g,i}^P
  =\frac{\widetilde Y_{g,i}^P}
        {n^{-1}\sum_{j=1}^{n}\widetilde Y_{g,j}^P}.
  \label{eq:emphatic-p-label}
\end{equation}
Because interest and normalized parent mass have unit mean on their
respective data marginals, exact arithmetic gives
\begin{equation}
  Y_{g,i}^P
  =\frac{i_g^P(s_i')+\gamma R_g^P(s_i)}{1+\gamma}.
  \label{eq:emphatic-p-label-simplified}
\end{equation}
The implementation computes the full-data denominator explicitly.
This label contains no action importance ratio and no evolving slow
network: it combines successor interest with normalized predecessor
mass carried along behavior transitions. For a scalar state function
$h$, define the empirical reverse-transition operator $\mathcal B_g^P$
at each observed successor by
\begin{equation}
  (\mathcal B_g^P h)(s')=\E_{\Gamma_g}[h(S)\mid S'=s'].
  \label{eq:emphatic-p-reverse-operator}
\end{equation}
Let $M_g^{P,\star}$ be the unrestricted conditional regression target.
By Equation~\eqref{eq:emphatic-conditional-minimizer}, on the
successor-data support it is
\begin{equation}
  M_g^{P,\star}(s')
  =\frac{i_g^P(s')+\gamma(\mathcal B_g^P R_g^P)(s')}{1+\gamma}.
  \label{eq:emphatic-p-conditional-target}
\end{equation}
The online network starts from the parent parameters and is fitted with
Equation~\eqref{eq:emphatic-poisson-loss} using the fixed labels
$Y_i^\chi=Y_{g,i}^P$. Recall the reference observation $s_\circ$ in
Equation~\eqref{eq:emphatic-mass-parameterization}; the successor
reference value is set to $\beta_{g+1}^P=i_g^P(s_\circ)$.
This is an imposed anchor. If $s_\circ$ occurs among the observed
successors, it need not equal the unrestricted conditional target there.

For the ideal population reference, recall the stationary law
$\Gamma_\mu$ and positive behavior distribution $d_\mu$ in
Equation~\eqref{eq:emphatic-population-transition-law}. Define the
behavior-policy transition matrix
$P_\mu(s,s')=\sum_a\mu(a\mid s)P(s'\mid s,a)$.
Freeze a positive interest function $i^P$ normalized by
$\E_{d_\mu}[i^P]=1$, and replace the empirical conditional expectations
by expectations under $\Gamma_\mu$. For any normalized predecessor
multiplier, the resulting unrestricted conditional target has mean
$(1+\gamma)/(1+\gamma)=1$ under $d_\mu$. At an exact fixed point,
therefore, the raw output equals its normalized multiplier, which we
denote by $R^P$. Define the associated state mass
$e^P(s)=d_\mu(s)R^P(s)$. Multiplication of the population conditional
fixed-point equation by $d_\mu(s')$ gives
\begin{equation}
  e^P=\frac{d_\mu\odot i^P+\gamma P_\mu^\top e^P}{1+\gamma}.
  \label{eq:emphatic-p-flow}
\end{equation}
Define the effective propagation factor $\lambda_P=\gamma/(1+\gamma)$,
so $1-\lambda_P=1/(1+\gamma)$. This is the factor induced by the
$P$-label normalization, not an ETD eligibility-trace parameter. Solving
with the identity matrix $\mathrm{Id}$ gives
\begin{equation}
  e^P=(1-\lambda_P)
      (\mathrm{Id}-\lambda_PP_\mu^\top)^{-1}(d_\mu\odot i^P).
  \label{eq:emphatic-p-resolvent}
\end{equation}
This reference starts from interest-weighted behavior states and
propagates under $\mu$. In this idealization, $R^P=e^P/d_\mu$ describes
the corresponding change in state importance. Unit interest has the
neutral fixed point $R^P\equiv1$, irrespective of behavior--target policy
disagreement.

Both mass learners use $\gamma=0.995$ and $1000$ Adam updates with
minibatches of $256$ uniformly sampled transition occurrences, with
replacement. Each role has fresh optimizer state and the configured
emphatic learning rate; the Adam coefficients are those specified in
Section~\ref{app:critic-successor-update}. All labels are detached
during differentiation. The $J$ labels evolve through its private slow
network, while the $P$ labels remain fixed throughout its fit.

\subsubsection{Deriving consumer weights from the masses.}
\label{app:emphatic-consumer-weights}
Recall the frozen parent raw masses $M_g^\chi$ for
$\chi\in\{J,P\}$ and the current-observation empirical measure $D_g$.
Let $\overline m_g^\chi$ be the full-data mean mass,
$R_g^\chi$ the normalized mass multiplier (agreeing with
Equation~\eqref{eq:emphatic-p-parent-normalization} for $P$), and
$v_{M,g}^\chi$ its mean squared deviation from one:
\begin{equation}
  \overline m_g^\chi=\E_{D_g}[M_g^\chi],
  \qquad
  R_g^\chi(s)=\frac{M_g^\chi(s)}{\overline m_g^\chi},
  \qquad
  v_{M,g}^\chi=\E_{D_g}[(R_g^\chi-1)^2].
  \label{eq:emphatic-normalized-mass}
\end{equation}
Since $\E_{D_g}[R_g^\chi]=1$, its second moment equals
$1+v_{M,g}^\chi$. Define the shrinkage coefficient $\alpha_g^\chi$ and
the consumer weight function $w_g^\chi$ by
\begin{equation}
  \alpha_g^\chi=(1+v_{M,g}^\chi)^{-1/2},
  \qquad
  w_g^\chi(s)=1+\alpha_g^\chi(R_g^\chi(s)-1).
  \label{eq:emphatic-shrinkage}
\end{equation}
Substitution gives the main-text expression
(Equation~\eqref{eq:emphatic-weights}):
\begin{equation}
  w_g^\chi(s)
  =1+\frac{M_g^\chi(s)-\E_{D_g}[M_g^\chi]}
            {\sqrt{\E_{D_g}[(M_g^\chi)^2]}},
  \qquad \chi\in\{J,P\}.
  \label{eq:emphatic-consumer-formula}
\end{equation}
The moments are computed once over all current-state occurrences and
remain fixed during subsequent minibatch computations.

For finite positive masses, $0<\alpha_g^\chi\le1$, so
\begin{equation}
  w_g^\chi=(1-\alpha_g^\chi)+\alpha_g^\chi R_g^\chi>0,
  \qquad \E_{D_g}[w_g^\chi]=1.
  \label{eq:emphatic-weight-properties}
\end{equation}
Multiplying $M_g^\chi$ by a positive constant leaves $R_g^\chi$ and
$w_g^\chi$ unchanged. Moreover,
\begin{equation}
  \E_{D_g}[(w_g^\chi-1)^2]
  =\frac{v_{M,g}^\chi}{1+v_{M,g}^\chi}<1.
  \label{eq:emphatic-weight-dispersion}
\end{equation}
Let $w_i^\chi=w_g^\chi(s_i)$ and let
$w^\chi=(w_1^\chi,\ldots,w_n^\chi)$ be the vector of occurrence weights.
Its algebraic effective sample size, denoted by
$\operatorname{ESS}(w^\chi)$, is
\begin{equation}
  \operatorname{ESS}(w^\chi)
  =\frac{(\sum_iw_i^\chi)^2}{\sum_i(w_i^\chi)^2}
  =n\frac{1+v_{M,g}^\chi}{1+2v_{M,g}^\chi}>\frac n2.
  \label{eq:emphatic-weight-ess}
\end{equation}
These identities establish positivity, normalization, and controlled
dispersion of the consumer weights. Scale invariance here concerns the
consumer map; the $J$ fitting rule is not scale invariant because its
additive source term $1-\gamma$ remains fixed when raw mass is rescaled.

\subsubsection{Induced measures and their use in updating the critic.}
\label{app:emphatic-consumer-measures}
Recall that $w_i^\chi=w_g^\chi(s_i)$ is the frozen weight of current
occurrence $i$. Let $\nu_g^\chi$ be the probability measure induced by
these weights on the observed roots. Together with the unweighted
empirical measure $D_g$, it is
\begin{equation}
  D_g=\frac1n\sum_{i=1}^{n}\delta_{s_i},
  \qquad
  \nu_g^\chi=\frac1n\sum_{i=1}^{n}w_i^\chi\delta_{s_i},
  \quad \chi\in\{J,P\}.
  \label{eq:critic-consumer-root-measures}
\end{equation}
Sampling from $\nu_g^\chi$ means selecting occurrence $i$ with
probability $w_i^\chi/n$. For any real-valued function $f$ that is finite
on the observed roots,
\begin{equation}
  \E_{S\sim\nu_g^\chi}[f(S)]
  =\frac1n\sum_{i=1}^{n}w_i^\chi f(s_i).
  \label{eq:critic-emphatic-measure-expectation}
\end{equation}
Recall the normalized mass multiplier $R_g^\chi$ and shrinkage coefficient
$\alpha_g^\chi$ in Equations~\eqref{eq:emphatic-normalized-mass}
and~\eqref{eq:emphatic-shrinkage}. Define the normalized-mass probability
measure $\overline\nu_g^\chi=n^{-1}\sum_iR_g^\chi(s_i)\delta_{s_i}$.
The consumer measure then satisfies
\begin{equation}
  \nu_g^\chi=(1-\alpha_g^\chi)D_g
             +\alpha_g^\chi\overline\nu_g^\chi.
  \label{eq:emphatic-measure-mixture}
\end{equation}
The induced measure is an explicit mixture of the empirical behavior
measure and the normalized-mass measure.

Emphasis enters the successor-critic update through the distribution of
its training states. Recall from Section~\ref{app:critic-training-bank}
that a root-policy pair is chosen uniformly from
$(D_g,\mu_g)$, $(\nu_g^J,\pi_g)$, and $(\nu_g^P,\mu_g)$.
The augmented root measures add an independent root risk in RB
(Equation~\eqref{eq:critic-augmented-root-measures}). Recall that
$\widetilde\eta_g$ is the augmented state law after the first consumer
transition, and $\widetilde\zeta_g$ is the law after an additional
geometric number of target-policy transitions
(Equations~\eqref{eq:critic-first-consumer-transition}
and~\eqref{eq:critic-zeta-distribution}).
The resulting training bank contains $n_\zeta=25{,}000$ pairs
$(S_\ell,U_\ell)$, indexed by $\ell=1,\ldots,n_\zeta$, where $S_\ell$
is the retained observation and $U_\ell$ is the additional next-availability draw in
Equation~\eqref{eq:critic-paired-supply}. Recall that $V_\theta$ is a
live scalar critic with parameter vector $\theta$, and that
$\mathcal T_g^u$ denotes its target-policy Bellman backup conditional on
next availability $u$. The prediction-minus-backup fitting error is
$\varepsilon_g(\theta;s,u)=V_\theta(s)-(\mathcal T_g^uV_\theta)(s)$,
as defined in Equation~\eqref{eq:critic-training-residual}.
The empirical fitting
objective $\widehat{\mathcal J}_{V,g}(\theta)$ is
\begin{equation}
  \widehat{\mathcal J}_{V,g}(\theta)
  =\frac1{n_\zeta}\sum_{\ell=1}^{n_\zeta}
       \varepsilon_g(\theta;S_\ell,U_\ell)^2,
  \label{eq:emphatic-critic-fitting-link}
\end{equation}
which is Equation~\eqref{eq:critic-empirical-fitting-objective}.
Running the prescribed finite optimization produces each successor
member $V_{g+1,j}$, for $j=1,\ldots,5$, and their output mean $v_{g+1}$
(Equation~\eqref{eq:critic-successor}). Thus emphatic weights affect
where successor critics are fitted. Their effect is incorporated
through root selection and continuation sampling; the training-bank
loss is not multiplied by $w_i^\chi$ a second time.

The same frozen weights enter the target-return and transmission
update channels. The induced measures $\nu_g^J$ and $\nu_g^P$ also
enter the target actor's shared Fisher geometry and its two channelwise
endpoint constraints (Section~\ref{sec:ahjp-updates}). Recall from
Equation~\eqref{eq:critic-policy-root-proposal}
that $\varpi_i=(1+w_i^J+w_i^P)/(3n)$ is the occurrence probability for
their shared residual-root proposal; the corresponding importance
factors are given in Equation~\eqref{eq:critic-policy-shared-evidence}.
Finally, $\nu_g^J$ selects the heuristic anchors used above, linking
current target-return emphasis to future confidence-based transmission
emphasis.

\subsubsection{Neutral mass and weight initialization.}
\label{app:emphatic-initialization}
Both initialization routes start from neutral masses,
$M_0^J=M_0^P=1$, so $w_0^J=w_0^P=1$ and
$\nu_0^J=\nu_0^P=D_0$.

\subsubsection{Cross-generation evolution and two-generation transmission.}
\label{app:emphatic-schedule}
During generation $g$, parent masses are
evaluated on the newly collected data $\mathcal D_g$. Their consumer
weights remain frozen while successor actors, critics, and masses are
constructed; actor differentiation treats the weights as constants.
The confidence-to-transmission-mass pathway is
\begin{equation}
    q_{H,g}
    \longrightarrow (c_{g,f})_{f=1}^{16}
     \longrightarrow i_g^P
     \longrightarrow M_{g+1}^P
    \longrightarrow (w_{g+1}^P,\nu_{g+1}^P)
                       \text{ on }\mathcal D_{g+1}
     \longrightarrow\pi_{g+2}.
  \label{eq:emphatic-confidence-delay}
\end{equation}
These arrows indicate computational dependence, including the paired
comparison with $\pi_g$ used to obtain $c_{g,f}$. Interest $i_g^P$
helps fit $M_{g+1}^P$ on $\mathcal D_g$, while the update producing
$\pi_{g+1}$ uses $M_g^P$. After the joint successor commit,
$M_{g+1}^P$ is evaluated and normalized on $\mathcal D_{g+1}$; its
weights can first affect the target update producing $\pi_{g+2}$.

Weights change both because mass functions are learned and because the
behavior data used to compute their moments change. Even an unchanged
mass can induce different weights on a new data set. Generation-$g$
training labels and normalization constants are therefore not reused as
generation-$(g+1)$ consumer weights. The $J$ mass follows the same
parent-consumption/successor-fitting schedule, although its fitting
labels evolve within a generation through the private slow network.

\subsubsection{Design interpretation and scope of the claims.}
\label{app:emphatic-scope}
The two roles allocate learning effort to different state references:
target-policy continuation from behavior-distributed states for $J$,
and confidence-weighted behavior continuation for $P$. In the implemented
algorithm, these references motivate the labels rather than identify the
fitted networks with exact state-distribution ratios. The resulting
weights enter the target-update objectives, shared Fisher geometry, and
endpoint constraints; they also determine critic-training root selection
and select heuristic-confidence anchors.
Equations~\eqref{eq:emphatic-weight-properties}--\eqref{eq:emphatic-weight-ess}
prove positivity, unit mean, and bounded occurrence-weight concentration.
Equation~\eqref{eq:emphatic-confidence-delay} specifies the earliest
target update reachable through newly learned confidence-based mass.

The stationary references in Equations~\eqref{eq:emphatic-j-resolvent}
and~\eqref{eq:emphatic-p-resolvent} are population fixed points under
$\Gamma_\mu$ in Equation~\eqref{eq:emphatic-population-transition-law}.
They require fixed policies, sufficient action support, a sufficient
Markov state, a positive stationary behavior distribution, and
unrestricted exact regression; the $P$ reference also requires fixed
normalized interest. Equality of two empirical state marginals alone
does not imply these population equations. Finite trajectory data,
anchored neural approximation, finite optimization, and evolving
generations need not realize these references. The empirical conditional
regression identities remain valid on the observed successor support
for their declared frozen labels, without implying that the fitted
networks attain those targets.

These results establish controlled reweighting and the stated
computational dependencies. They do not by themselves establish exact
occupancy correction, monotone reduction of visitation mismatch, or
improved return in every generation. The ESS bound concerns algebraic
weight concentration and does not remove temporal dependence in the
data. The confidence score summarizes standardized simulated improvement
under the frozen fitted kernel in Equation~\eqref{eq:critic-fitted-kernel};
its calibration and transfer to actual returns require separate
evidence. ETD's linear expected-update stability result and ACE's
exact-gradient theorem do not automatically extend to HBPO's fitted,
shrunken, and delayed weights.

\subsection{Heuristic Guidance: Construction, Estimation, and Behavior Fitting}
\label{app:heuristic-guidance-construction}

\subsubsection{Role of heuristic guidance and frozen inputs}

At generation \(g\), heuristic guidance converts long-horizon action
comparisons under the frozen simulation model
\(\widehat{\mathcal P}_g\) into a temporary stochastic controller
\(q_{H,g}\). The parent target policy \(\pi_g\) supplies both the
continuation policy used to evaluate actions and the reference
distribution used to regularize the controller. Recall the five frozen scalar critic members \(V_{g,j}\),
\(j\in\{1,\ldots,5\}\), and their output mean
\(v_g(s)=\frac15\sum_{j=1}^{5}V_{g,j}(s)\), which supplies a control
variate.
The emphatic probability measure \(\nu_g^J\), recalled below, selects
the starting observations. The model, parent policies, reference
critic, and emphatic weights remain fixed while the controller is
constructed.

Heuristic guidance acts on the behavior policy \(\mu\), which collects
the next generation of interaction data. We therefore define a
behavior-fitting objective \(H_g\) that attracts \(\mu\) toward
\(q_{H,g}\). The target policy \(\pi_g\) determines the controller's
preferences; the parameters differentiated in \(H_g\) are those of
\(\theta^\mu\).

Throughout this subsection, \(T=1000\) is the heuristic evaluation
horizon, denoted by \(H\) in the notation table's heuristic-utility
entries. In DSIS-RB, abbreviated RB below, each simulated continuation
retains one synthetic supply-risk draw. DSIS uses the corresponding
construction without a latent-risk coordinate.

The resulting controller provides action-distribution targets for the
behavior policy. It also supplies evidence for the finite-intervention confidence used
to fit the next generation's transmission emphatic mass \(M_{g+1}^P\). Here, \emph{paired returns} compare two \(T\)-step simulations
starting from the same anchor observation (a starting state selected
as described below) and sharing their exogenous and action-sampling
random inputs, including the root risk in RB: one follows
\(q_{H,g}\) at stages \(0,\ldots,8\) and \(\pi_g\) thereafter,
whereas the other follows \(\pi_g\) throughout. Their return difference measures the
simulated gain from this finite intervention. The confidence and
interest construction is specified in Appendix~\ref{app:emphatic-confidence}.
Here we develop the action preferences and the behavior-guidance
gradient. The controller has no separately trained actor parameters.

\subsubsection{Selecting anchor states}
\label{app:h-anchor-selection}

Let \(n\) be the number of collected observation occurrences and
\(s_i\) the observation at occurrence \(i\in\{1,\ldots,n\}\).
Write \(\delta_x\) for unit point mass at observation \(x\).
Recall the empirical measure
\(D_g=n^{-1}\sum_{i=1}^{n}\delta_{s_i}\) and the positive consumer
weights \(w_i^J\) for the target-return role, normalized so that
\(n^{-1}\sum_i w_i^J=1\). The heuristic uses the already defined
measure
\begin{equation}
  \nu_g^J=\frac1n\sum_{i=1}^{n}w_i^J\delta_{s_i}.
  \label{eq:h-anchor-measure}
\end{equation}
At each generation, we sample \(16\) observation occurrences
independently with replacement. Let \(f\in\{1,\ldots,16\}\) index
the selections, \(I_{g,f}\in\{1,\ldots,n\}\) be the selected
occurrence index, and \(\bar s_{g,f}\) its anchor observation:
\begin{equation}
  \Pr(I_{g,f}=i)=\frac{w_i^J}{n},
  \qquad
  \bar s_{g,f}=s_{I_{g,f}},
  \qquad f=1,\ldots,16.
  \label{eq:h-anchor-sampling}
\end{equation}
Anchors may occur anywhere in the collected trajectories; they are not
restricted to episode starts. Repeated observations and repeated
selections retain their multiplicity. These are the same anchor
occurrences used by the confidence construction.

Every controller query, whether at an anchor or at a subsequently
simulated state, starts a new local evaluation clock at \(k=0\).
Its utility horizon is always \(T=1000\), independently of the queried
observation's position in its originating trajectory.

\subsubsection{The return, centered utility, and simulation tapes}
\label{app:heuristic-utility}

We first expand the expectation \(\mathbb E_H\) used in
Section~\ref{sec:heuristic-guidance}. For a fixed synthetic risk \(z\in[0,1/2]\), recall that
\(\widehat{\mathcal P}_{g,z}\) is the frozen joint
reward--observation kernel conditional on that risk. Let \(S_k\),
\(\mathsf A_k\), and \(R_k\) denote the observation, action, and
normalized reward at local stage \(k\), and let
\(\mathbb E_z^{s,a}\) denote expectation under the rollout law
\begin{equation}
  \begin{gathered}
    S_0=s,\qquad \mathsf A_0=a,\\
    (R_k,S_{k+1})\sim
      \widehat{\mathcal P}_{g,z}(\cdot\mid S_k,\mathsf A_k),
       \qquad 0\le k<T,\\
    \mathsf A_k\sim\pi_g(\cdot\mid S_k),\qquad 1\le k<T.
  \end{gathered}
  \label{eq:h-rollout-law}
\end{equation}
Let \(m=\operatorname{Unif}[0,1/2]\) be the synthetic-risk prior.
For any integrable rollout functional \(\mathcal F\), the RB
expectation is
\begin{equation}
  \mathbb E_H[\mathcal F\mid s,a]
  =\int_{[0,1/2]}\mathbb E_z^{s,a}[\mathcal F]\,m(dz).
  \label{eq:h-expanded-expectation}
\end{equation}
The root risk is drawn independently of the query observation and
retained for the entire continuation; it is not inferred from that
observation or redrawn at each stage.
The kernel and its risk convention are specified in Appendix~\ref{app:critic-model-construction}.
For DSIS, omit \(z\) and the integral over \(m\), and use its frozen
simulation kernel. The finite return in the main text is therefore
\begin{equation}
  G_{T,g}(s,a)
  =\mathbb E_H\!\left[\sum_{k=0}^{T-1}R_k\,\middle|\,s,a\right].
  \label{eq:h-finite-return}
\end{equation}
For RB, this includes the risk integral in
Equation~\eqref{eq:h-expanded-expectation}; for DSIS it uses the
risk-free rollout law. Write \(c_k\) for the period cost on the
simulated trajectory. Rewards retain the task's normalization
\(R_k=-c_k/495\), and the return is undiscounted.

Recall that \(\mathcal A=\{0,\ldots,8\}^2\) is the set of \(81\)
regular/expedited order pairs. Let \(\widehat G_{T,g}(s,a)\) denote
the sample estimate of this return constructed below. The centered estimated utility, denoted by
\(u_g\) in the main text and \(h_g\) in the notation table's
heuristic-utility entries, is
\begin{equation}
  u_g(s,a)=h_g(s,a)
  =\widehat G_{T,g}(s,a)
   -\sum_{a'\in\mathcal A}\pi_g(a'\mid s)\widehat G_{T,g}(s,a').
  \label{eq:h-centered-utility}
\end{equation}
Thus
\(\sum_{a\in\mathcal A}\pi_g(a\mid s)u_g(s,a)=0\).
We next construct the coupled simulations and the critic-based
estimator that supplies \(\widehat G_{T,g}\).

A \emph{simulation tape} records the random inputs used to generate a
continuation. A \emph{trajectory} records the resulting observations,
actions, and rewards. A trajectory thus depends on the starting state,
policy, and model as well as the tape. We use tapes so that alternative
first actions can be compared using common random inputs, rather than
generating an independent collection of inputs for every alternative.
This couples the simulation noise in the comparisons and permits
identical continuation work to be shared.

Under the ideal sampling law, each fixed query observation uses
\(32\) independent tapes,
indexed by \(b\in\{1,\ldots,32\}\). In RB, let \(Z_b\sim m\)
be tape \(b\)'s root risk, and let
\(U^D_{b,k},U^B_{b,k}\sim\operatorname{Unif}(0,1)\) be its demand
and next-availability inputs. For each action \(a'\in\mathcal A\),
let \(\epsilon_{b,k,a'}\) be a standard Gumbel variable, with
cumulative distribution function \(x\mapsto\exp(-e^{-x})\), and
write \(\boldsymbol\epsilon_{b,k}
=(\epsilon_{b,k,a'})_{a'\in\mathcal A}\).
These primitive inputs are mutually independent across tapes,
stages, action coordinates, and input types, including independence
from the risks. A utility-evaluation tape \(\mathfrak t_b\) is
\begin{equation}
  \mathfrak t_b
  =\left(Z_b,
      \{U^D_{b,k},U^B_{b,k},\boldsymbol\epsilon_{b,k}\}_{k=0}^{T-1}
    \right).
  \label{eq:h-simulation-tape}
\end{equation}
The action noise at \(k=0\) is unused because the first action is
forced; the final action and transition need not be sampled when
evaluating the control estimator below. DSIS omits the risk
and stochastic-availability inputs. At each distinct query state, we
enumerate all \(81\) first actions in \(\mathcal A\) and evaluate
them using these \(32\) tapes in total. Each tape supports \(81\) coupled trajectories, one for each
forced first action. Write \(S_{b,k}^{a}\) and \(\mathsf A_{b,k}^{a}\) for the
stage-\(k\) observation and action on tape \(b\)'s branch whose
forced first action is \(a\). Thus \(S_{b,0}^{a}=s\) and
\(\mathsf A_{b,0}^{a}=a\). The finite-logit categorical target
policy has \(\pi_g(a'\mid s')>0\) for every action and queried
observation. For \(1\le k<T\), branch \(a\) samples its action as
\begin{equation}
  \mathsf A_{b,k}^{a}
  =
  \operatorname*{arg\,max}_{a'\in\mathcal A}
  \left\{
    \log\pi_g(a'\mid S_{b,k}^{a})+\epsilon_{b,k,a'}
  \right\}.
  \label{eq:h-coupled-action}
\end{equation}
Demand and next availability are obtained by applying the model's
inverse cumulative distribution functions to \(U^D_{b,k}\) and
\(U^B_{b,k}\), respectively, using the branch's current observation
and, in RB, its fixed risk \(Z_b\).
The branches share the tape's risk and random inputs, but their
realized actions and successors can differ because their conditional
distributions can differ. Sharing a tape preserves the prescribed
marginal rollout law for every first action.

\subsubsection{Estimating the return with the frozen critic}
\label{app:h-control-estimator}

Let \(R\) and \(S'\) denote a one-step reward and successor
observation under \(\widehat{\mathcal P}_{g,z}\). Define
\(Q^{v,H}_{g,z,k}\) as the undiscounted control backup at local
stage \(k\), where the superscripts indicate the frozen reference
\(v_g\) and the heuristic evaluation:
\begin{equation}
  Q^{v,H}_{g,z,k}(s,a)
  =
  \begin{cases}
    \mathbb E_{\widehat{\mathcal P}_{g,z}}
      [R+v_g(S')\mid s,a],&0\le k\le T-2,\\[2pt]
    \mathbb E_{\widehat{\mathcal P}_{g,z}}
      [R\mid s,a],&k=T-1.
  \end{cases}
  \label{eq:h-control-backup}
\end{equation}
This is exactly the evaluated backup \(L_{g,z,k}^{\mathrm H}\)
listed in the notation table, with \(T=H\): it uses \(v_g\) at
nonterminal successor stages and zero control at the horizon. This quantity is distinct from the completed, discounted
action value \(Q_g^{\pi_g}\) in Equation~\eqref{eq:critic-residual-completion}.
The one-step expectations integrate the model's demand support and,
in RB, both next-availability outcomes.

For branch \(a\) on tape \(b\), let
\(\widetilde G_{T,g,Z_b}^{(b)}(s,a)\) denote the complete one-tape
control-variate estimator, defined by
\begin{equation}
    \widetilde G_{T,g,Z_b}^{(b)}(s,a)
    =Q^{v,H}_{g,Z_b,0}(s,a)
    +\sum_{k=1}^{T-1}
    \left[
      \sum_{a'\in\mathcal A}
      \pi_g(a'\mid S_{b,k}^{a})
      Q^{v,H}_{g,Z_b,k}(S_{b,k}^{a},a')
      -v_g(S_{b,k}^{a})
    \right].
  \label{eq:h-complete-estimator}
\end{equation}
The first term evaluates the forced first action. Each later term is
the target-policy average of the one-step backup, minus the reference
value at the visited observation. The sampled continuation determines
where these control residuals are evaluated.

The same conditional-expectation and telescoping argument used in the
proof of Theorem~\ref{thm:critic-residual-completion}\ yields the finite-horizon identity below.
The theorem itself concerns an infinite discounted return; here the
return is undiscounted and the terminal control contribution is zero.
Fix \(s,a,z\), and write \(\widetilde G_{T,g,z}(s,a)\) for one
generic complete tape estimator conditional on risk \(z\),
suppressing the tape and branch indices on both the estimator and
the trajectory. The tower property gives
\begin{equation}
  \begin{aligned}
    \mathbb E_z^{s,a}[\widetilde G_{T,g,z}(s,a)]
    &=\mathbb E_z^{s,a}\left[\begin{aligned}
      &R_0+v_g(S_1)\\
      &+\sum_{k=1}^{T-2}
        \{R_k+v_g(S_{k+1})-v_g(S_k)\}
       +R_{T-1}-v_g(S_{T-1})
    \end{aligned}\right]\\
    &=\mathbb E_z^{s,a}\left[\sum_{k=0}^{T-1}R_k\right].
  \end{aligned}
  \label{eq:h-finite-telescope}
\end{equation}
All reference values cancel. For each fixed starting observation, bounded actions and finite
demand support give finitely many reachable observations over this
finite horizon. Period costs and the finite-parameter frozen
networks are finite on that set, uniformly over \(z\in[0,1/2]\).
Integrating this identity against \(m(dz)\) therefore recovers
\(G_{T,g}(s,a)\).

The critic need not equal the finite-horizon value function: its role
is to supply terms whose expectations cancel. This is an identity in
expectation, so the estimate need not equal the realized reward sum on
an individual tape. Rewards already enter through
\(Q^{v,H}_{g,z,k}\); an additional sampled reward sum would count them
again.

There are two changes from the discounted critic backup. Recall that
\(\gamma=0.995\) is the critic discount factor. Before the last
period, the successor value has coefficient \(1\) instead of
\(\gamma\), because every reward in this return receives equal
weight. At the last period, no successor value is included, because
rewards after period \(T-1\) are outside the objective.
Recall the discounted backup \(L_{g,z}v_g\) and the state-transition
operator \(\widehat P_{g,z}^{a}\), evaluated on \(v_g\):
\begin{equation}
  \begin{aligned}
    L_{g,z}v_g(s,a)
    &=\mathbb E_{\widehat{\mathcal P}_{g,z}}
        [R+\gamma v_g(S')\mid s,a],\\
    (\widehat P_{g,z}^{a}v_g)(s)
    &=\mathbb E_{\widehat{\mathcal P}_{g,z}}[v_g(S')\mid s,a].
  \end{aligned}
  \label{eq:h-recalled-discounted-backup}
\end{equation}
Writing \(\mathbf 1_{\{k<T-1\}}\) for the indicator of
\(k<T-1\), the required adjustment is
\begin{equation}
  Q^{v,H}_{g,z,k}(s,a)
  =
  L_{g,z}v_g(s,a)
  +\bigl(\mathbf 1_{\{k<T-1\}}-\gamma\bigr)
     (\widehat P_{g,z}^{a}v_g)(s).
  \label{eq:h-backup-adjustment}
\end{equation}
Keeping the discounted backup unchanged would break the cancellation.
Keeping a terminal value would add
\(\mathbb E_z^{s,a}[v_g(S_T)]\) to the expected estimate.

\subsubsection{Averaging and computing action differences}
\label{app:h-centered-utility}

Average the \(32\) complete tape estimates to obtain the return
estimate used in Equation~\eqref{eq:h-centered-utility}:
\begin{equation}
  \widehat G_{T,g}(s,a)
  =\frac1{32}\sum_{b=1}^{32}
       \widetilde G_{T,g,Z_b}^{(b)}(s,a).
  \label{eq:h-mean-return}
\end{equation}
Recall that \(u_g=h_g\) is this estimate centered under
\(\pi_g(\cdot\mid s)\), rather than under a uniform action law. For a fixed query observation independent of its utility tapes,
hold the model, policy, and reference critic fixed. In the next
display, \(\mathbb E\) averages only over the \(32\) utility tapes,
including their risks in RB. The preceding expectation identity gives
\begin{equation}
    \mathbb E[\widehat G_{T,g}(s,a)]=G_{T,g}(s,a),\quad
    \mathbb E[u_g(s,a)]
    =G_{T,g}(s,a)
      -\sum_{a'}\pi_g(a'\mid s)G_{T,g}(s,a').
  \label{eq:h-fixed-query-unbiasedness}
\end{equation}
These statements concern the complete fixed-state return and utility
estimators in exact arithmetic. They do not assert conditional
unbiasedness after selecting a state using the same cached utility
estimates. Nonlinear transformations of their estimates, including
the controller below, do not inherit unbiasedness automatically.
Utilities remain in cumulative normalized-reward units, without
division by \(T\).

Only action differences are needed to construct \(u_g\). For any fixed reference action \(a_\star\in\mathcal A\),
\begin{equation}
    u_g(s,a)
    = [\widehat G_{T,g}(s,a)-\widehat G_{T,g}(s,a_\star)]
    -\sum_{a'}\pi_g(a'\mid s)
       [\widehat G_{T,g}(s,a')-\widehat G_{T,g}(s,a_\star)].
  \label{eq:h-reference-action}
\end{equation}
The implementation either accumulates tape-level differences against
the reference order pair \(a_\star=(0,0)\), or accumulates coupled
branch estimates and centers them afterwards. On the same
tape, branches reaching the same observation at the same stage have
identical future contributions under their shared inputs. Their
subsequent work can be shared, and a tail common to all branches can
be omitted. These operations preserve the centered utilities in exact
arithmetic; an accumulator with an omitted common tail need not
represent an absolute return.

The identities above use exact arithmetic and the same evaluated
frozen policy and control function throughout. On RB graphics processors, the heuristic service evaluates bfloat16
copies of the frozen actor and critics, with 32-bit floating-point
accumulation. Discounted residual evaluation uses 32-bit critic
inference and 64-bit residual algebra. Equation~\eqref{eq:h-backup-adjustment}
therefore compares backups of the same mathematical functions; it
does not assert bitwise equality between these numerical services.

\subsubsection{The regularization coefficient}
\label{app:h-regularization-scale}

The coefficient \(\tau>0\) controls how strongly the controller is
penalized for departing from \(\pi_g\): larger values give more weight
to staying near the target prior. The shared hyperparameter-selection protocol tunes a positive,
dimensionless multiplier \(\kappa\), distinct from the fitted
supply-recovery parameter \(\widehat\kappa_g^{\mathrm{rec}}\).
At initialization, let \(\bar s_{0,f}^{\mathrm{cal}}\),
\(f=1,\ldots,16\), denote calibration anchors sampled with
replacement from the initial emphatic measure \(\nu_0^J\) of
Equation~\eqref{eq:h-anchor-measure}, using a separate random-number
domain. Let \(u_0^{\mathrm{cal}}\) be their centered utility
estimates, computed with the same \(81\) forced first actions,
\(32\) tapes, and \(T=1000\) horizon. Define the calibration scale
\(S_{\mathrm{cal}}\) and set
\begin{equation}
  \begin{aligned}
    S_{\mathrm{cal}}
    &=\operatorname{median}_{f=1,\ldots,16}
      \left[
        \max_{a\in\mathcal A}u_0^{\mathrm{cal}}(\bar s_{0,f}^{\mathrm{cal}},a)
        -\min_{a\in\mathcal A}u_0^{\mathrm{cal}}(\bar s_{0,f}^{\mathrm{cal}},a)
      \right],\\
    \tau&=\kappa S_{\mathrm{cal}}.
  \end{aligned}
  \label{eq:h-tau-calibration}
\end{equation}
For these \(16\) ranges, the median is the mean of the eighth and
ninth ordered values. The construction requires
\(0<S_{\mathrm{cal}}<\infty\), hence \(0<\tau<\infty\).
The implementation rejects a zero or nonfinite calibration scale;
no positive floor is substituted. Thus \(S_{\mathrm{cal}}\) supplies a one-time scale
conversion; it is not an additional tuned parameter. The resulting
\(\tau\) stays fixed throughout the run, with no separate normalization
of later states' utilities.

\subsubsection{The Hellinger-regularized controller}
\label{app:h-derivation}

Recall that \(u_g\) is the centered utility in
Equation~\eqref{eq:h-centered-utility}, and that the positive scale
\(\tau=\kappa S_{\mathrm{cal}}\) is fixed by
Equation~\eqref{eq:h-tau-calibration}. At a fixed query observation,
write \(p=(p_a)_{a\in\mathcal A}\) for the target prior with
\(p_a=\pi_g(a\mid s)>0\), and \(u_a=u_g(s,a)\).
Let \(q=(q_a)_{a\in\mathcal A}\) be a candidate distribution for
the controller's action at \(s\), where \(q_a\) is the probability
assigned to action \(a\). The optimization ranges over the entire
simplex
\(\Delta(\mathcal A)=\{q:q_a\ge0,\ \sum_{a\in\mathcal A}q_a=1\}\),
the set of all probability distributions on the \(81\) actions.
The prior \(p\) supplies the regularization reference; no initial
candidate \(q\) is required, because the maximizing distribution is
computed by the scalar root solve derived below. The controller is
defined by
\begin{equation}
  q_{H,g}(\cdot\mid s)
  =
  \operatorname*{arg\,max}_{q\in\Delta(\mathcal A)}
  \left\{
    \sum_a q_a u_a
    -2\tau\sum_a(\sqrt{q_a}-\sqrt{p_a})^2
  \right\}.
  \label{eq:h-appendix-controller}
\end{equation}
The displayed penalty uses the unnormalized squared Hellinger
discrepancy \(\sum_a(\sqrt{q_a}-\sqrt{p_a})^2\); the convention
with a factor \(1/2\) gives half this value.

Since \(p\) and \(q\) have unit mass, denote the same objective by
\(\Phi(q)\) and expand it as
\begin{equation}
  \Phi(q)=\sum_a q_a u_a+4\tau\sum_a\sqrt{p_aq_a}-4\tau.
  \label{eq:h-controller-objective}
\end{equation}
At an interior point \(q_a>0\) for all \(a\), its Hessian entries
are
\begin{equation}
  \frac{\partial^2\Phi}{\partial q_a^2}
  =-\tau\sqrt{p_a}\,q_a^{-3/2}<0,
  \qquad
  \frac{\partial^2\Phi}{\partial q_a\partial q_{a'}}=0
  \quad(a\ne a').
  \label{eq:h-controller-hessian}
\end{equation}
The objective is strictly concave. Its maximizer has full support: if \(q_a=0\), transfer a small mass
\(\varepsilon>0\) from any positive coordinate to action \(a\).
Because \(p_a>0\), the gain in the square-root term is of order
\(\sqrt{\varepsilon}\), whereas the utility change and the loss at
the donating coordinate are of order \(\varepsilon\). Compactness gives existence, and strict concavity
gives uniqueness.

Introducing a multiplier \(\lambda_g(s)\) for the unit-mass constraint
yields
\begin{equation}
  u_a+2\tau\sqrt{\frac{p_a}{q_a}}-\lambda_g(s)=0.
  \label{eq:h-controller-stationarity}
\end{equation}
Therefore,
\begin{equation}
    q_{H,g}(a\mid s)=
    \pi_g(a\mid s)
    \left(\frac{2\tau}{\lambda_g(s)-u_g(s,a)}\right)^2
  ,
  \qquad \lambda_g(s)>\max_a u_g(s,a),
  \label{eq:h-controller-solution}
\end{equation}
where \(\lambda_g(s)\) is determined by
\begin{equation}
  \sum_a \pi_g(a\mid s)
    \left(\frac{2\tau}{\lambda_g(s)-u_g(s,a)}\right)^2=1.
  \label{eq:h-controller-root}
\end{equation}
The left-hand side is continuous and strictly decreasing in
\(\lambda_g(s)\), tends to infinity at the lower boundary, and tends
to zero as \(\lambda_g(s)\to\infty\). Hence a unique scalar root
determines the entire action distribution.

For a bounded numerical bracket, let \(u_{\max}\) be the largest
utility, \(p_{\max}\) the total prior mass of all maximizing
actions (not the largest prior probability), and \(\delta\) the
positive multiplier gap:
\begin{equation}
  u_{\max}=\max_a u_a,\qquad
  p_{\max}=\sum_{a:u_a=u_{\max}}p_a,\qquad
  \delta=\lambda_g(s)-u_{\max}.
  \label{eq:h-root-bracket-notation}
\end{equation}
At the normalization root, the maximizing actions contribute
\((2\tau/\delta)^2p_{\max}\), and every denominator is at least
\(\delta\). Hence
\((2\tau/\delta)^2p_{\max}\le1\le(2\tau/\delta)^2\), giving
\begin{equation}
  2\tau\sqrt{p_{\max}}\le\delta\le2\tau.
  \label{eq:h-root-bracket}
\end{equation}
The implementation uses \(64\) bisection iterations in
\(\log\delta\), retaining log prior probabilities. This solves for one scalar normalizer per query; the resulting
probabilities satisfy the controller formula up to floating-point
and root-solve error, without adding a probability floor.

Since \(q=p\) is feasible and \(\sum_a p_a u_a=0\), optimality
of the exact controller gives
\begin{equation}
  \sum_{a\in\mathcal A}q_{H,g}(a\mid s)u_a
  \ge 2\tau\sum_{a\in\mathcal A}
    \bigl(\sqrt{q_{H,g}(a\mid s)}-\sqrt{p_a}\bigr)^2
  \ge0.
  \label{eq:h-estimated-improvement}
\end{equation}
This is a property of the estimated first-action replacement objective
under the frozen model. It neither guarantees improvement in the exact model return after
Monte Carlo action selection nor establishes improvement from
repeatedly executing the controller or in the real environment.

\subsubsection{Selecting fitting locations through short controller rollouts}
\label{app:h-fitting-locations}

Recall the \(16\) anchor occurrences \(\bar s_{g,f}\) from
Equation~\eqref{eq:h-anchor-sampling}. From each, generate \(8\)
outer model trajectories under the controller. The superscript
\(\mathrm O\) labels these outer trajectories: let
\(S^{\mathrm O}_{g,f,b,k}\) and \(\mathsf A^{\mathrm O}_{g,f,b,k}\)
denote their stage-\(k\) observations and actions, with
\(b\in\{1,\ldots,8\}\) indexing outer paths and
\(S^{\mathrm O}_{g,f,b,0}=\bar s_{g,f}\).
In RB, each path draws a risk \(Z^{\mathrm O}_{g,f,b}\) independently
from \(m=\operatorname{Unif}[0,1/2]\), the prior recalled in
Equation~\eqref{eq:h-expanded-expectation}, and retains it throughout.
At stages \(k=0,\ldots,8\), sample
\begin{equation}
  \mathsf A^{\mathrm O}_{g,f,b,k}
  \sim q_{H,g}(\cdot\mid S^{\mathrm O}_{g,f,b,k})
  \label{eq:h-outer-action}
\end{equation}
and then sample the reward and successor from
\(\widehat{\mathcal P}_{g,Z^{\mathrm O}_{g,f,b}}
(\cdot\mid S^{\mathrm O}_{g,f,b,k},\mathsf A^{\mathrm O}_{g,f,b,k})\)
in RB, or from the risk-free model in DSIS.
Let \(\xi_g\) be the empirical fitting measure of the pre-action
observations at these nine stages. Recalling that \(\delta_x\)
denotes unit point mass at \(x\), set
\begin{equation}
    \xi_g=
    \frac1{16\cdot8\cdot9}
    \sum_{f=1}^{16}\sum_{b=1}^{8}\sum_{k=0}^{8}
      \delta_{S^{\mathrm O}_{g,f,b,k}}.
  \label{eq:h-fitting-measure}
\end{equation}
This panel contains exactly \(1152\) observation occurrences,
including repetitions. Each occurrence receives equal fitting weight;
the emphatic weights have already entered through anchor selection.

At every retained observation, the fitting target is the full
distribution \(q_{H,g}(\cdot\mid s)\), obtained from a utility query
with a new local horizon of \(1000\) periods. The nine-stage rollout
selects fitting locations without reducing the utility horizon at
later stages.

Utility estimation and outer trajectory generation use separate
random-number domains. In RB, an inner controller query samples its
own utility-tape risks and does not read the outer trajectory's latent
risk, so the controller remains a function of the visible observation.
Repeated queries at the same observation reuse the generation's cached
estimate. Separate outer paths use separately addressed random inputs,
including when their anchors coincide. Thus \(1152\) counts fitting
occurrences, not necessarily distinct utility queries; observations
along a path are also temporally dependent.

\subsubsection{Deriving the behavior-guidance gradient}
\label{app:h-behavior-gradient}

Recall that \(\xi_g\) is the equal-weight empirical measure of the
\(1152\) pre-action observations \(S^{\mathrm O}_{g,f,b,k}\) in
Equation~\eqref{eq:h-fitting-measure}; \(\mathrm O\) denotes the
outer controller trajectories. Let \(\theta^\mu\) parameterize the
candidate behavior policy \(\mu_{\theta^\mu}\), and recall the
parent policy \(\mu_g=\mu_{\theta_g^\mu}\) at parameters
\(\theta_g^\mu\). Assume that the candidate probabilities are
strictly positive and differentiable in a neighborhood of those
parent parameters, as for the finite-logit categorical actor.
For two action distributions \(p=(p_a)\) and \(q=(q_a)\), define
the categorical Fisher--Rao distance \(d_{\mathrm{FR}}\) by
\begin{equation}
  d_{\mathrm{FR}}(p,q)
  =2\arccos\left(\sum_{a\in\mathcal A}\sqrt{p_aq_a}\right).
  \label{eq:h-fisher-rao-distance}
\end{equation}
With \(\xi_g\) and all controller targets fixed, define the
behavior-fitting objective
\begin{equation}
  H_g(\mu_{\theta^\mu})
  =-\frac12\,\mathbb E_{s\sim\xi_g}\!\left[
    d_{\mathrm{FR}}^2\bigl(
      \mu_{\theta^\mu}(\cdot\mid s),q_{H,g}(\cdot\mid s)\bigr)
    \right].
  \label{eq:h-behavior-objective}
\end{equation}
Maximizing \(H_g\) reduces the distance between the behavior policy
and the controller at the retained observations. Define its behavior-guidance gradient \(g_{H,g}\) at the parent
behavior parameters by
\begin{equation}
  \begin{aligned}
    g_{H,g}
    &=
    \left.\nabla_{\theta^\mu}H_g(\mu_{\theta^\mu})
    \right|_{\theta^\mu=\theta_g^\mu}\\
    &=
    -2\left.\nabla_{\theta^\mu}
    \mathbb E_{s\sim\xi_g}
    \left[
      \arccos^2\left(
        \sum_a
          \sqrt{\mu_{\theta^\mu}(a\mid s)q_{H,g}(a\mid s)}
      \right)
    \right]\right|_{\theta^\mu=\theta_g^\mu},
  \end{aligned}
  \label{eq:h-behavior-gradient}
\end{equation}
recovering Equation~\eqref{eq:h-guidance-gradient}.

For an explicit expression, define the affinity \(\mathcal B_g(s)\),
angle \(\varphi_g(s)\), and behavior-policy score vector
\(\operatorname{sc}_g^\mu(s,a)\) at the parent parameters:
\begin{equation}
  \begin{aligned}
    \mathcal B_g(s)
    &=\sum_a\sqrt{\mu_g(a\mid s)q_{H,g}(a\mid s)},&
    \varphi_g(s)&=\arccos\mathcal B_g(s),\\
    \operatorname{sc}_g^\mu(s,a)
    &=\left.\nabla_{\theta^\mu}
      \log\mu_{\theta^\mu}(a\mid s)
      \right|_{\theta^\mu=\theta_g^\mu}.
  \end{aligned}
  \label{eq:h-affinity-and-score}
\end{equation}
Let \(\mathcal B_{\theta^\mu}(s)\) denote the same affinity with
\(\mu_g\) replaced by \(\mu_{\theta^\mu}\). Differentiation gives
\begin{equation}
  \left.\nabla_{\theta^\mu}
      \mathcal B_{\theta^\mu}(s)\right|_{\theta^\mu=\theta_g^\mu}
  =
  \frac12\sum_a
    \sqrt{\mu_g(a\mid s)q_{H,g}(a\mid s)}
    \,\operatorname{sc}_g^\mu(s,a).
  \label{eq:h-affinity-gradient}
\end{equation}
Apply the chain rule to
\(-2\arccos^2\mathcal B_{\theta^\mu}(s)\) and use the score
identity
\(\sum_a\mu_g(a\mid s)\operatorname{sc}_g^\mu(s,a)=0\), which
follows by differentiating \(\sum_a\mu_{\theta^\mu}(a\mid s)=1\).
This gives
\begin{equation}
    g_{H,g}
    =
    \mathbb E_{s\sim\xi_g}
    \left[
      \frac{2\varphi_g(s)}{\sin\varphi_g(s)}
      \sum_a
      \left(
        \sqrt{\mu_g(a\mid s)q_{H,g}(a\mid s)}
        -\mathcal B_g(s)\mu_g(a\mid s)
      \right)
      \operatorname{sc}_g^\mu(s,a)
    \right]
  .
  \label{eq:h-explicit-gradient}
\end{equation}
The expectation is the arithmetic average over the \(1152\) retained
occurrences, and the action sum includes all \(81\) actions.
The subtracted term \(\mathcal B_g(s)\mu_g(a\mid s)\) contributes
zero after the score sum. It is a tangent-space centering of the
same Fisher--Rao gradient; there is no additional action-importance
ratio or state-weight correction in this fitting objective.

At \(\mu_g(\cdot\mid s)=q_{H,g}(\cdot\mid s)\), use
\(\lim_{\varphi\to0}\varphi/\sin\varphi=1\); the centered
action coefficients are zero, so the contribution is zero.
For stable evaluation, let \(\mathbf r_g(s)\) be the residual
between the square-root controller vector and its projection onto
the square-root behavior vector. Square roots below act
coordinatewise, \(\|\cdot\|_2\) is the Euclidean norm, and
\(\operatorname{atan2}(y,x)\) is the two-argument angle function:
\begin{equation}
  \begin{aligned}
    \mathbf r_g(s)
    &=\sqrt{q_{H,g}(\cdot\mid s)}
       -\mathcal B_g(s)\sqrt{\mu_g(\cdot\mid s)},\\
    \sin\varphi_g(s)&=\|\mathbf r_g(s)\|_2,\\
    \varphi_g(s)
    &=\operatorname{atan2}
      \bigl(\|\mathbf r_g(s)\|_2,\mathcal B_g(s)\bigr).
  \end{aligned}
  \label{eq:h-stable-angle}
\end{equation}
This evaluates the same gradient without directly differentiating an
inverse cosine at coincident distributions.

During differentiation, the sampled observations, controller targets,
utility estimates, and parent simulation components remain fixed.
No gradient passes through action selection, model trajectories, the
critic, or the controller's scalar root. Recall that \(\tau=\kappa S_{\mathrm{cal}}\), with tuned multiplier
\(\kappa\) and initialization scale \(S_{\mathrm{cal}}\), is fixed
by Equation~\eqref{eq:h-tau-calibration}. It affects \(g_{H,g}\)
through the controller and its induced fitting locations, and adds
no multiplicative factor to this gradient. The resulting \(g_{H,g}\) enters the behavior update through
the coupled-update rule defined in the main text.

\subsection{Delayed Heuristic--Behavior--Target Coupling and Policy Updates}
\label{app:policy-updates}
\label{app:actor-derivation}

This subsection assembles the shared advantage evidence, the direct
behavior and target objectives, the transmission objective, and the fused
finite actor updates, in that order. All updates use the same frozen
generation-$g$ parents: behavior policy $\mu_g$, target policy $\pi_g$,
reference critic $v_g$, simulation model, and normalized consumer weights
$w_i^J,w_i^P$ defined below. The simulation model is the public kernel in
DSIS and the fitted kernel in DSIS-RB. Once generated, residual samples,
controller targets, and heuristic fitting locations are also held fixed
during actor differentiation. In particular, the target successor uses
$\mu_g$, not the simultaneously constructed $\mu_{g+1}$: guidance that
changes behavior in generation $g$ can first enter the discrepancy used
to construct $\pi_{g+2}$. The corresponding delay through the transmission
mass is specified in Appendix~\ref{app:emphatic-schedule}.

The behavior actor combines the advantage objective $A$ with the heuristic
objective $H_g$; the target actor combines its advantage objective $J$ with
the transmission objective $P$. Recall that $H_g$ is the negative half
squared Fisher--Rao distance to the frozen heuristic controller, averaged
over its fitting observations, and write $g_H=g_{H,g}$ for its parent
gradient from Equations~\eqref{eq:h-behavior-objective}
and~\eqref{eq:h-explicit-gradient}. We differentiate the candidate behavior
policy in $H_g$, holding the controller and its fitting measure fixed. The results below
concern these frozen policy-update objectives and their induced policy
changes.

\subsubsection{Constructing the shared advantage face}
\label{app:critic-policy-evidence}
\label{app:critic-policy-error}

We first specify the exact, base, and sampled advantages used by all three
advantage channels. Recall that the generation batch contains $n=25{,}000$
observation occurrences $s_i$, indexed by $i=1,\ldots,n$, and that
$D_g=n^{-1}\sum_i\delta_{s_i}$ is their empirical measure, with
$\delta_{s_i}$ the unit point mass. The frozen weights satisfy
$w_i^J,w_i^P>0$ and $n^{-1}\sum_iw_i^J=n^{-1}\sum_iw_i^P=1$;
their weighted measures are $\nu_g^\chi=n^{-1}\sum_iw_i^\chi\delta_{s_i}$
for $\chi\in\{J,P\}$. All action sums below range over the finite action
set $\mathcal A$. Write $\pi_i(a)=\pi_g(a\mid s_i)$ and
$\mu_i(a)=\mu_g(a\mid s_i)$. For action-indexed arrays $x,y$, use
$\langle x,y\rangle=\sum_{a\in\mathcal A}x(a)y(a)$, componentwise if
one array is vector-valued. Recall the exact discounted model action value
$Q_g^{\pi_g}$ and its one-step base $Q_g^0$ from
Equations~\eqref{eq:critic-model-action-value} and~\eqref{eq:critic-dense-base}.
Their centered advantages are
\begin{equation}
  \begin{aligned}
    \mathbf a_i(a)
    &=A_g^{\mathrm{adv}}(s_i,a)
      =Q_g^{\pi_g}(s_i,a)-\sum_b\pi_i(b)Q_g^{\pi_g}(s_i,b),\\
    \mathbf a_i^0(a)
    &=A_g^{\mathrm{base}}(s_i,a)
      =Q_g^0(s_i,a)-\sum_b\pi_i(b)Q_g^0(s_i,b).
  \end{aligned}
  \label{eq:critic-exact-and-base-advantages}
\end{equation}
All three channels use target-policy centering. The completed evidence
evaluates a first action followed by $\pi_g$, while the one-step base
approximates that continuation through $v_g$, including when the evidence
is used to update behavior.

Centering does not generally remove the critic's continuation error.
Recall that $\gamma\in(0,1)$ is the critic discount, $Z\sim m$ is a
synthetic risk with prior $m=\operatorname{Unif}[0,1/2]$, drawn
independently at the query root and retained along
the continuation, and $\Delta_{g,z}=V_{g,z}^{\pi_g}-v_g$ is the
conditional model-value error. Here $V_{g,z}^{\pi_g}$ is the exact
discounted value at fixed risk $z$, and
$\widehat P_{g,z}^{a},\widehat P_{g,z}^{\pi_g}$ are the one-step
observation-transition operators at that risk, with the action fixed to
$a$ or averaged under $\pi_g$, respectively. In DSIS the risk variable
is omitted. Theorem~\ref{thm:critic-residual-completion}, followed by
subtraction of the $\pi_i$-weighted action mean, gives
\begin{equation}
  \mathbf a_i(a)-\mathbf a_i^0(a)
  =\gamma\mathbb E_{Z\sim m}\!\left[
    \bigl((\widehat P_{g,Z}^{a}-\widehat P_{g,Z}^{\pi_g})
      \Delta_{g,Z}\bigr)(s_i)
  \right].
  \label{eq:critic-centered-error}
\end{equation}
Here the operators act on the function $\Delta_{g,Z}$ before evaluation
at $s_i$. Consequently, the base recovers the entire centered action
vector exactly if and only if the omitted continuation
$Q_g^{\pi_g}(s_i,a)-Q_g^0(s_i,a)$ is constant across actions. On-policy
sampling does not impose this condition. The shared advantage face
therefore includes the centered residual correction derived earlier.

The base is evaluated at every batch occurrence. For residual completion,
the implementation draws $K_{\mathrm{res}}=128$ occurrences with replacement.
Let $\varpi_i$ denote the probability of occurrence $i$ and
$\mathsf I_k$ the occurrence index in draw $k$; the draws are independent
and identically distributed (iid) conditional on the frozen inputs:
\begin{equation}
  \varpi_i=\frac{1+w_i^J+w_i^P}{3n},\qquad
  \mathsf I_k\overset{\mathrm{iid}}{\sim}
  \operatorname{Categorical}(\varpi_1,\ldots,\varpi_n),
  \quad k=1,\ldots,K_{\mathrm{res}}.
  \label{eq:critic-policy-root-proposal}
\end{equation}
Recall that $\widehat A_g^{(k)}$ denotes the sampled discounted
target-policy advantage from Equation~\eqref{eq:critic-sampled-advantage},
using the independent bank of $32$ continuation tapes assigned to draw
$k$. It is centered under $\pi_g$ at the selected occurrence.
Repeated selections of the same occurrence receive distinct banks.
Define the centered residual correction by
\begin{equation}
  \Delta\widehat{\mathbf a}^{(k)}(a)
  =\widehat A_g^{(k)}(s_{\mathsf I_k},a)
     -\mathbf a_{\mathsf I_k}^0(a).
  \label{eq:coupling-sampled-correction}
\end{equation}
Let $\chi\in\{0,J,P\}$ index the behavior-advantage, target-advantage,
and transmission roles, respectively, and set $w_i^0=1$.
For finite actionwise coefficients $\ell=(\ell_i(a))_{i,a}$, define the
sampled linear evidence functional
\begin{equation}
    \widehat\Lambda_g^\chi[\ell]
    =\frac1n\sum_{i=1}^n
       w_i^\chi\langle\ell_i,\mathbf a_i^0\rangle +\frac1{K_{\mathrm{res}}}
       \sum_{k=1}^{K_{\mathrm{res}}}
       \frac{w_{\mathsf I_k}^\chi}{n\varpi_{\mathsf I_k}}
       \left\langle\ell_{\mathsf I_k},
                         \Delta\widehat{\mathbf a}^{(k)}\right\rangle.
  \label{eq:critic-policy-shared-evidence}
\end{equation}
All three channels share the same root draws and continuation banks.
Since each weight vector has empirical mean one,
$\sum_i\varpi_i=1$ and $0<w_i^\chi/(n\varpi_i)\le3$ in exact arithmetic.
In floating-point arithmetic, the implementation normalizes
$1+w_i^J+w_i^P$ by its computed sum and reuses that same probability
vector for root sampling and inverse-proposal factors.
For vector-valued coefficients, the functional acts componentwise.

Define the occurrence-level estimated advantage vector
$\widehat{\mathbf a}_i$ by the following algebraic representation of
the main-text evidence:
\begin{equation}
  \widehat{\mathbf a}_i
  =\mathbf a_i^0+
    \frac1{K_{\mathrm{res}}\varpi_i}
    \sum_{k:\mathsf I_k=i}\Delta\widehat{\mathbf a}^{(k)}.
  \label{eq:coupling-occurrence-evidence}
\end{equation}
An empty sum is zero, so an occurrence not selected for residual
completion retains its base advantage. Therefore,
\begin{equation}
  \widehat\Lambda_g^\chi[\ell]
    =\frac1n\sum_iw_i^\chi
       \langle\ell_i,\widehat{\mathbf a}_i\rangle,
  \qquad
  \langle\pi_i,\widehat{\mathbf a}_i\rangle=0.
  \label{eq:coupling-face-representation}
\end{equation}
This identity does not require materializing a completed action-value
table at all roots: the implementation retains the dense base and sampled
corrections separately. Distinct occurrences of the same observation may
have different realized corrections. The main-text abbreviations are
$\Lambda_A=\widehat\Lambda_g^0$,
$\Lambda_J=\widehat\Lambda_g^J$, and
$\Lambda_P=\widehat\Lambda_g^P$.

\noindent\begin{minipage}{\linewidth}
\begin{theorem}[Conditional mean of the shared advantage evidence]
\label{thm:coupling-shared-evidence}
Let $\mathscr G_g$ be the sigma-field generated by the frozen generation
inputs before residual-root and continuation-bank sampling. Expectations
below are taken under the residual-sampling law conditional on these
inputs. Assume the integrability conditions of Theorem~\ref{thm:critic-residual-completion}.
Conditionally on $\mathscr G_g$, draw roots with
Equation~\eqref{eq:critic-policy-root-proposal} and use the unbiased
centered continuation estimators from
Equations~\eqref{eq:critic-geometric-unbiasedness}
and~\eqref{eq:critic-sampled-advantage}. In exact arithmetic, for every finite
$\mathscr G_g$-measurable coefficient array $\ell$ and every role
$\chi\in\{0,J,P\}$,
\begin{equation}
  \mathbb E[\widehat{\mathbf a}_i\mid\mathscr G_g]=\mathbf a_i,
  \qquad
  \mathbb E[\widehat\Lambda_g^\chi[\ell]\mid\mathscr G_g]
  =\frac1n\sum_iw_i^\chi\langle\ell_i,\mathbf a_i\rangle.
  \label{eq:coupling-fixed-query-mean}
\end{equation}
\end{theorem}
\end{minipage}
\begin{proof}
For a root draw $k$, let $\mathbf 1\{\mathsf I_k=i\}$ be the indicator
of selecting occurrence $i$. Conditional unbiasedness of the
continuation bank gives
\[
 \mathbb E[\mathbf 1\{\mathsf I_k=i\}
       \Delta\widehat{\mathbf a}^{(k)}\mid\mathscr G_g]
 =\varpi_i(\mathbf a_i-\mathbf a_i^0),
\]
Substitution in
Equation~\eqref{eq:coupling-occurrence-evidence} cancels the proposal
factor and proves the first identity. The second follows from
Equation~\eqref{eq:coupling-face-representation}, conditional linearity,
and measurability of the frozen weights and coefficients.
\end{proof}
The result applies componentwise to frozen parent-policy score
coefficients. Nonlinear transformations,
gradient norms, fusion coefficients, and adaptively selected candidates
do not inherit this unbiasedness statement.

\paragraph{Further consequences of critic evaluation error.}
The following consequence also applies to fixed query observations
outside the empirical batch.
For such observations, $A_g^{\mathrm{adv}}(s,\cdot)$ and
$A_g^{\mathrm{base}}(s,\cdot)$ denote the
$\pi_g(\cdot\mid s)$-centered versions of $Q_g^{\pi_g}(s,\cdot)$
and $Q_g^0(s,\cdot)$, respectively, as in
Equation~\eqref{eq:critic-exact-and-base-advantages}.
\begin{corollary}[Action-dependent evaluation error survives centering]
\label{cor:critic-action-error}
Under Theorem~\ref{thm:critic-residual-completion}, fix an observation
$s$ satisfying its assumptions. The one-step base recovers all centered
action values at $s$ exactly if and only if
$Q_g^{\pi_g}(s,a)-Q_g^0(s,a)$ is constant across actions.
At batch occurrences, Equation~\eqref{eq:critic-centered-error}
gives the corresponding centered correction.
\end{corollary}
\begin{proof}
Subtract the $\pi_g$-weighted mean over actions from
\eqref{eq:critic-omitted-continuation-error}. Writing $\Delta_g^Q(s,a)=Q_g^{\pi_g}(s,a)-Q_g^0(s,a)$ for the actionwise
error, equality of the centered vectors means
$\Delta_g^Q(s,a)=\sum_{a'\in\mathcal A}\pi_g(a'\mid s)\Delta_g^Q(s,a')$ for every action $a$.
The right-hand side is independent of $a$, which proves necessity;
a constant vector also satisfies this equality, proving sufficiency.
\end{proof}

On-policy sampling does not impose this constancy condition or make
$\Delta_{g,z}$ vanish. To express its consequence for a policy update,
let $\upsilon$ be any fixed state-sampling distribution, including one
obtained on-policy. Let $\vartheta$ denote actor parameters, distinct
from critic parameters $\theta$, and let $\vartheta_g$ be the
parent actor parameter, so that $\pi_g=\pi_{\vartheta_g}$.
For a frozen scalar action-evaluation function $f$, define the local actor surrogate $\mathcal J_g^{\mathrm{actor}}$ and
the parent-policy score vector $\operatorname{sc}_g$ by
\begin{equation}
  \mathcal J_g^{\mathrm{actor}}(\vartheta;f,\upsilon)
  =\E_{S\sim\upsilon}\!\left[\sum_a\pi_\vartheta(a\mid S)f(S,a)\right],
  \qquad
  \operatorname{sc}_g(s,a)=
  \left.\nabla_\vartheta\log\pi_\vartheta(a\mid s)\right|_{\vartheta=\vartheta_g}.
  \label{eq:critic-actor-surrogate}
\end{equation}
In the expectation below, $S\sim\upsilon$ and
$\mathsf A\mid S\sim\pi_g(\cdot\mid S)$; the shorthand $Q$ in
$\mathcal J_g^{\mathrm{actor}}$ denotes the frozen function $Q_g^{\pi_g}$.
Assume the policy is differentiable with positive action probabilities,
the surrogate expectations are finite, and differentiation can be
interchanged with the state expectation. Also assume absolute score
integrability, where $\|\cdot\|_2$ denotes the Euclidean norm:
\begin{equation}
  \E_{\upsilon,\pi_g}\!\left[
    \|\operatorname{sc}_g(S,\mathsf A)\|_2
    \max_{a'\in\mathcal A}\bigl(|Q_g^{\pi_g}(S,a')|+|Q_g^0(S,a')|\bigr)
  \right]<\infty.
  \label{eq:critic-score-integrability}
\end{equation}
For a finite empirical state measure these integrability and
interchange conditions hold whenever the pointwise scores and action
values are finite. Then
\begin{equation}
  \left.\nabla_\vartheta\mathcal J_g^{\mathrm{actor}}(\vartheta;Q,\upsilon)\right|_{\vartheta_g}
  -\left.\nabla_\vartheta\mathcal J_g^{\mathrm{actor}}(\vartheta;Q_g^0,\upsilon)\right|_{\vartheta_g}
  =\E_{\upsilon,\pi_g}[\operatorname{sc}_g(S,\mathsf A)(A_g^{\mathrm{adv}}(S,\mathsf A)-A_g^{\mathrm{base}}(S,\mathsf A))].
  \label{eq:critic-actor-score-error}
\end{equation}
This follows from
$\nabla_\vartheta\pi_\vartheta=\pi_\vartheta\nabla_\vartheta\log\pi_\vartheta$
and $\sum_a\nabla_\vartheta\pi_\vartheta(a\mid s)=0$, the same baseline
cancellation used in policy-gradient methods \citep{sutton1999policy}.
Residual completion supplies the missing population score term
without requiring the scalar critics to have converged. This is a
statement about the frozen, fixed-state surrogate. Interpreting it
as the gradient of a trajectory-return objective additionally requires
that objective's correct state and risk occupancy measure.

\emph{Why the supply-chain structure makes the correction relevant.}
A regular order changes the pipeline immediately, but its inventory
and cost consequences occur after the four-period lead time.
For example, with $v_g\equiv0$ and fixed expedited quantity,
\begin{equation}
  Q_g^0(s,(q^r,q^e))
  =\int\E_z^{s,(q^r,q^e)}[R_0]m(dz)
  \quad\text{is independent of }q^r.
  \label{eq:critic-delayed-order-example}
\end{equation}
Immediate cost contains no regular-order charge, and the current
regular quantity does not change end-of-period inventory. When supply
is available, however, different regular quantities enter different
successor pipelines and can produce different future holding and
backlog costs. The residual continuation recovers precisely the
long-term action differences omitted by this base. An approximate
nonzero critic can already capture part of those differences; the
same identity completes the part it misses. This issue exists in DSIS
as well as DSIS-RB, regardless of whether data collection is on-policy.

\subsubsection{Direct advantage learning for behavior and target}
\label{app:coupling-direct-gradients}

Suppress generation indices on the parent distributions and write
$m_{\theta^\mu}(a\mid s)=\mu_{\theta^\mu}(a\mid s)$ and
$p_{\theta^\pi}(a\mid s)=\pi_{\theta^\pi}(a\mid s)$ for the two
parameterized candidate policies. Let $\theta_g^\mu,\theta_g^\pi$
be their frozen parent parameter vectors, so
$m_{\theta_g^\mu}=\mu$ and $p_{\theta_g^\pi}=\pi$.
The symbols $m,p$ in the local objectives denote candidate distributions;
this use of $m$ is distinct from the synthetic-risk prior above.
Write $m_i(a)=m_{\theta^\mu}(a\mid s_i)$ and
$p_i(a)=p_{\theta^\pi}(a\mid s_i)$ for their occurrence-level rows.
Assume positive, differentiable candidate probabilities near the parents
and finite values and derivatives at the retained observations.
Recall that $\tau=\kappa S_{\mathrm{cal}}>0$ is the fixed run-level
reward scale from Equation~\eqref{eq:h-tau-calibration}; $\kappa>0$
is the tuned dimensionless multiplier and $S_{\mathrm{cal}}>0$ is the
initialization utility-range scale.

The \emph{score function} of a policy is its log-probability gradient.
The parent behavior and target scores are
\begin{equation}
  \begin{aligned}
    \operatorname{sc}_g^\mu(s,a)
    &=\left.\nabla_{\theta^\mu}\log m_{\theta^\mu}(a\mid s)
      \right|_{\theta_g^\mu},\\
    \operatorname{sc}_g(s,a)
    &=\left.\nabla_{\theta^\pi}\log p_{\theta^\pi}(a\mid s)
      \right|_{\theta_g^\pi}.
  \end{aligned}
  \label{eq:coupling-parent-scores}
\end{equation}
The target score is the $\operatorname{sc}_g$ of the notation table, with
$\vartheta_g=\theta_g^\pi$. Write
$\operatorname{sc}_i^\mu(a)=\operatorname{sc}_g^\mu(s_i,a)$ and
$\operatorname{sc}_i(a)=\operatorname{sc}_g(s_i,a)$.

Recall that $\widehat\Lambda_g^\chi$ is the sampled linear advantage
functional in Equation~\eqref{eq:critic-policy-shared-evidence}; role
$0$ uses unit weights and role $J$ uses $w_i^J$. The two direct surfaces
are
\begin{equation}
  A(m)=\frac1\tau\widehat\Lambda_g^0[m-\mu],
  \qquad
  J(p)=\widehat\Lambda_g^J[p-\pi].
  \label{eq:coupling-direct-surfaces}
\end{equation}
During differentiation, the subtracted parents $\mu,\pi$ and the
weights, base advantages, sampled root indices, and continuation
corrections defining $\widehat\Lambda_g^\chi$ remain fixed. In particular, the
appearance of $\pi_g$ in the centering and continuation estimator does
not create an additional derivative. Only the candidate distribution
depends on the differentiated actor parameters.

\begin{theorem}[Parent gradients of the direct advantage surfaces]
\label{thm:coupling-direct-gradients}
Under the positivity and differentiability assumptions above, define
$g_A=\left.\nabla_{\theta^\mu}A(m_{\theta^\mu})\right|_{\theta_g^\mu}$
and $g_J=\left.\nabla_{\theta^\pi}(J(p_{\theta^\pi})/\tau)
\right|_{\theta_g^\pi}$. Inside the functionals, the arrays
$\operatorname{sc}^{\mu},\operatorname{sc}$ have rows
$\operatorname{sc}_i^\mu,\operatorname{sc}_i$; multiplication by a
policy probability is actionwise. With all evidence and parents frozen,
\begin{equation}
  g_A=\frac1\tau\widehat\Lambda_g^0
       [\mu\,\operatorname{sc}^{\mu}]
      =\frac1{n\tau}\sum_{i,a}
        \mu_i(a)\widehat{\mathbf a}_i(a)
        \operatorname{sc}_i^\mu(a),
  \label{eq:coupling-gA}
\end{equation}
and
\begin{equation}
  g_J=\frac1\tau\widehat\Lambda_g^J[\pi\,\operatorname{sc}]
      =\frac1{n\tau}\sum_{i,a}w_i^J\pi_i(a)
        \widehat{\mathbf a}_i(a)\operatorname{sc}_i(a).
  \label{eq:coupling-gJ}
\end{equation}
\end{theorem}

\begin{proof}
For behavior, linearity and the chain rule give
\begin{equation}
  \begin{aligned}
    \nabla_{\theta^\mu}A(m_{\theta^\mu})
    &=\frac1{n\tau}\sum_{i,a}\widehat{\mathbf a}_i(a)
       \nabla_{\theta^\mu}m_{\theta^\mu}(a\mid s_i)\\
    &=\frac1{n\tau}\sum_{i,a}\widehat{\mathbf a}_i(a)
       m_{\theta^\mu}(a\mid s_i)
       \nabla_{\theta^\mu}\log m_{\theta^\mu}(a\mid s_i).
  \end{aligned}
  \label{eq:coupling-behavior-chain-rule}
\end{equation}
Evaluation at $\theta_g^\mu$ proves Equation~\eqref{eq:coupling-gA}.
For target, $J$ is expressed in unscaled advantage units, so
\begin{equation}
  \begin{aligned}
    \nabla_{\theta^\pi}\frac{J(p_{\theta^\pi})}{\tau}
    &=\frac1{n\tau}\sum_{i,a}w_i^J\widehat{\mathbf a}_i(a)
       \nabla_{\theta^\pi}p_{\theta^\pi}(a\mid s_i)\\
    &=\frac1{n\tau}\sum_{i,a}w_i^J\widehat{\mathbf a}_i(a)
       p_{\theta^\pi}(a\mid s_i)
       \nabla_{\theta^\pi}\log p_{\theta^\pi}(a\mid s_i).
  \end{aligned}
  \label{eq:coupling-target-chain-rule}
\end{equation}
Evaluation at $\theta_g^\pi$ proves Equation~\eqref{eq:coupling-gJ}.
\end{proof}

Behavior therefore learns whether its first-action changes are supported
by target-continuation evidence; no separate $Q^{\mu_g}$ is estimated.
Target uses the same action evidence under $\nu_g^J$. For both surfaces,
the coefficient rows $m_i-\mu_i$ and $p_i-\pi_i$ sum to zero.
For any positive differentiable policy $q$, differentiating its
probability normalization gives
$\sum_a q(a\mid s)\nabla\log q(a\mid s)=0$, where the gradient is
taken with respect to that policy's parameters. Thus action-independent
baselines cancel in these two channels. The transmission coefficients
below do not generally have this cancellation property.

\subsubsection{Advantage-filtered behavior-to-target transmission}
\label{app:coupling-transmission}

Recall that $\mu_i(a),\pi_i(a)$ are the frozen parent probabilities,
$\widehat{\mathbf a}_i$ is the sampled discounted target-centered
advantage, and $\widehat\Lambda_g^P$ is their linear functional with
transmission weights $w_i^P$ in
Equation~\eqref{eq:critic-policy-shared-evidence}. Write
$\delta_i(a)=\mu_i(a)-\pi_i(a)$ for the parent probability discrepancy,
distinct from the point mass $\delta_{s_i}$ and the earlier controller
multiplier gap. Write $\delta=(\delta_i(a))_{i,a}$ for the full array;
products, absolute values, and ratios inside the functional below act
actionwise. The transmission surface is
\begin{equation}
  P(p)=\frac1\tau\widehat\Lambda_g^P
       \left[|\delta|\left(\frac p\pi-1\right)\right],
  \label{eq:coupling-P-surface}
\end{equation}
For an arbitrary candidate $p_{\theta^\pi}$,
\begin{equation}
    \nabla_{\theta^\pi}P(p_{\theta^\pi})
    =\frac1\tau\widehat\Lambda_g^P
       \left[\frac{|\delta|}{\pi}
             \nabla_{\theta^\pi}p_{\theta^\pi}\right]=\frac1\tau\widehat\Lambda_g^P
       \left[|\delta|\frac{p_{\theta^\pi}}{\pi}
             \nabla_{\theta^\pi}\log p_{\theta^\pi}\right].
  \label{eq:coupling-P-chain-rule}
\end{equation}
The parent denominator $\pi$, the discrepancy $\delta$, and the evidence
are constants here. Evaluating at $p_{\theta_g^\pi}=\pi$ cancels the
probability ratio. Define
$g_P=\left.\nabla_{\theta^\pi}P(p_{\theta^\pi})\right|_{\theta_g^\pi}$;
then
\begin{equation}
  g_P=\frac1\tau\widehat\Lambda_g^P[|\delta|\operatorname{sc}]
     =\frac1{n\tau}\sum_{i,a}w_i^P|\delta_i(a)|
       \widehat{\mathbf a}_i(a)\operatorname{sc}_i(a).
  \label{eq:coupling-gP}
\end{equation}
Disagreement supplies the magnitude and advantage supplies the sign.
Unlike $g_J$, this score coefficient is not multiplied by $\pi_i(a)$;
an action rarely selected by target can still receive substantial
transmission when behavior assigns it appreciably more mass.

Centered evidence is essential because, in general,
$\sum_a|\delta_i(a)|(p_i(a)/\pi_i(a)-1)\ne0$.
Replacing advantages by uncentered action values changes $P$.
Both the dense base and sampled correction are therefore centered before
entering this channel.

\noindent\begin{minipage}{\linewidth}
\begin{theorem}[Selective transmission in the advantage face]
\label{thm:coupling-selective-transmission}
Fix an occurrence on the finite action set $\mathcal A$. Write
$\pi=(\pi_a)_a$ for a strictly positive target distribution,
$\mu=(\mu_a)_a$ for a behavior distribution, and $e=(e_a)_a$ for
finite centered evidence with $\sum_a\pi_ae_a=0$. The evidence can be
the exact vector $\mathbf a_i$ or the realized sampled vector
$\widehat{\mathbf a}_i$. Let $\tau>0$. Define the discrepancy
$\delta_a$, its relative magnitude $h_a$, and the signed score
coefficient $b_a$ by
\begin{equation}
  \delta_a=\mu_a-\pi_a,\qquad
  h_a=\frac{|\delta_a|}{\pi_a},\qquad
  b_a=\frac{|\delta_a|e_a}{\tau}.
  \label{eq:coupling-reflection-coefficients}
\end{equation}
Write $\operatorname{sign}(x)=1$ for $x>0$ and $-1$ for $x<0$.
For $\delta_ae_a\ne0$,
\begin{equation}
  b_a=\frac{\delta_a|e_a|}{\tau}
       \operatorname{sign}(\delta_ae_a).
  \label{eq:coupling-reflection-identity}
\end{equation}
Thus the sign of the discrepancy is retained in the coefficient $b_a$
when $\delta_ae_a>0$ and reversed when $\delta_ae_a<0$.
For a nonnegative auxiliary step $\eta$, define the distribution
$q_\eta^R$ (with $R$ labeling reflection) and its normalizing constant
$Z_\eta$ by
\begin{equation}
  q_{\eta,a}^R
    =\frac{\pi_a\exp(\eta h_ae_a/\tau)}{Z_\eta},\qquad
  Z_\eta=\sum_b\pi_b\exp(\eta h_be_b/\tau).
  \label{eq:coupling-reflected-distribution}
\end{equation}
It satisfies
\begin{equation}
  \langle q_\eta^R-\pi,e\rangle
  =\frac1{Z_\eta}\sum_a\pi_ae_a
      \left[\exp(\eta h_ae_a/\tau)-1\right]\ge0.
  \label{eq:coupling-reflected-gain}
\end{equation}
For $\eta>0$, the inequality is strict if and only if
$|\delta_a||e_a|>0$ for at least one action. In addition,
\begin{equation}
  \left.\frac{d}{d\eta}\langle q_\eta^R-\pi,e\rangle
  \right|_{\eta=0}
  =\frac1\tau\sum_a|\delta_a|e_a^2.
  \label{eq:coupling-reflected-initial-slope}
\end{equation}
\end{theorem}
\end{minipage}

\begin{proof}
For one occurrence, the transmission contribution of action $a$ to a
candidate $p$ is
\begin{equation}
  T_a(p)=\frac{|\delta_a|}{\tau\pi_a}(p_a-\pi_a)e_a.
  \label{eq:coupling-coordinate-contribution}
\end{equation}
The omitted occurrence weight is positive and does not change signs.
Whenever $\delta_a\ne0$, $T_a(p)$ is positive precisely when the
candidate's coordinate change is supported by the evidence. We examine
the four nonzero sign combinations separately.
\begin{enumerate}[label=(\roman*),leftmargin=*]
  \item \emph{Positive advantage and increased behavioral mass:}
  $e_a>0$, $\delta_a>0$. Behavior assigns more probability to a favorable
  action. Here $b_a=\delta_ae_a/\tau>0$, so increasing $p_a$ contributes
  positively to $P$. The behavioral deviation is retained.

  \item \emph{Negative advantage and decreased behavioral mass:}
  $e_a<0$, $\delta_a<0$. Behavior assigns less probability to an
  unfavorable action. Here $b_a=(-\delta_a)e_a/\tau<0$, and decreasing
  $p_a$ makes both $(p_a-\pi_a)$ and $e_a$ negative, so $T_a(p)>0$.
  This beneficial reduction is also retained.

  \item \emph{Positive advantage and decreased behavioral mass:}
  $e_a>0$, $\delta_a<0$. Behavior underweights a favorable action.
  Following that decrease would make $T_a(p)<0$. Instead,
  $b_a=(-\delta_a)e_a/\tau>0$ rewards an increase, reflecting the
  behavioral deviation.

  \item \emph{Negative advantage and increased behavioral mass:}
  $e_a<0$, $\delta_a>0$. Behavior overweights an unfavorable action.
  Following that increase would again make $T_a(p)<0$. Here
  $b_a=\delta_ae_a/\tau<0$ rewards a decrease, reflecting this deviation.
\end{enumerate}
These cases establish Equation~\eqref{eq:coupling-reflection-identity}.
If $e_a=0$ or $\delta_a=0$, both $b_a$ and the coordinate contribution
are zero. Coordinate statements describe contributions to the surface;
feasible distributions must also obey the shared unit-mass constraint.

We next show that the auxiliary distribution respects this constraint
while preserving a nonnegative total advantage. Positivity of $\pi$ and
finiteness of the evidence give $0<Z_\eta<\infty$. Consequently
$q_\eta^R$ is a positive probability distribution. Centering gives
\[
  \langle q_\eta^R-\pi,e\rangle
  =\frac1{Z_\eta}\sum_a\pi_ae_a\exp(\eta h_ae_a/\tau)
  =\frac1{Z_\eta}\sum_a\pi_ae_a
       [\exp(\eta h_ae_a/\tau)-1].
\]
If $e_a>0$, then $h_a\ge0$ makes the bracket nonnegative.
If $e_a<0$, the bracket is nonpositive, so its product with $e_a$ is
again nonnegative. If $e_a=0$ or $h_a=0$, the product is zero.
For $\eta>0$, it is strictly positive exactly when $h_a|e_a|>0$.
Since all $\pi_a>0$, summation proves the inequality and its strictness
condition.

Finally, let $N_\eta$ be the numerator on the right-hand side of
Equation~\eqref{eq:coupling-reflected-gain}. At $\eta=0$,
$N_0=0$ and $Z_0=1$. The quotient rule therefore gives
\[
  \left.\frac{d}{d\eta}\frac{N_\eta}{Z_\eta}\right|_{0}
  =N'_0
  =\frac1\tau\sum_a\pi_ah_ae_a^2
  =\frac1\tau\sum_a|\delta_a|e_a^2,
\]
which proves Equation~\eqref{eq:coupling-reflected-initial-slope}.
\end{proof}

Here favorable and unfavorable mean positive and negative centered
evidence relative to $\pi$. The theorem does not require behavior to be
better than target in aggregate: beneficial deviations in the first two
cases are retained even when behavior has harmful deviations elsewhere.
The latter two cases use the disagreement magnitude while reversing its
unsupported direction.

The auxiliary distribution also explains the surface itself. For any
candidate action distribution $p$ at the same occurrence, setting
$\eta=1$ and using probability normalization gives
\begin{equation}
  \begin{aligned}
    \sum_a(p_a-\pi_a)\log\frac{q_{1,a}^R}{\pi_a}
    &=\sum_a(p_a-\pi_a)
       \left(\frac{h_ae_a}{\tau}-\log Z_1\right)\\
    &=\frac1\tau\sum_a|\mu_a-\pi_a|
       \left(\frac{p_a}{\pi_a}-1\right)e_a.
  \end{aligned}
  \label{eq:coupling-log-teacher-equivalence}
\end{equation}
The normalizer cancels because $\sum_a(p_a-\pi_a)=0$. The implementation
evaluates the right-hand side directly; $q_\eta^R$ is a proof device, not
an additional actor or a policy-mixing step.

With exact evidence $e=\mathbf a_i$, the theorem concerns the frozen
model's first-action evaluation. With $e=\widehat{\mathbf a}_i$, it is a
pathwise statement about the realized occurrence-level evidence; repeated
states need not define a single common auxiliary policy. The specific
auxiliary path has nonnegative advantage gain, but an arbitrary increase
of $P$ need not increase the unweighted advantage face. Nor does the
theorem assign signs to every coordinate of a parameterized actor update.
For independent categorical logits at this occurrence, the actual
parent logit gradient for this occurrence is the positive occurrence
weight times the following vector, where $b=(b_a)_a$ recalls the
coefficients in Equation~\eqref{eq:coupling-reflection-coefficients}:
\begin{equation}
  b-\pi\sum_a b_a,
  \label{eq:coupling-logit-centering}
\end{equation}
because differentiating the softmax couples all action probabilities.
Shared network parameters further couple states. The channel can increase
behavior--target disagreement. It vanishes identically when $\mu=\pi$;
a zero parameter gradient alone does not imply a constant finite surface.

\subsubsection{Why the advantage channels use the scale $1/\tau$}
\label{app:coupling-tau-scale}

Recall that $g_H=g_{H,g}$ is the parent gradient of the heuristic
objective $H_g$ in Equation~\eqref{eq:h-explicit-gradient}.
The heuristic objective is a dimensionless squared statistical distance;
its utility dependence enters through the regularized controller.
The unscaled advantage surfaces instead carry reward units. Since
$\tau=\kappa S_{\mathrm{cal}}$ in Equation~\eqref{eq:h-tau-calibration}
has the same units as the heuristic utility, division by $\tau$ expresses
the three advantage channels in compatible units. The next result also
quantifies the utility-induced scale of guidance while accounting for
any behavior--target mismatch already present at the parent.

\begin{theorem}[Scale of heuristic guidance]
\label{thm:coupling-heuristic-scale}
Recall the frozen centered heuristic utility $u_g$, the
Hellinger-regularized controller $q_{H,g}$ from
Equation~\eqref{eq:h-controller-solution}, and its finite empirical
fitting measure $\xi_g$ from Equation~\eqref{eq:h-fitting-measure}.
The utility is the undiscounted finite-horizon quantity, distinct from
the discounted advantages $\widehat{\mathbf a}_i$.
Recall also $d_{\mathrm{FR}}(p,q)=2\arccos\sum_a\sqrt{p_aq_a}$
for action distributions $p,q$. Assume finite utilities and actor scores
on this panel, positive parent probabilities, and $\tau>0$.
Define the utility range $R_g(s)$ and the root-mean-square parent
distance $D_{\mu,\pi}$ by
\begin{equation}
  R_g(s)=\max_a u_g(s,a)-\min_a u_g(s,a),\qquad
  D_{\mu,\pi}
  =\left(\mathbb E_{\xi_g}d_{\mathrm{FR}}^2(\mu_g,\pi_g)\right)^{1/2}.
  \label{eq:coupling-utility-range}
\end{equation}
For symmetric matrices, write $X\succeq Y$ when $X-Y$ is positive
semidefinite and $X\succ0$ when $X$ is positive definite; a superscript
$\top$ denotes transpose. Let the behavior-actor matrix $F_\mu\succ0$
satisfy
\begin{equation}
  F_\mu\succeq\frac12\mathbb E_{s\sim\xi_g}
     \sum_a\mu_g(a\mid s)\operatorname{sc}_g^\mu(s,a)
                   \operatorname{sc}_g^\mu(s,a)^\top.
  \label{eq:coupling-fisher-domination}
\end{equation}
The actor matrix defined below has this property. Use
$\|x\|_G=(x^\top Gx)^{1/2}$ for a positive-definite $G$.
The numerical factor $\pi$ in the following bounds is the circle constant.
Then
\begin{equation}
  d_{\mathrm{FR}}(\pi_g(\cdot\mid s),q_{H,g}(\cdot\mid s))
  \le\frac{\pi R_g(s)}{4\tau},
  \label{eq:coupling-controller-scale}
\end{equation}
and
\begin{equation}
  \|g_H\|_{F_\mu^{-1}}
  \le\sqrt2\left[
       D_{\mu,\pi}+\frac{\pi}{4\tau}
       \left(\mathbb E_{\xi_g}R_g(S)^2\right)^{1/2}
     \right].
  \label{eq:coupling-guidance-scale}
\end{equation}
In particular, at matched parents $\mu_g=\pi_g$, the natural norm is
$O(\tau^{-1})$ as $\tau\to\infty$ for fixed fitting locations,
utility evidence, parent policies, and score vectors.
\end{theorem}

\begin{proof}
At one fitting location, abbreviate $p=\pi_g(\cdot\mid s)$,
$q=q_{H,g}(\cdot\mid s)$, $u_a=u_g(s,a)$, and
$\lambda=\lambda_g(s)$, the controller's unit-mass multiplier.
Set $B_{\pi,H}=\sum_a\sqrt{p_aq_a}$. Here $\mathbb E_q$ and
$\operatorname{Var}_q$ denote expectation and variance over actions
drawn from $q$.
This affinity is between target and controller, distinct from the
behavior--controller affinity in Equation~\eqref{eq:h-affinity-and-score}.
The stationarity equation~\eqref{eq:h-controller-stationarity} gives
$\lambda-u_a=2\tau\sqrt{p_a/q_a}$. Its first two moments under $q$ are
\[
  \mathbb E_q(\lambda-u)=2\tau B_{\pi,H},\qquad
  \mathbb E_q[(\lambda-u)^2]=4\tau^2.
\]
Thus $\operatorname{Var}_q(u)=4\tau^2(1-B_{\pi,H}^2)$.
Set $\psi=\arccos B_{\pi,H}\in[0,\pi/2]$. Since
$2\psi\le\pi\sin\psi$ and a variable of range $R_g(s)$ has variance
at most $R_g(s)^2/4$,
\[
  d_{\mathrm{FR}}(p,q)=2\psi
  \le\frac{\pi}{2\tau}\sqrt{\operatorname{Var}_q(u)}
  \le\frac{\pi R_g(s)}{4\tau}.
\]

For the guidance gradient, apply Cauchy--Schwarz to the centered formula
in Equation~\eqref{eq:h-explicit-gradient}. Recall the
behavior--controller affinity
$\mathcal B_g(s)=\sum_a\sqrt{\mu_g(a\mid s)q_{H,g}(a\mid s)}$
and angle $\varphi_g(s)=\arccos\mathcal B_g(s)$.
At each state, the centered
factor $\sqrt{q_{H,g}/\mu_g}-\mathcal B_g$ has second moment
$1-\mathcal B_g^2=\sin^2\varphi_g$ under $\mu_g$; this cancels the
denominator $\sin\varphi_g$ in that formula. The coincident case follows
by continuity. For any parameter direction $v$, a second
Cauchy--Schwarz inequality over the fitting locations gives
\begin{equation}
    |g_H^\top v|
    \le
      \left(\mathbb E_{\xi_g}
          d_{\mathrm{FR}}^2(\mu_g,q_{H,g})\right)^{1/2}
          \times
      \left(\mathbb E_{\xi_g}\sum_a\mu_g(a\mid S)
          [\operatorname{sc}_g^\mu(S,a)^\top v]^2\right)^{1/2}.
  \label{eq:coupling-guidance-dual-bound}
\end{equation}
Equation~\eqref{eq:coupling-fisher-domination} bounds the last factor by
$\sqrt2\|v\|_{F_\mu}$. Taking the dual norm, then using the metric
triangle inequality and Minkowski's inequality, proves
Equation~\eqref{eq:coupling-guidance-scale}.
\end{proof}

For a more direct interpretation, hold the finite fitting locations,
utility values, parent policies, and score vectors fixed and let
$\tau\to\infty$. The remainders below are uniform over this finite
panel and action set; the vector remainder is measured in the Euclidean
parameter norm. At each state, put $\epsilon_\tau=1/(2\tau)$ and
$y_\tau=\lambda_g(s)/(2\tau)$. The controller normalization is
$\sum_a\pi_g(a\mid s)(y_\tau-\epsilon_\tau u_g(s,a))^{-2}=1$.
At $(y_\tau,\epsilon_\tau)=(1,0)$ its derivative with respect to
$y_\tau$ is $-2$, so the implicit function theorem and target centering
give $y_\tau=1+O(\epsilon_\tau^2)$. Expanding the inverse square
therefore yields
\begin{equation}
  q_{H,g}(a\mid s)=\pi_g(a\mid s)
     \left[1+\frac{u_g(s,a)}{\tau}+O(\tau^{-2})\right].
  \label{eq:coupling-controller-expansion}
\end{equation}
At $\mu_g=\pi_g$, this implies, pointwise on the fixed panel,
$\sqrt{\pi_gq_{H,g}}=\pi_g[1+u_g/(2\tau)+O(\tau^{-2})]$ and
$\mathcal B_g=1+O(\tau^{-2})$. Also
$2\varphi_g/\sin\varphi_g=2+O(\tau^{-2})$, with the continuous
value $2$ at $\varphi_g=0$. Substitution in the centered guidance
gradient of Equation~\eqref{eq:h-explicit-gradient} yields
\begin{equation}
  g_H=\frac1\tau\mathbb E_{\xi_g}\sum_a\pi_g(a\mid S)
        u_g(S,a)\operatorname{sc}_g^\mu(S,a)+O(\tau^{-2}).
  \label{eq:coupling-guidance-expansion}
\end{equation}
This fixed-panel expansion does not differentiate through the controller
rollouts that originally selected $\xi_g$. In general, the whole
gradient need not decay as $1/\tau$: if $u_g\equiv0$, then
$q_{H,g}=\pi_g$ for every $\tau$, while guidance can remain nonzero
when $\mu_g\ne\pi_g$. The mismatch term in
Theorem~\ref{thm:coupling-heuristic-scale} accounts for this component.

There is also an exact unit-conversion interpretation. Multiplying all
frozen return evidence and calibration utilities by $c>0$ multiplies
$u_g,\widehat{\mathbf a}_i$, and $\tau$ by $c$. The controller, $H_g$,
and the scaled surfaces $A,J/\tau,P$ are unchanged. This is an invariance
of the frozen update interface, not a claim that arbitrary critic
optimization is invariant to a change of reward units. It does not make
the four gradient norms equal; their relative contributions are handled
by the fusion below.

\subsubsection{Fisher geometry and the Fuse construction}
\label{app:coupling-fusion}

The score-based Fisher metric and inverse-Fisher preconditioning follow
the natural-policy-gradient construction of \citet{kakade2001natural}.
Here we specify the frozen consumer measures and the fusion rule.
Recall the four empirical state measures: $D_g$ is the unweighted
generation batch, $\nu_g^J$ and $\nu_g^P$ use the positive mean-one
weights $w_i^J$ and $w_i^P$, and $\xi_g$ is the heuristic fitting
measure from Equation~\eqref{eq:h-fitting-measure}.
For a parent actor $q$ and a frozen probability measure $\nu$ on
observations, let $\operatorname{sc}_q(s,a)$ be its log-probability
gradient at its own parent parameters. Define its empirical Fisher
matrix $\mathsf F(q,\nu)$ by
\begin{equation}
  \mathsf F(q,\nu)=\mathbb E_{s\sim\nu}\sum_aq(a\mid s)
       \operatorname{sc}_q(s,a)\operatorname{sc}_q(s,a)^\top,
  \label{eq:coupling-channel-fisher}
\end{equation}
Let $I$ denote the identity matrix on the relevant actor parameter
space and $\lambda_F>0$ the damping coefficient. The common matrix
for each actor's two channels is
\begin{equation}
  \begin{aligned}
    F_\mu&=\tfrac12\left[
       \mathsf F(\mu_g,D_g)+\mathsf F(\mu_g,\xi_g)\right]
       +\lambda_F I,\\
    F_\pi&=\tfrac12\left[
       \mathsf F(\pi_g,\nu_g^J)+\mathsf F(\pi_g,\nu_g^P)\right]
       +\lambda_F I.
  \end{aligned}
  \label{eq:coupling-actor-fishers}
\end{equation}
Thus the common matrix contains a single ridge term $\lambda_F I$.
The matrices are positive definite and actor-specific; the two policies
do not share a parameter-space update.

For either $(g_1,g_2)=(g_A,g_H)$ or $(g_1,g_2)=(g_J,g_P)$, use the
corresponding fixed matrix $F=F_\mu$ or $F=F_\pi$. In this subsection,
$i\in\{1,2\}$ indexes channels, rather than data occurrences.
Define their natural norms $n_i$, their maximum $n_{\max}$, and the
scalar denominator $\Delta_F$ by
\begin{equation}
  n_i=\sqrt{g_i^\top F^{-1}g_i},\quad
  n_{\max}=\max(n_1,n_2),\quad
  \Delta_F=n_1^2+6n_1n_2+n_2^2.
  \label{eq:coupling-natural-norms}
\end{equation}
For $n_{\max}>0$, define the fusion coefficients $\alpha_1,\alpha_2$
and the ideal parameter direction $d$ by
\begin{equation}
  \begin{gathered}
    \alpha_1=\frac{n_{\max}(n_1+3n_2)}{\Delta_F},\qquad
    \alpha_2=\frac{n_{\max}(n_2+3n_1)}{\Delta_F},\\
    d=\operatorname{Fuse}_F(g_1,g_2)
      =F^{-1}(\alpha_1g_1+\alpha_2g_2).
  \end{gathered}
  \label{eq:coupling-fusion-definition}
\end{equation}
If both norms are zero, set both coefficients and $d$ to zero. For
$n_{\max}>0$, the normalized contributions
\begin{equation}
  \omega_1=\frac{n_1(n_1+3n_2)}{\Delta_F},\qquad
  \omega_2=\frac{n_2(n_2+3n_1)}{\Delta_F}
  \label{eq:coupling-fusion-votes}
\end{equation}
satisfy $\omega_1+\omega_2=1$ and
$\alpha_i n_i=n_{\max}\omega_i$. The $\alpha_i$ themselves need not
sum to one.

\begin{theorem}[Variational characterization of the fusion weights]
\label{thm:coupling-fusion-weights}
Let $n_1,n_2>0$, and define the two-coordinate probability simplex
$\Delta_2=\{\omega\in\mathbb R^2:\omega_i\ge0,\ \omega_1+\omega_2=1\}$.
For probability vectors $x,y$ with $y_i>0$, define the Pearson
divergence $\chi^2(x\|y)=\sum_i(x_i-y_i)^2/y_i$.
The weights in Equation~\eqref{eq:coupling-fusion-votes} are the unique
solution of
\begin{equation}
  \min_{\omega\in\Delta_2}
  \left\{
    \chi^2\!\left(\omega\,\middle\|
               \frac{(n_1,n_2)}{n_1+n_2}\right)
    +\chi^2\!\left(\omega\,\middle\|(1/2,1/2)\right)
  \right\}.
  \label{eq:coupling-vote-objective}
\end{equation}
\end{theorem}
\begin{proof}
Write $r_i=n_i/(n_1+n_2)$ for the normalized natural strengths.
The objective equals
$\sum_i\omega_i^2(1/r_i+2)-2$. Its constrained minimizer obeys
$\omega_i\propto(1/r_i+2)^{-1}$, yielding
Equation~\eqref{eq:coupling-fusion-votes}. The positive diagonal
quadratic coefficients make the objective strictly convex, proving
uniqueness.
\end{proof}
For one zero norm, use the extended convention: a summand with $y_i=0$
is zero if $x_i=0$ and $+\infty$ if $x_i>0$. The only finite feasible
weight then assigns all mass to the nonzero channel, agreeing with
Equation~\eqref{eq:coupling-fusion-votes}. The double-zero case uses
the separately defined zero direction and requires no normalized votes.
The two terms balance relative natural-gradient strength and equal
channel contributions.

Equal positive norms give $\alpha_1=\alpha_2=1/2$. If $n_1>0,n_2=0$,
the coefficients are $(1,3)$ and $d=F^{-1}g_1$. The second gradient
contributes zero, but its frozen finite objective need not be constant
away from the parent. A vanishing channel has a continuously vanishing
contribution, and the fused direction extends continuously to the
double-zero case without a threshold or denominator floor.

\begin{theorem}[Bounded fusion and channel compatibility]
\label{thm:coupling-fusion}
Let $\theta_\star$ denote the parent parameter vector of the actor
under consideration, and let $U_1,U_2$ be two differentiable channel
objectives on a neighborhood of $\theta_\star$. Let $F\succ0$ be
fixed and $g_i=\nabla U_i(\theta_\star)$ for $i\in\{1,2\}$.
Construct $n_i,\alpha_i,d$ as in
Equations~\eqref{eq:coupling-natural-norms}--\eqref{eq:coupling-fusion-definition};
the votes $\omega_i$ in Equation~\eqref{eq:coupling-fusion-votes} are
used only when $n_{\max}>0$. Recall
$\|x\|_F=(x^\top Fx)^{1/2}$.
When $n_1n_2>0$, define the unit natural directions $v_i$ and their
Fisher cosine $c_F$ by
\begin{equation}
  v_i=\frac{F^{-1}g_i}{n_i},\qquad
  c_F=\frac{g_1^\top F^{-1}g_2}{n_1n_2}\in[-1,1].
  \label{eq:coupling-fisher-cosine}
\end{equation}
For each $i\in\{1,2\}$, let $j=3-i$ denote the other channel. Then
\begin{equation}
  \begin{aligned}
    d&=n_{\max}(\omega_1v_1+\omega_2v_2),\\
    n_{\max}|\omega_1-\omega_2|
       &\le\|d\|_F\le n_{\max},\\
    g_i^\top d&=n_{\max}n_i(\omega_i+c_F\omega_j),\qquad j\ne i.
  \end{aligned}
  \label{eq:coupling-fusion-bounds}
\end{equation}
Both directional derivatives are nonnegative if and only if
\begin{equation}
  c_F\ge-\min\left\{\frac{\omega_1}{\omega_2},
                           \frac{\omega_2}{\omega_1}\right\}.
  \label{eq:coupling-common-ascent-condition}
\end{equation}
The channel with the larger natural norm always has a nonnegative
directional derivative. If $n_1\ge n_2$, the weaker channel satisfies
\begin{equation}
  g_2^\top d\ge
  -n_{\max}n_2\frac{n_1^2-n_2^2}{\Delta_F}.
  \label{eq:coupling-conflict-bound}
\end{equation}

For any norms, hold $\alpha_1,\alpha_2$ fixed and set
$U=\alpha_1U_1+\alpha_2U_2$. Then
$\nabla U(\theta_\star)^\top d=\|d\|_F^2$.
If $U$ is twice continuously differentiable along the update segment and
$\nabla^2U\succeq-LF$ there for a nonnegative curvature constant $L$, then for a parameter-step
length $t\ge0$,
\begin{equation}
  U(\theta_\star+td)-U(\theta_\star)
  \ge t\left(1-\frac{Lt}{2}\right)\|d\|_F^2,
  \qquad t\ge0.
  \label{eq:coupling-finite-fused-bound}
\end{equation}
The bound applies whenever the entire segment is in the stated domain.
\end{theorem}

\begin{proof}
The identity $\alpha_i n_i=n_{\max}\omega_i$ gives the representation
of $d$. Each $v_i$ has unit $F$-norm, so the triangle and reverse-triangle
inequalities give the two norm bounds. Since
$g_i^\top v_i=n_i$ and $g_i^\top v_j=n_ic_F$, taking the inner product
with $g_i$ gives its slope. Both slopes are nonnegative exactly when
$c_F\ge-\omega_1/\omega_2$ and
$c_F\ge-\omega_2/\omega_1$, which is
Equation~\eqref{eq:coupling-common-ascent-condition}.

Furthermore,
$\omega_1-\omega_2=(n_1^2-n_2^2)/\Delta_F$.
If $n_1\ge n_2$, then $\omega_1\ge\omega_2$ and $c_F\ge-1$ gives
$g_1^\top d\ge n_{\max}n_1(\omega_1-\omega_2)\ge0$ and the stated
lower bound for $g_2^\top d$. The zero-norm cases follow directly from
positive definiteness of $F$ and the definition of $d$.

Finally, with the coefficients frozen,
$\nabla U(\theta_\star)=\alpha_1g_1+\alpha_2g_2=Fd$.
Taylor's formula with integral remainder yields
\[
  U(\theta_\star+td)-U(\theta_\star)
  =t\|d\|_F^2+
    \int_0^t(t-s)d^\top\nabla^2U(\theta_\star+sd)d\,ds.
\]
The Hessian lower bound makes the integral at least
$-Lt^2\|d\|_F^2/2$, proving the result.
\end{proof}

Likewise, suppose each $U_i$ is twice continuously differentiable on
the segment and $\nabla^2U_i\succeq-L_iF$ there for constants $L_i\ge0$.
For the same nonnegative step length $t$,
\begin{equation}
  U_i(\theta_\star+td)-U_i(\theta_\star)
  \ge t g_i^\top d-\frac{L_it^2}{2}\|d\|_F^2.
  \label{eq:coupling-finite-channel-bound}
\end{equation}
When both norms are positive, the slope is
$g_i^\top d=n_{\max}n_i(\omega_i+c_F\omega_j)$; the displayed form also
covers zero-norm channels. Strictly positive initial slopes therefore
give simultaneous improvement for sufficiently small positive steps. Behavior uses
$(U_1,U_2)=(A,H_g)$ and target uses $(U_1,U_2)=(J/\tau,P)$.
No derivative passes through the parent Fisher matrices, norms, or
fusion coefficients.

These are conditional compatibility guarantees, not unconditional
common ascent. For example, $F=1$, $g_1=2$, and $g_2=-1$ give
$(\alpha_1,\alpha_2)=(10/17,14/17)$ and $d=6/17$, so the second
channel decreases to first order. A larger natural norm describes
gradient strength in this geometry, not statistical confidence.

\subsubsection{Numerical updates and endpoint feasibility control}
\label{app:coupling-numerics}

\paragraph{Fisher damping and its selection.}
The ridge $\lambda_F$ is a fixed positive hyperparameter that stabilizes
the linear systems and makes the common Fisher matrices invertible.
We use
\begin{equation}
  \lambda_F=0.0067202476327024195
  \label{eq:coupling-ridge}
\end{equation}
for both actors, all initialization sources, and all generations.
This value was inherited from an earlier cold-start study of the
precursor implementation that jointly varied the critic learning rate,
emphatic learning rate, and Fisher damping. The study evaluated $12$ fixed Sobol configurations,
mapping damping logarithmically over $[10^{-4},10^{-2}]$, with three
training seeds per configuration. Each run used $60$ generations of
$25{,}000$ transitions. Configurations were ranked by the mean stochastic
target-policy cost across the three seeds on common tuning tapes, with
ties resolved by configuration index. The selected damping was then
fixed and inherited by the present implementation. The current seven
hyperparameter configurations vary the critic learning rate, emphatic
learning rate, and heuristic multiplier $\kappa$; they do not retune
$\lambda_F$. It is not adapted from within-run returns or gradient norms.

\paragraph{Numerical natural-gradient directions.}
Two conjugate-gradient (CG) solves estimate the channel natural norms.
Let $\widetilde\alpha_1,\widetilde\alpha_2$ be the coefficients obtained
by applying Equation~\eqref{eq:coupling-fusion-definition} to those
estimated norms. After freezing these coefficients, a separate CG solve
uses the right-hand side $\widetilde\alpha_1g_1+\widetilde\alpha_2g_2$. The final direction
is not formed by adding the two approximate channel solutions.
Fisher-vector products avoid constructing a dense parameter-space matrix.
The J14 r3 implementation uses a base limit of $256$ iterations and
Euclidean relative residual tolerance $10^{-4}$. The retained J12
implementation uses the same base limit and tolerance $5\times10^{-5}$.
Numerical recovery retains the same frozen equations and data rather
than resampling evidence.

\begin{theorem}[Residual bound for a numerical fused solve]
\label{thm:coupling-cg-error}
Let $F\succ0$ and let $\widetilde\alpha_1,\widetilde\alpha_2$ be the
fixed coefficients actually passed to the fused CG solve, possibly
computed from approximate channel norms. Define its exact solution
$d_{\widetilde\alpha}=F^{-1}(\widetilde\alpha_1g_1+
\widetilde\alpha_2g_2)$. For a computed direction $\widetilde d$, define
the linear-system residual $r_{\mathrm{CG}}$ and its dual norm
$\varepsilon_{\mathrm{CG}}$ by
\begin{equation}
  r_{\mathrm{CG}}=(\widetilde\alpha_1g_1+\widetilde\alpha_2g_2)
                   -F\widetilde d,\qquad
  \varepsilon_{\mathrm{CG}}=\|r_{\mathrm{CG}}\|_{F^{-1}}.
  \label{eq:coupling-cg-residual}
\end{equation}
With the exact channel norms $n_i=\|g_i\|_{F^{-1}}$,
\begin{equation}
  \|\widetilde d-d_{\widetilde\alpha}\|_F=\varepsilon_{\mathrm{CG}},\qquad
  |g_i^\top(\widetilde d-d_{\widetilde\alpha})|
       \le n_i\varepsilon_{\mathrm{CG}}.
  \label{eq:coupling-cg-error-bound}
\end{equation}
If $d$ and $\alpha_i$ denote the ideal exact-norm fusion from
Equation~\eqref{eq:coupling-fusion-definition}, then also
\begin{equation}
  \|\widetilde d-d\|_F
  \le\varepsilon_{\mathrm{CG}}
       +\sum_{i=1}^2|\widetilde\alpha_i-\alpha_i|n_i.
  \label{eq:coupling-total-direction-error}
\end{equation}
\end{theorem}
\begin{proof}
The residual identity gives
$d_{\widetilde\alpha}-\widetilde d=F^{-1}r_{\mathrm{CG}}$.
Taking the $F$-norm proves the equality, and dual Cauchy--Schwarz
proves the slope bound. Finally,
$d_{\widetilde\alpha}-d=\sum_i(\widetilde\alpha_i-\alpha_i)F^{-1}g_i$;
the triangle inequality and $\|F^{-1}g_i\|_F=n_i$ prove the last bound.
\end{proof}
Here $\varepsilon_{\mathrm{CG}}$ is an analytical residual norm; the
runtime stopping tolerance uses a Euclidean relative residual and does
not directly equal this quantity. The coefficient-error term must also
be controlled to transfer the exact-norm guarantees of
Theorem~\ref{thm:coupling-fusion} to a numerical direction. Endpoint
control below evaluates the numerical candidate directly.

\paragraph{Finite updates and the two channel budgets.}
Recall that the Fuse directions in
Equation~\eqref{eq:coupling-fusion-definition} use exact solves at the
frozen parents. Let $t_\mu,t_\pi\in[0,1]$ denote behavior and target
step fractions. Because all channels are maximization objectives,
the ideal updates use the ascent sign:
\begin{equation}
  \begin{aligned}
    \theta_{g+1}^\mu
      &=\theta_g^\mu+t_\mu\operatorname{Fuse}_{F_\mu}(g_A,g_H),\\
    \theta_{g+1}^\pi
      &=\theta_g^\pi+t_\pi\operatorname{Fuse}_{F_\pi}(g_J,g_P).
  \end{aligned}
  \label{eq:coupling-actor-updates}
\end{equation}
For either actor, let $\theta_\star$ be its parent parameters and
$\widetilde d$ its computed direction. Endpoint feasibility control
(EFC) chooses a step fraction $t\in[0,1]$. Let $\operatorname{FP32}$
denote rounding to the stored single-precision floating-point parameter
representation. EFC evaluates the actual candidate
$\theta(t)=\operatorname{FP32}(\theta_\star+t\widetilde d)$.
Let $q$ be the parent policy, $q_t$ the policy represented by
$\theta(t)$, and $\nu$ one of the actor's channel probability measures.
Define the averaged Pearson discrepancy $\mathcal D_{2,\nu}$ by
\begin{equation}
  \begin{aligned}
    \mathcal D_{2,\nu}(q_t\|q)
    &=\mathbb E_{s\sim\nu}\sum_a
       \frac{(q_t(a\mid s)-q(a\mid s))^2}{q(a\mid s)}\\
    &=\mathbb E_{s\sim\nu}\sum_a
       \frac{q_t(a\mid s)^2}{q(a\mid s)}-1.
  \end{aligned}
  \label{eq:coupling-endpoint-discrepancy}
\end{equation}
Let $\varepsilon_{\mathrm{EFC}}=0.01$ be the per-channel endpoint
budget. The numerical constraint
$\mathcal D_{2,\nu}(q_t\|q)\le\varepsilon_{\mathrm{EFC}}$
is checked separately for
$\nu=D_g,\xi_g$ for behavior and $\nu=\nu_g^J,\nu_g^P$ for target.
Checking both measures separately prevents one channel's small movement
from concealing a large movement under the other.

EFC first tests $t=1$. If infeasible, it initializes the feasible lower
endpoint at the parent $t=0$, the upper endpoint at $1$, and performs
$24$ midpoint probes. A feasible probe replaces the lower endpoint and
its cached parameter vector; an infeasible probe replaces the upper
endpoint. The algorithm commits that exact cached feasible vector,
without reconstructing it from the returned scalar. Thus the committed
parameters are exactly those that passed the implemented numerical
checks. This search does not require global monotonicity of feasibility
and need not return the largest feasible step. The mathematical bounds
below assume the stated discrepancy inequality for that endpoint;
finite-precision checks alone are not interval-arithmetic certificates
of the exact inequality.

\begin{corollary}[Finite policy-change bounds under each channel measure]
\label{cor:coupling-endpoint-bounds}
Let $\nu$ be a probability measure on observations, let $q$ be positive
on the finite action set, and let $q_t$ be a candidate policy. Recall
that $t$ labels the endpoint step fraction. Define, at each observation,
the total variation and forward Kullback--Leibler divergence by
$\operatorname{TV}(q_t,q)=\tfrac12\sum_a|q_t(a)-q(a)|$ and
$\operatorname{KL}(q_t\|q)=\sum_aq_t(a)\log(q_t(a)/q(a))$,
using natural logarithms and the convention $0\log0=0$.
If $\mathcal D_{2,\nu}(q_t\|q)\le\varepsilon_{\mathrm{EFC}}$, then
\begin{equation}
  \mathbb E_\nu\operatorname{TV}(q_t,q)
     \le\frac{\sqrt{\varepsilon_{\mathrm{EFC}}}}2,
  \qquad
  \mathbb E_\nu\operatorname{KL}(q_t\|q)
     \le\log(1+\varepsilon_{\mathrm{EFC}}).
  \label{eq:coupling-endpoint-tv-kl}
\end{equation}
For any frozen real action payoff $f(s,a)$ with finite integrated
conditional variance $\mathbb E_{S\sim\nu}
\operatorname{Var}_{a\sim q(\cdot\mid S)}f(S,a)<\infty$,
\begin{equation}
  \begin{aligned}
    &\left|\mathbb E_\nu\sum_a
       (q_t(a\mid S)-q(a\mid S))f(S,a)\right|\\
    &\qquad\le\sqrt{\varepsilon_{\mathrm{EFC}}}
       \left[\mathbb E_\nu
          \operatorname{Var}_{a\sim q(\cdot\mid S)}f(S,a)
       \right]^{1/2}.
  \end{aligned}
  \label{eq:coupling-endpoint-payoff-bound}
\end{equation}
\end{corollary}

\begin{proof}
Set $\bar f(s)=\sum_aq(a\mid s)f(s,a)$. Probability normalization
allows replacing $f$ by $f-\bar f$ in the policy difference. Applying
Cauchy--Schwarz under the joint measure $\nu(ds)q(a\mid s)$ gives
the following bound, where $\mathbb E_{\nu,q}$ denotes expectation
under that measure:
\[
  \left|\mathbb E_{\nu,q}
     \left[\left(\frac{q_t}{q}-1\right)(f-\bar f)\right]\right|
  \le\sqrt{\mathcal D_{2,\nu}(q_t\|q)}
      \sqrt{\mathbb E_\nu\operatorname{Var}_{q}(f)}.
\]
Similarly, $\operatorname{TV}(q_t,q)=
\tfrac12\mathbb E_q|q_t/q-1|$ gives the total-variation bound.
Recall the pointwise Pearson divergence
$\chi^2(q_t\|q)=\sum_a(q_t(a)-q(a))^2/q(a)$, whose $\nu$-average
is $\mathcal D_{2,\nu}(q_t\|q)$. At each state, Jensen's inequality
under $q_t$ yields
\[
  \operatorname{KL}(q_t\|q)
  =\mathbb E_{q_t}\log(q_t/q)
  \le\log\mathbb E_{q_t}(q_t/q)
  =\log\left(1+\chi^2(q_t\|q)\right).
\]
Averaging the displayed inequality over $\nu$ and applying Jensen
again proves the KL bound.
\end{proof}

Under the corollary's discrepancy premise with the implemented budget,
the average total-variation change is at most $0.05$ under each channel
measure. These constraints are empirical
averages, not uniform bounds at every state. EFC controls distribution
movement and does not check objective improvement or the
curvature-dependent step conditions in Theorem~\ref{thm:coupling-fusion}.
The fusion result characterizes the ideal proposed direction; the
residual theorem describes numerical direction error, and the endpoint
corollary bounds finite policy change whenever its discrepancy premise
holds.

\section{Experiments Protocol}
\label{app:protocol}

We compare HBPO with nine native methods:
PPO \citep{schulman2017ppo},
A2C \citep{mnih2016asynchronous},
Discrete SAC \citep{haarnoja2018sac,christodoulou2019discrete},
ACE \citep{imani2018emphatic},
PPO-81 \citep{schulman2017ppo},
TRPO \citep{schulman2015trpo},
GRPO \citep{shao2024deepseekmath},
TPO \citep{kaddour2026tpo}, and
Guided Policy Search (GPS) \citep{levine2013gps,levine2016visuomotor}. Each method
provides a warm initialization; Cold supplies a tenth HBPO initialization.

\subsection{Native baseline selection}
\label{app:protocol-native}

Each method is tuned independently in DSIS and DSIS-RB using twelve
joint configurations from an eight-dimensional scrambled Sobol sequence
(seed 1413164337). Table~\ref{tab:native-search-space} gives the search
ranges. Learning rates use logarithmic sampling on \([10^{-5},10^{-3}]\)
unless specified otherwise; clipping and GAE intervals use linear
sampling, and discrete choices occupy equal intervals.

Each candidate trains for 500,000 real interactions. The eight non-GPS
methods use tuning seed 0; GPS uses seeds \(5,6,7\). Tuning and formal
training use separate random streams even when seed labels coincide.
Candidates are evaluated with sampled actions on 100 trajectories of
1,000 periods per risk level, using the cost measure defined below.
Within each method and tuning seed, candidates share evaluation tapes
and action-randomness streams. We select the lowest validation cost,
averaged over the three training seeds for GPS, and break ties by trial
identifier. Tables~\ref{tab:native-selected-dsis} and
\ref{tab:native-selected-rb} give the selected parameters, rounded to
six significant figures. These settings remain fixed for all formal
seeds and their paired native/HBPO comparisons.

\begin{table}[!htbp]
\centering\small
\caption{Native search spaces. The DSIS-RB collection multiplier applies only
to GRPO and TPO.}
\label{tab:native-search-space}
\begin{tabular}{lp{.75\linewidth}}
\toprule
Method & Searched dimensions \\
\midrule
PPO, PPO-81 & Learning rate; clip ratio $[.1,.3]$; GAE $\lambda\in[.90,.99]$;
 epochs $\{5,10\}$; rollout length $\{256,512,1024\}$;
 entropy coefficient $\{0,.001,.01\}$.\\
A2C & Learning rate; GAE $\lambda\in[.90,.99]$;
 rollout length $\{20,50,100\}$; entropy coefficient $\{0,.001,.01\}$.\\
Discrete SAC & Learning rate; minibatch $\{128,256,512\}$;
 target update rate $\{.005,.01,.02\}$; target entropy ratio $\{.1,.25,.5\}$.\\
ACE & Actor and critic learning rates; ETD $\lambda\in\{0,.5,.9\}$;
 behavior temperature $\{.5,1,2\}$.\\
TRPO & KL limit: log $[.003,.03]$; value learning rate;
 GAE $\lambda\in[.90,.99]$; rollout length $\{512,1024,2048\}$.\\
GRPO & Actor and value learning rates; clip ratio $[.1,.3]$;
 group size $\{4,8\}$; actor epochs $\{1,2,4\}$; KL coefficient $\{0,.04\}$;
 RB collection multiplier $\{4,8,16\}$.\\
TPO & Actor and value learning rates; temperature $\eta$: log $[.001,1]$;
 group size $\{4,8\}$; actor epochs $\{1,2,4\}$;
 RB collection multiplier $\{4,8,16\}$.\\
GPS & Actor and dual learning rates; guide learning rate: log $[10^{-4},10^{-2}]$;
 penalty: log $[.01,10]$; per-step KL limit: log $[.001,.1]$;
 inner iterations $\{2,4,8\}$; model horizon $\{64,256,1000\}$;
 model paths $\{8,16,32\}$.\\
\bottomrule
\end{tabular}
\end{table}

\begin{table}[!htbp]
\centering\small
\caption{Selected native configurations for DSIS.}
\label{tab:native-selected-dsis}
\begin{tabular}{llp{.68\linewidth}}
\toprule
Method & Trial & Selected parameters \\
\midrule
PPO & 03 & Learning rate $2.80429\times10^{-4}$; clip $.214343$;
 GAE $\lambda=.938579$; 5 epochs; rollout 512; entropy 0.\\
A2C & 04 & Learning rate $1.20204\times10^{-4}$; GAE $\lambda=.915092$;
 rollout 20; entropy $.001$.\\
Discrete SAC & 11 & Learning rate $1.65339\times10^{-4}$; minibatch 512;
 target update rate $.02$ (Polyak coefficient $.98$); entropy ratio $.1$.\\
ACE & 01 & Actor learning rate $2.09424\times10^{-5}$;
 critic/follow-on learning rate $6.92692\times10^{-4}$;
 ETD $\lambda=0$; behavior temperature $.5$.\\
PPO-81 & 04 & Learning rate $1.20204\times10^{-4}$; clip $.133537$;
 GAE $\lambda=.914021$; 5 epochs; rollout 512; entropy $.001$.\\
TRPO & 03 & KL limit $.0158867$; value learning rate $1.39131\times10^{-4}$;
 GAE $\lambda=.938579$; rollout 1024.\\
GRPO & 07 & Actor learning rate $7.73584\times10^{-4}$;
 value learning rate $3.35597\times10^{-4}$; clip $.263714$;
 group size 8; 4 actor epochs; KL coefficient 0.\\
TPO & 04 & Actor learning rate $1.20204\times10^{-4}$;
 value learning rate $2.16455\times10^{-5}$; $\eta=.00293339$;
 group size 4; 2 actor epochs.\\
GPS & 01 & Actor learning rate $2.09424\times10^{-5}$;
 guide learning rate $.00692692$; dual learning rate $1.76382\times10^{-5}$;
 penalty $.0387192$; KL limit $.00179181$;
 8 inner iterations; model horizon 64; 16 model paths.\\
\bottomrule
\end{tabular}
\end{table}

\begin{table}[!htbp]
\centering\small
\caption{Selected native configurations for DSIS-RB.}
\label{tab:native-selected-rb}
\begin{tabular}{llp{.68\linewidth}}
\toprule
Method & Trial & Selected parameters \\
\midrule
PPO & 08 & Learning rate $5.66943\times10^{-4}$; clip $.119578$;
 GAE $\lambda=.927511$; 10 epochs; rollout 256; entropy 0.\\
A2C & 11 & Learning rate $1.65339\times10^{-4}$; GAE $\lambda=.984611$;
 rollout 100; entropy 0.\\
Discrete SAC & 04 & Learning rate $1.20204\times10^{-4}$; minibatch 128;
 target update rate $.005$ (Polyak coefficient $.995$); entropy ratio $.25$.\\
ACE & 10 & Actor learning rate $7.52789\times10^{-5}$;
 critic/follow-on learning rate $5.70075\times10^{-5}$;
 ETD $\lambda=0$; behavior temperature 2.\\
PPO-81 & 03 & Learning rate $2.80429\times10^{-4}$; clip $.214343$;
 GAE $\lambda=.938579$; 5 epochs; rollout 512; entropy 0.\\
TRPO & 00 & KL limit $.0173037$; value learning rate $9.30087\times10^{-5}$;
 GAE $\lambda=.954778$; rollout 2048.\\
GRPO & 07 & Actor learning rate $7.73584\times10^{-4}$;
 value learning rate $3.35597\times10^{-4}$; clip $.263714$;
 group size 8; 4 actor epochs; KL coefficient 0; collection multiplier 4.\\
TPO & 00 & Actor learning rate $3.32686\times10^{-4}$;
 value learning rate $9.30087\times10^{-5}$; $\eta=.0669775$;
 group size 8; 4 actor epochs; collection multiplier 16.\\
GPS & 01 & Same selected numerical configuration as DSIS GPS.\\
\bottomrule
\end{tabular}
\end{table}

\subsection{HBPO selection}
\label{app:protocol-hpo}

In each environment, we evaluate the seven configurations in
Table~\ref{tab:hbpo-search} on all ten initializations (nine warm sources
and Cold), using tuning seeds \(5,6,7\). Candidates share the initial
learner state, first data batch, and calibration scale within each
initialization and seed.
Each candidate runs for 15 generations of 25,000 real interactions,
including the first batch, and is validated with sampled target actions
on 100 trajectories of 1,000 periods per risk level. The grid varies
value learning rate, emphatic learning rate, and the heuristic scale
multiplier \(\kappa\), with other settings fixed. The calibration of
\(\kappa\) is specified in Appendix~\ref{app:h-regularization-scale}.

\begin{table}[!htbp]
\centering\small
\caption{The complete HBPO candidate set. H5 is used in the final experiments.}
\label{tab:hbpo-search}
\begin{tabular}{lrrr}
\toprule
Candidate & Value learning rate & Emphatic learning rate & $\kappa$\\
\midrule
H0 & $5.66943\times10^{-4}$ & $1.56958\times10^{-5}$ & $1/4$\\
H1 & $1.79283\times10^{-4}$ & $1.56958\times10^{-5}$ & $1/4$\\
H2 & $1.79283\times10^{-3}$ & $1.56958\times10^{-5}$ & $1/4$\\
H3 & $5.66943\times10^{-4}$ & $4.96345\times10^{-6}$ & $1/4$\\
H4 & $5.66943\times10^{-4}$ & $4.96345\times10^{-5}$ & $1/4$\\
H5 & $5.66943\times10^{-4}$ & $1.56958\times10^{-5}$ & $1/32$\\
H6 & $5.66943\times10^{-4}$ & $1.56958\times10^{-5}$ & $2$\\
\bottomrule
\end{tabular}
\end{table}

For each tuning seed, split initializations into initial-cost groups
\(C_0\le1010\) and \(C_0>1010\). For each candidate, compute percentage
improvement \(100(1-C_{15}/C_0)\), average within each group, and then
average these group means across seeds with nonempty groups. Denote the
two scores by \(P\) and \(R\), respectively. Rank by each score,
breaking ties by the other score and then candidate ID. Select the
highest-\(P\) candidate in the intersection of the two top-three lists;
if the intersection is empty, select the highest-\(R\) candidate among
the \(P\) top three. If one group is empty for every seed, use the
other ranking. This rule selects H5 in both environments:
value learning rate \(5.66943\times10^{-4}\), emphatic learning rate
\(1.56958\times10^{-5}\), and \(\kappa=1/32\). One configuration is
shared across all initializations and formal seeds within each
environment; ablations use it without retuning.

\subsection{Evaluation protocol}
\label{app:protocol-evaluation}

\paragraph{Budgets and seeds.}
Each formal run uses 2M real interactions and one of five training seeds,
\(0,\ldots,4\). Warm comparisons branch from the same native 0.5M
checkpoint into native continuation and HBPO, each receiving another
1.5M interactions. Cold receives 2M without a native prefix. Tuning and
formal-evaluation data are separate, and formal results do not influence
hyperparameter or checkpoint selection. Simulation and evaluation are
accounted for separately from real training interactions.

\paragraph{Information access.}
All methods have the same public-information permissions and visible
observations. DSIS's stochastic law is public; DSIS-RB's true stochastic
kernels, episode risks, and future random tapes are hidden. In DSIS-RB,
HBPO and GPS fit the same model class to their own budgeted observations.
Modeling assumptions are given in Appendix~\ref{app:critic-prior-comparison}.

\paragraph{Test cost and reporting.}
We freeze the final policy and sample its actions; HBPO always reports
the target policy \(\pi\). Each policy is tested on 500 independent
trajectories of 1,000 periods, including all periods from the initial
state. DSIS reports mean undiscounted raw cost per period. DSIS-RB uses
500 trajectories at each \(\rho\in\{0,.1,.2,.3,.4,.5\}\) and reports
\(C_s^{\mathrm{RB}}=\sum_{k=0}^5\omega_k C_s(0.1k)\), where
\(\omega=(.1,.2,.2,.2,.2,.1)\) and \(C_s(\rho)\) is the mean cost for
training seed \(s\) at risk \(\rho\). This is normalized trapezoidal
risk-AUC. We report the mean, median, and sample standard deviation
across the five training seeds, using unrounded costs for comparisons.
Ablations are paired with their corresponding Full run by source,
seed, and evaluation bank, as detailed in Appendix~\ref{app:ablations}.

\paragraph{Cost-collapse thresholds.}
\label{app:cost_thre}
Mean demand is two units per period. Expedited replacement costs
\(2\times10=20\), and \(k\) periods of mean demand in backlog incur
\(2k\times495=990k\) per period. The reference cost
\(20+990k\) gives 1010 for one period and 3980 for four periods.
We classify costs as Non-collapse (\(C\le1010\)), Collapse
(\(1010<C\le3980\)), or Severe Collapse (\(C>3980\)). These are
economic severity thresholds, not estimates of actual delivery delay.
Endpoint recovery means a native 2M cost above 1010 paired with an
HBPO 2M cost at or below 1010.

\section{Extended Results}
\label{app:results}

\subsection{Detailed Main Results}
\label{app:results_main}

Tables~\ref{tab:dsis-seeds} and~\ref{tab:rb-seeds} expand
Table~\ref{tab:main-results} using the same 190 final-policy costs:
90 native endpoints, 90 warm HBPO endpoints, and 10 Cold endpoints.
Each policy is evaluated after a total budget of 2M environment interactions;
warm HBPO includes the 0.5M native prefix and reports the final target policy
$\pi$. H denotes warm HBPO, and Cold has no native prefix or paired native
endpoint. All summaries retain all five training seeds. Evaluation protocols
and cost grades follow Appendix~\ref{app:protocol-evaluation}.

\paragraph{Precision and table conventions.}
Seed costs retain their full effective decimal precision: five places for DSIS
and six for DSIS-RB, without k/M abbreviations. Only binary floating-point
representation tails are removed. Means and medians are exact at this
precision; sample standard deviations use the $n-1=4$ denominator and are
displayed to nine decimal places. Black, ochre, and red seed entries indicate
$c\le1010$, $1010<c\le3980$, and $c>3980$, respectively. Bold marks the lower
paired mean or median; underlining marks HBPO costs $\le1010$ whose
seed-matched native endpoints exceed 1010. Comparisons use the full-precision
values, not the abbreviated main-table entries.

\paragraph{DSIS.}
HBPO lowers mean cost in all nine warm-source comparisons, median cost in
eight (TRPO is the exception), and sample standard deviation in all nine.
It lowers cost in 40 of 45 seed-matched comparisons. All 12 native endpoints
above 1010 have lower-cost HBPO counterparts, and 10 cross to costs $\le1010$.
All 33 native endpoints at or below 1010 remain within this range under HBPO,
with 28 achieving lower costs. The two HBPO endpoints still above 1010 are
A2C seed 2 (6638.13434) and ACE seed 0 (1207.57231).

\paragraph{DSIS-RB.}
HBPO lowers mean cost in eight of nine warm-source comparisons, median cost in
seven (PPO and A2C are the exceptions), and sample standard deviation in all
nine. It lowers cost in 41 of 45 seed-matched comparisons. Among the 24 native
endpoints above 1010, 23 have lower-cost HBPO counterparts and 19 cross to
costs $\le1010$. All 21 native endpoints at or below 1010 remain within this
range under HBPO, with 18 achieving lower costs. Discrete SAC is the mean-cost
exception: 2479.0853802 increases to 2514.9923076 (1.4484\%), although its
median and standard deviation decrease. Its seed 0 worsens from 1183.646256
to 5359.057081, moving from Collapse to Severe Collapse. The five HBPO
endpoints still above 1010 are PPO seed 2, A2C seed 2, Discrete SAC seeds 0
and 1, and ACE seed 0.

\paragraph{Cold starts and scope of the comparisons.}
All five Cold endpoints are at or below 1010 in each environment.
DSIS has mean 45.103804 and range [42.67906, 47.51533]; DSIS-RB has mean
72.1150812 and range [69.675982, 74.916383]. Warm comparisons pair native and
HBPO endpoints by environment, source, training seed, and interaction budget;
they do not assume identical evaluation random banks across campaigns.
Crossing the threshold relative to a native 2M endpoint is distinct from
recovery relative to the 0.5M initialization. These descriptive results
characterize the evaluated seeds and do not establish statistical significance
or a task-independent stability guarantee.

\clearpage
\begin{table}[p]
\caption{Detailed final-policy costs on DSIS (mean cost per period; lower is better),
expanding Table~\ref{tab:main-results}. Panel (a) retains five decimal
places for every seed; panel (b) reports exact means and medians, with sample
standard deviations displayed to nine decimal places. Formatting follows the
conventions in Appendix~\ref{app:results_main}.}
\label{tab:dsis-seeds}
\centering
\begingroup
\small
\definecolor{appcollapse}{RGB}{165,115,0}
\definecolor{appsevere}{RGB}{175,34,39}
\setlength{\tabcolsep}{2.5pt}
\renewcommand{\arraystretch}{1.0}
\begin{tabular*}{\linewidth}{@{\extracolsep{\fill}}lrrrrr@{}}
\toprule
\multicolumn{6}{c}{\textbf{(a) Individual training seeds}} \\
\midrule
Policy & Seed 0 & Seed 1 & Seed 2 & Seed 3 & Seed 4 \\
\midrule
PPO & 21.00574 & 21.16667 & 56.89301 & \textcolor{appsevere}{13817.30298} & 21.33039 \\
H-PPO & 20.73032 & 21.25826 & 51.41226 & \underline{158.87953} & 20.75846 \\
\addlinespace[1pt]
A2C & 23.65804 & 28.68256 & \textcolor{appsevere}{23012.57348} & 30.43475 & 23.38418 \\
H-A2C & 32.28773 & 20.76730 & \textcolor{appsevere}{6638.13434} & 20.88466 & 20.74205 \\
\addlinespace[1pt]
Discrete SAC & 27.82865 & 36.65902 & 38.56467 & 27.45829 & 345.08867 \\
H-Discrete SAC & 26.56887 & 33.98975 & 33.06797 & 32.39846 & 55.35698 \\
\addlinespace[1pt]
ACE & \textcolor{appcollapse}{1679.44062} & 22.57928 & 22.34593 & \textcolor{appsevere}{7504.31927} & \textcolor{appcollapse}{1116.35635} \\
H-ACE & \textcolor{appcollapse}{1207.57231} & 319.84945 & 21.40372 & \underline{77.84574} & \underline{37.53063} \\
\addlinespace[1pt]
PPO-81 & 22.35240 & 22.70225 & 21.70040 & 22.96933 & 21.49762 \\
H-PPO-81 & 21.37053 & 22.01671 & 21.16084 & 20.89490 & 21.48099 \\
\addlinespace[1pt]
TRPO & \textcolor{appsevere}{5026.80023} & \textcolor{appsevere}{12492.90475} & 26.03542 & 29.66418 & 25.40624 \\
H-TRPO & \underline{102.58326} & \underline{219.28574} & 25.62822 & 30.75901 & 25.13818 \\
\addlinespace[1pt]
GRPO & 37.64513 & 32.93183 & 32.32662 & 50.67199 & 30.85771 \\
H-GRPO & 22.46585 & 26.81074 & 20.99287 & 23.23004 & 21.89809 \\
\addlinespace[1pt]
TPO & \textcolor{appsevere}{4207.30851} & \textcolor{appsevere}{4804.99275} & \textcolor{appsevere}{4560.12880} & \textcolor{appsevere}{4171.18351} & \textcolor{appsevere}{4399.49216} \\
H-TPO & \underline{45.11618} & \underline{44.84068} & \underline{45.91967} & \underline{44.60019} & \underline{45.65733} \\
\addlinespace[1pt]
GPS & 123.18311 & 120.45602 & 106.86388 & 117.79971 & 124.03338 \\
H-GPS & 39.81263 & 41.73971 & 41.15723 & 44.38921 & 35.46236 \\
\midrule
Cold & 44.56601 & 47.51533 & 45.09705 & 45.66157 & 42.67906 \\
\bottomrule
\end{tabular*}
\par\medskip
\begin{tabular*}{\linewidth}{@{\extracolsep{\fill}}lrrr@{}}
\toprule
\multicolumn{4}{c}{\textbf{(b) Across all five seeds}} \\
\midrule
Policy & Mean & Median & Std \\
\midrule
PPO & 2787.539758 & 21.33039 & 6165.844492000 \\
H-PPO & \textbf{54.607766} & \textbf{21.25826} & 59.767170003 \\
\addlinespace[1pt]
A2C & 4623.746602 & 28.68256 & 10279.667193796 \\
H-A2C & \textbf{1346.563216} & \textbf{20.88466} & 2958.082369644 \\
\addlinespace[1pt]
Discrete SAC & 95.119860 & 36.65902 & 139.827358107 \\
H-Discrete SAC & \textbf{36.276406} & \textbf{33.06797} & 11.055155831 \\
\addlinespace[1pt]
ACE & 2069.008290 & 1116.35635 & 3121.642702425 \\
H-ACE & \textbf{332.840370} & \textbf{77.84574} & 503.623773687 \\
\addlinespace[1pt]
PPO-81 & 22.244400 & 22.35240 & 0.632540611 \\
H-PPO-81 & \textbf{21.384794} & \textbf{21.37053} & 0.418009117 \\
\addlinespace[1pt]
TRPO & 3520.162164 & \textbf{29.66418} & 5463.192396034 \\
H-TRPO & \textbf{80.678882} & 30.75901 & 84.111535246 \\
\addlinespace[1pt]
GRPO & 36.886656 & 32.93183 & 8.114700906 \\
H-GRPO & \textbf{23.079518} & \textbf{22.46585} & 2.240041009 \\
\addlinespace[1pt]
TPO & 4428.621146 & 4399.49216 & 262.226888622 \\
H-TPO & \textbf{45.226810} & \textbf{45.11618} & 0.552130068 \\
\addlinespace[1pt]
GPS & 118.467220 & 120.45602 & 6.932895316 \\
H-GPS & \textbf{40.512228} & \textbf{41.15723} & 3.276596300 \\
\midrule
Cold & 45.103804 & 45.09705 & 1.753226456 \\
\bottomrule
\end{tabular*}
\endgroup
\end{table}
\clearpage

\begin{table}[p]
\caption{Detailed final-policy costs on DSIS-RB (normalized risk-AUC cost; lower is better),
expanding Table~\ref{tab:main-results}. Panel (a) retains six decimal
places for every seed; panel (b) reports exact means and medians, with sample
standard deviations displayed to nine decimal places. Formatting follows the
conventions in Appendix~\ref{app:results_main}.}
\label{tab:rb-seeds}
\centering
\begingroup
\small
\definecolor{appcollapse}{RGB}{165,115,0}
\definecolor{appsevere}{RGB}{175,34,39}
\setlength{\tabcolsep}{2.5pt}
\renewcommand{\arraystretch}{1.0}
\begin{tabular*}{\linewidth}{@{\extracolsep{\fill}}lrrrrr@{}}
\toprule
\multicolumn{6}{c}{\textbf{(a) Individual training seeds}} \\
\midrule
Policy & Seed 0 & Seed 1 & Seed 2 & Seed 3 & Seed 4 \\
\midrule
PPO & 70.061774 & 75.460918 & \textcolor{appsevere}{66970.754322} & 75.131193 & 74.882238 \\
H-PPO & 52.679426 & 81.929137 & \textcolor{appsevere}{5893.013522} & 55.296788 & 112.014382 \\
\addlinespace[1pt]
A2C & 74.484857 & 69.637995 & \textcolor{appsevere}{8036.258056} & 807.481221 & 69.608571 \\
H-A2C & 51.574127 & 51.333329 & \textcolor{appsevere}{7397.851829} & 125.837686 & 653.867411 \\
\addlinespace[1pt]
Discrete SAC & \textcolor{appcollapse}{1183.646256} & \textcolor{appsevere}{10969.145360} & 71.162997 & 69.717788 & 101.754500 \\
H-Discrete SAC & \textcolor{appsevere}{5359.057081} & \textcolor{appsevere}{7028.880845} & 61.773085 & 58.874363 & 66.376164 \\
\addlinespace[1pt]
ACE & \textcolor{appsevere}{9955.781036} & \textcolor{appsevere}{12184.555697} & \textcolor{appsevere}{22312.909190} & \textcolor{appsevere}{13801.992411} & \textcolor{appsevere}{15519.232290} \\
H-ACE & \textcolor{appsevere}{4013.167881} & \underline{109.714766} & \underline{797.749040} & \underline{121.330187} & \underline{85.117775} \\
\addlinespace[1pt]
PPO-81 & 191.330923 & 103.245154 & 96.113375 & 95.326541 & 95.308221 \\
H-PPO-81 & 124.935134 & 80.015061 & 86.630289 & 84.824489 & 84.067253 \\
\addlinespace[1pt]
TRPO & 427.649648 & 365.795013 & 317.922635 & \textcolor{appcollapse}{1996.454276} & 292.815271 \\
H-TRPO & 316.810653 & 289.684982 & 303.129556 & \underline{235.322284} & 166.282749 \\
\addlinespace[1pt]
GRPO & 537.104180 & \textcolor{appcollapse}{1243.313035} & \textcolor{appcollapse}{1102.386111} & \textcolor{appcollapse}{3234.895086} & \textcolor{appcollapse}{1215.766356} \\
H-GRPO & 80.778016 & \underline{92.341207} & \underline{107.239713} & \underline{93.158126} & \underline{222.144895} \\
\addlinespace[1pt]
TPO & \textcolor{appsevere}{11220.857853} & \textcolor{appcollapse}{1616.815238} & \textcolor{appcollapse}{2856.305500} & \textcolor{appcollapse}{1994.682157} & \textcolor{appcollapse}{3392.571200} \\
H-TPO & \underline{238.316432} & \underline{199.456230} & \underline{219.981811} & \underline{214.844334} & \underline{163.411692} \\
\addlinespace[1pt]
GPS & \textcolor{appcollapse}{1767.627848} & \textcolor{appsevere}{6215.901052} & \textcolor{appcollapse}{2410.230847} & \textcolor{appcollapse}{1922.649659} & \textcolor{appcollapse}{2013.114622} \\
H-GPS & \underline{233.994143} & \underline{200.358208} & \underline{188.219671} & \underline{239.709108} & \underline{215.895433} \\
\midrule
Cold & 69.675982 & 73.103120 & 74.916383 & 70.041407 & 72.838514 \\
\bottomrule
\end{tabular*}
\par\medskip
\begin{tabular*}{\linewidth}{@{\extracolsep{\fill}}lrrr@{}}
\toprule
\multicolumn{4}{c}{\textbf{(b) Across all five seeds}} \\
\midrule
Policy & Mean & Median & Std \\
\midrule
PPO & 13453.2580890 & \textbf{75.131193} & 29917.189972738 \\
H-PPO & \textbf{1238.9866510} & 81.929137 & 2601.790788363 \\
\addlinespace[1pt]
A2C & 1811.4941400 & \textbf{74.484857} & 3494.322436684 \\
H-A2C & \textbf{1656.0928764} & 125.837686 & 3219.614352805 \\
\addlinespace[1pt]
Discrete SAC & \textbf{2479.0853802} & 101.754500 & 4770.066181326 \\
H-Discrete SAC & 2514.9923076 & \textbf{66.376164} & 3409.927238739 \\
\addlinespace[1pt]
ACE & 14754.8941248 & 13801.992411 & 4697.178735828 \\
H-ACE & \textbf{1025.4159298} & \textbf{121.330187} & 1696.948305507 \\
\addlinespace[1pt]
PPO-81 & 116.2648428 & 96.113375 & 42.095435700 \\
H-PPO-81 & \textbf{92.0944452} & \textbf{84.824489} & 18.517339470 \\
\addlinespace[1pt]
TRPO & 680.1273686 & 365.795013 & 737.643085494 \\
H-TRPO & \textbf{262.2460448} & \textbf{289.684982} & 61.918285399 \\
\addlinespace[1pt]
GRPO & 1466.6929536 & 1215.766356 & 1029.108387150 \\
H-GRPO & \textbf{119.1323914} & \textbf{93.158126} & 58.344879455 \\
\addlinespace[1pt]
TPO & 4216.2463896 & 2856.305500 & 3977.588834515 \\
H-TPO & \textbf{207.2020998} & \textbf{214.844334} & 28.139939341 \\
\addlinespace[1pt]
GPS & 2865.9048056 & 2013.114622 & 1887.677624216 \\
H-GPS & \textbf{215.6353126} & \textbf{215.895433} & 21.804066539 \\
\midrule
Cold & 72.1150812 & 72.838514 & 2.213379602 \\
\bottomrule
\end{tabular*}
\endgroup
\end{table}
\clearpage

\providecommand{\transmissionfigdir}{figures/transmissionfigdir}
\IfFileExists{\transmissionfigdir/a2c-dsis.pdf}{}{%
  \IfFileExists{figure/transmission_appendix/a2c-dsis.pdf}{%
    \renewcommand{\transmissionfigdir}{figure/transmission_appendix}%
  }{}%
}

\subsection{Additional Transmission Analyses}
\label{app:transmission-results}

\begin{figure}[p]
    \centering
    \includegraphics[width=0.49\linewidth]{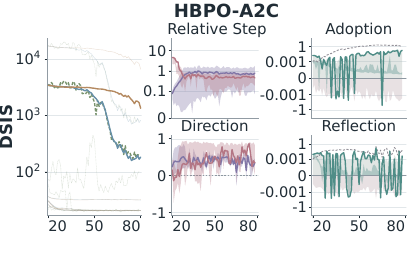}\hfill
    \includegraphics[width=0.49\linewidth]{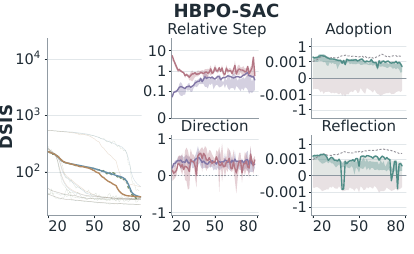}\par\medskip
    \includegraphics[width=0.49\linewidth]{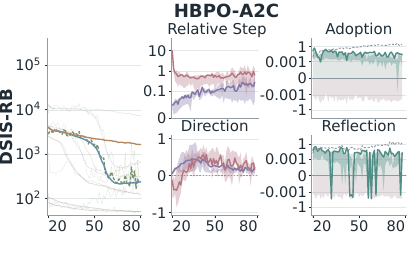}\hfill
    \includegraphics[width=0.49\linewidth]{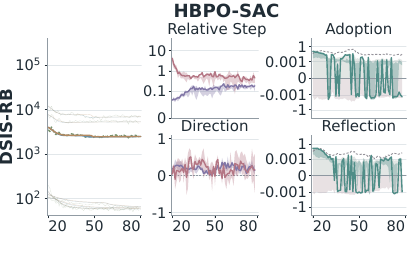}
    \caption{Additional transmission analysis for A2C (left) and Discrete SAC
    (right). Top: DSIS; bottom: DSIS-RB. Conventions follow
    Appendix~\ref{app:transmission-results}.}
    \label{fig:app-transmission-a2c-sac}
\end{figure}

\begin{figure}[p]
    \centering
    \includegraphics[width=0.49\linewidth]{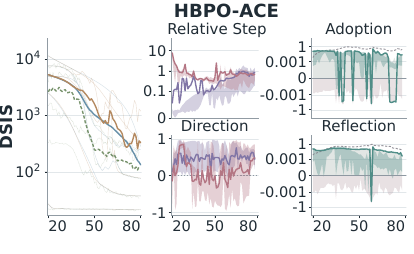}\hfill
    \includegraphics[width=0.49\linewidth]{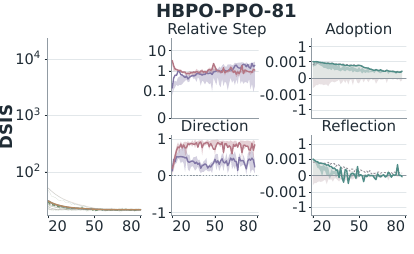}\par\medskip
    \includegraphics[width=0.49\linewidth]{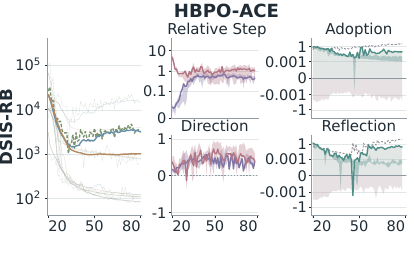}\hfill
    \includegraphics[width=0.49\linewidth]{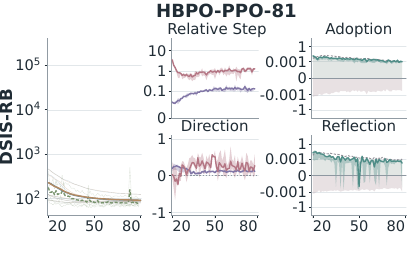}
    \caption{Additional transmission analysis for ACE (left) and PPO-81
    (right). Top: DSIS; bottom: DSIS-RB. Conventions follow
    Appendix~\ref{app:transmission-results}.}
    \label{fig:app-transmission-ace-ppo81}
\end{figure}

\begin{figure}[p]
    \centering
    \includegraphics[width=0.49\linewidth]{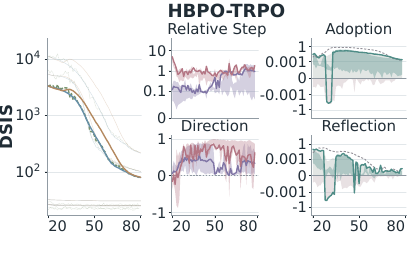}\hfill
    \includegraphics[width=0.49\linewidth]{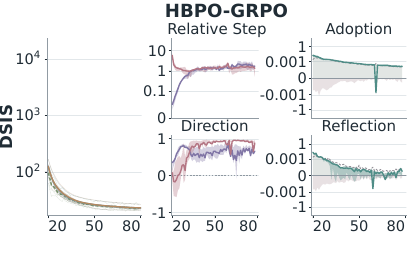}\par\medskip
    \includegraphics[width=0.49\linewidth]{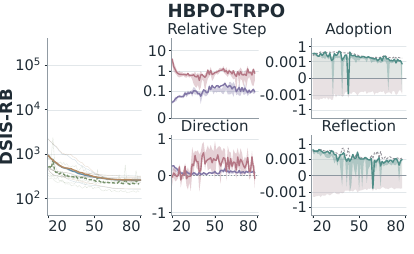}\hfill
    \includegraphics[width=0.49\linewidth]{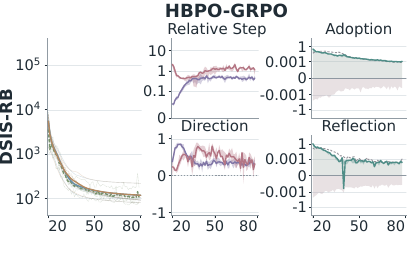}
    \caption{Additional transmission analysis for TRPO (left) and GRPO
    (right). Top: DSIS; bottom: DSIS-RB. Conventions follow
    Appendix~\ref{app:transmission-results}.}
    \label{fig:app-transmission-trpo-grpo}
\end{figure}

\begin{figure}[p]
    \centering
    \includegraphics[width=0.49\linewidth]{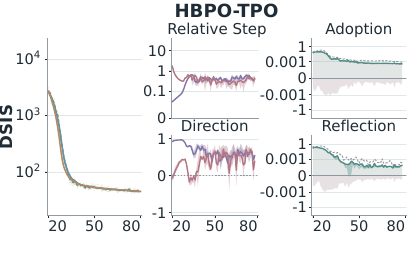}\hfill
    \includegraphics[width=0.49\linewidth]{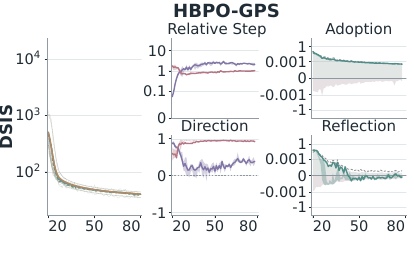}\par\medskip
    \includegraphics[width=0.49\linewidth]{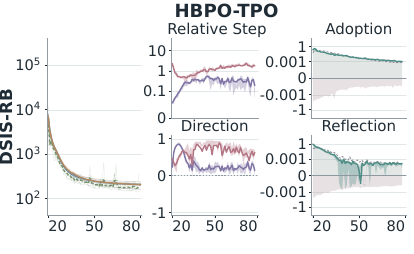}\hfill
    \includegraphics[width=0.49\linewidth]{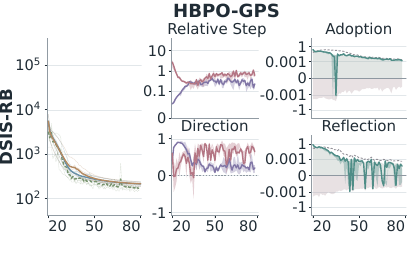}
    \caption{Additional transmission analysis for TPO (left) and GPS
    (right). Top: DSIS; bottom: DSIS-RB. Conventions follow
    Appendix~\ref{app:transmission-results}.}
    \label{fig:app-transmission-tpo-gps}
\end{figure}

Figures~\ref{fig:app-transmission-a2c-sac}--\ref{fig:app-transmission-tpo-gps}
complete the transmission analysis in Section~\ref{sec:transmission recog}.
They show the remaining 16 source--environment combinations: A2C, Discrete SAC,
ACE, PPO-81, TRPO, GRPO, TPO, and GPS on DSIS and DSIS-RB. Together with the
four Cold and PPO combinations in Figure~\ref{fig:p-transmission-analysis},
these cover all ten initializations in both environments. Each combination
contains five training seeds; HBPO-SAC denotes the Discrete SAC initialization.

\paragraph{Reading the panels.}
Within each combination, the left panel shows standard evaluation costs for
$\mu$ (blue) and $\pi$ (ochre), together with the green dashed $q_H$ reference.
The latter averages $H$-weighted immediate costs over the original nine-step
rollouts and is not a comparable long-horizon policy evaluation. The middle
panels show relative step $\varrho$ and direction cosine $c$, with purple for
$q_H\to\mu$ and rose for $\mu\to\pi$. The right panels show adoption and
reflection: solid teal curves are net responses $G_\pm$, gray dashed curves
are opportunities $B_\pm$, and pale green/pink backgrounds are mean favorable
and adverse contributions above/below zero. Positive $G_+$ and $G_-$ denote
favorable local responses in the adoption and reflection regions, respectively.

\paragraph{Aggregation and timing.}
Curves are arithmetic means across seeds; cost panels also retain individual
seed traces, while mechanism bands show the 25th--75th percentiles.
The signed contribution backgrounds are not uncertainty bands. Statistics are
computed before axis transformations, without smoothing or subsampling.
Costs use logarithmic axes; relative steps and selection responses use
inverse-hyperbolic-sine axes with linear scales $10^{-2}$ and $10^{-6}$,
respectively. Cost-axis limits are shared within each environment; mechanism
limits are shared across all combinations. These limits include the extrema
from all 20 combinations. Costs retain checkpoint time $t$;
mechanism panels pair the $q_H\to\mu$ update at $g$ with the $\mu\to\pi$
update and selection responses at $g+1$, following
$\pi_g\to q_{H,g}\to\mu_{g+1}\to\pi_{g+2}$. All panels here are warm starts
and display generation plus 20. These are descriptive diagnostics of the
recorded trajectories, rather than additional intervention experiments.

\clearpage

\clearpage

\clearpage

\subsection{Emphatic Weighting and Final Cost}
\label{app:emphatic-results}

\begin{figure}[!htb]
    \centering
    \includegraphics[width=\linewidth]{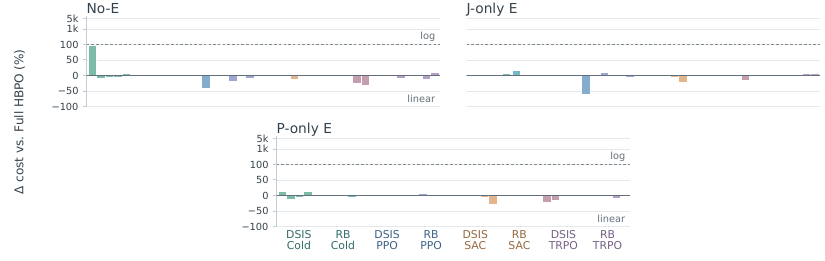}
    \caption{Emphatic-weight ablations at the final target-policy endpoint.
    No-E disables both emphatic branches; J-only E and P-only E retain
    only the indicated weighting branch. Each bar gives the percentage
    cost change relative to matched Full HBPO; positive values favor Full.
    Each environment--initialization block contains seeds 0--4 from left
    to right; darker bars mark ablation costs above 1010. All panels share
    the group order shown below and the H/P ablation scale, linear up to
    100\% and logarithmic above. All emphatic-ablation bars lie in the
    linear region.}
    \label{fig:emphatic-ablation}
\end{figure}

Figure~\ref{fig:emphatic-ablation} shows that emphatic weighting does not
provide a clear, consistent improvement in final target-policy cost after
the full $2$M-interaction training budget. We compare Full HBPO with No-E,
J-only E, and P-only E across Cold, PPO, Discrete SAC, and TRPO
initializations in both environments, retaining all five training seeds.
The interventions change the emphatic weighting branches while retaining
the J and P policy objectives (Appendix~\ref{app:emphatic-interventions}).
Against No-E, Full achieves a lower final cost in only $19$ of $40$ matched
runs and a lower mean in four of the eight environment--initialization
settings. The DSIS Cold mean favors Full, but this difference is driven
mainly by one seed; J-only E and P-only E likewise show mixed endpoint
effects. Thus, these results do not support a general long-run final-cost
advantage from emphatic weighting in the evaluated settings. This is a
descriptive comparison, not a formal significance or equivalence test,
and it does not determine whether weighting affects earlier learning.

\clearpage

\clearpage
\section{Ablation Details and Additional Diagnostics}
\label{app:ablations}

This appendix specifies the nine implemented interventions and the
additional diagnostics underlying the component and transmission analyses.
The interventions measure changes in the complete learning process at a
fixed setting; the observational diagnostics describe actual updates.

\subsection{Matched ablation panel and reporting}
\label{app:ablation-panel}

The ablation panel comprises PPO, Discrete SAC, TRPO, and Cold
initializations in each environment, with training seeds $0$--$4$.
Each intervention therefore has $40$ matched runs. We evaluate two
heuristic interventions, four coupling interventions, and three emphatic
interventions, giving $360$ intervention endpoints and $40$ shared Full
references. These are separate changes to Full, not a factorial
combination of multiple intervention families. Every variant uses the
selected H5 learning rates and $\kappa=1/32$ without retuning. Warm runs
use the same source checkpoint after $0.5$M interactions and another
$60$ generations of $25{,}000$ real transitions; Cold uses $80$
generations. Thus the final comparison is at $2$M real interactions.
The source policy, initialization procedure, scientific seed, and
initial real-data batch are matched to the corresponding Full run.
Subsequent policies, collected data, fitted models, and simulation
work can differ as a consequence of the intervention.

The final target policy is evaluated by sampled actions under the
formal protocol of Appendix~\ref{app:protocol}: $500$ episodes of
$1{,}000$ periods in DSIS, and the same number at each of the six risk
levels in DSIS-RB, followed by normalized trapezoidal risk aggregation.
These measurements do not feed back into training. The endpoint record
contains all five seeds in every cell. It combines qualified historical
DSIS runs for SAC, TRPO, and Cold with the subsequent DSIS PPO and
DSIS-RB runs. Each variant is compared with the Full reference evaluated
on its own matched bank. In particular, later Full reruns or reevaluations
used for the main comparison and generation diagnostics do not replace
these ablation denominators. This preserves within-comparison pairing
when the historical and subsequent evaluation banks differ.

For an environment--initialization--seed coordinate $i$, the plotted
change for variant $v$ is
\begin{equation}
  \Delta_{v,i}=100\left(\frac{c_{v,i}}{c_{\mathrm{Full},i}}-1\right).
  \label{eq:ablation-paired-cost-change}
\end{equation}
Positive values favor Full. An average paired percentage is the
arithmetic mean of these ratios; it is distinct from the percentage
change between two group means. Dark bars in the main ablation figures
indicate that the \emph{variant's} final cost exceeds $1010$.
Across-seed dispersion uses the sample standard deviation of the five
training-seed costs, rather than the standard error over evaluation
episodes. The ablation outcome evidence is an endpoint comparison;
it does not establish learning-curve smoothness, faster convergence,
or a training-curve area under the curve. Real interaction budgets are
matched, while model simulation and numerical work are accounted for
separately and need not be equal after a component is removed.

\subsection{Heuristic-guidance interventions}
\label{app:h-interventions}

\paragraph{No-H.}
This intervention deletes the entire finite heuristic objective,
so that $H(\theta)\equiv0$ and $g_H=0$, and removes the associated
H-occupancy construction and confidence calibration. The H state
channel in the behavior Fisher and endpoint constraints is replaced
by the complete real-data panel $D_g$ with unit weights. Consequently,
the two behavior channels use the same data measure and their averaged
Fisher reduces to the A-channel Fisher. The guidance-derived P interest
is set to one at the root and successor states. The A, J, and P
objectives, complete residual corrections, and remaining emphatic
learning are retained; in particular, setting P interest to one does
not disable the P emphatic learner. The initial utility-scale
calibration for $\tau$ can still require model queries. No-H therefore
tests the total contribution of the guidance module, including its
state weighting and geometry, rather than only setting a single
gradient vector to zero.

\paragraph{Random-H.}
At each queried public state, we construct a random target around this
variant's own frozen parent behavior policy. Let $q_{H,g}^{\mathrm{shadow}}$
be the simulation-derived controller at that same state under the same
parent policies and model. Draw a Gaussian action vector $z_g(s,\cdot)$
and choose $\beta_g(s)\ge0$ so that
\begin{align}
  q_{R,g}(a\mid s)
   &=\frac{\mu_g(a\mid s)\exp(\beta_g(s)z_g(s,a))}
           {\sum_b\mu_g(b\mid s)\exp(\beta_g(s)z_g(s,b))},\\
  d_{\mathrm{FR}}(\mu_g,q_{R,g})
   &=d_{\mathrm{FR}}(\mu_g,q_{H,g}^{\mathrm{shadow}}).
  \label{eq:ablation-random-h-radius}
\end{align}
For zero radius, the target is $\mu_g$. For a positive radius, directions
are rejected only if their limiting maximum-action vertex cannot reach
that radius; returns are not used to select a random direction.
A bracketed scalar solve finds the radius-matching $\beta_g(s)$.
The random field is addressed by environment, initialization, training
seed, generation, and public state, so repeated visits to the same
state within a generation use the same direction. Hidden risk does
not enter the address.

The random controller replaces the informative controller in both
occupancy rollouts and the paired calibration rollouts. The original
confidence-to-interest construction is retained. Because the shadow
controller still sets the local radius, Random-H retains model and
return information about the magnitude of guidance. The matching is
local to this arm's parent policy and visited states; it does not force
the Random-H and Full trajectories to visit the same states or to have
identical future exploration. This comparison tests the value of the
informative guidance direction within the resulting learning process.

\subsection{Behavior--target coupling interventions}
\label{app:coupling-interventions}

\paragraph{No-A.}
We remove the complete A channel from behavior learning, including
both the dense base term and sampled residual correction, and set its
finite objective to zero. The remaining H channel is passed through
the same fusion rule. The original behavior Fisher, both state-channel
endpoint constraints, target update, and critic construction are
retained. The removed signal is target-continuation advantage consumed
by the behavior actor.

\paragraph{No-P.}
We set the explicit P receiver gradient and finite P objective to zero
and recompute the target fusion with the unchanged J channel. The
original J and P state measures still define the target Fisher and
endpoint constraints. Behavior continues to generate the real training
data and can influence the target through data, critics, state weights,
and geometry. Thus No-P removes the explicit transfer objective; it
does not remove every effect of behavior on target learning.

\paragraph{KL-P.}
We replace P by forward-KL imitation of the parent behavior policy.
Up to a constant independent of the candidate target parameters,
the maximization objective is
\begin{equation}
  K_g(\theta)=\frac1{|D_g|}\sum_{s\in D_g}w_g^P(s)
        \sum_a\mu_g(a\mid s)\log\pi_\theta(a\mid s).
  \label{eq:ablation-kl-objective}
\end{equation}
This replacement has no advantage filter. Write $g_K=\nabla K_g$,
let $g_P$ be the complete original P gradient including the residual
correction, and retain the original target Fisher $F_\pi$. For positive
norms, the receiver is
\begin{equation}
  \widetilde g_K=
    \frac{\sqrt{g_P^\top F_\pi^{-1}g_P}}
         {\sqrt{g_K^\top F_\pi^{-1}g_K}}g_K.
  \label{eq:ablation-kl-norm-match}
\end{equation}
The same frozen scale multiplies the finite KL objective. If the P
reference norm is zero, the replacement is zero; a zero KL gradient
with a nonzero reference norm is recorded as an unmatchable direction
and also yields zero. The scaled receiver and J are fused using the
usual target geometry and endpoint constraints. This matches the
parent natural-gradient norm, not the finite policy displacement.

\paragraph{Near-on-policy.}
This arm maintains $\mu_g=\pi_g$ at committed generation boundaries.
It first computes the accepted AH behavior successor as a temporary
guide $\widetilde\mu_{g+1}$, then updates the target using J and
norm-matched KL imitation of this fresh guide. For this matching,
the reference P objective also uses
$\widetilde\mu_{g+1}-\pi_g$ as its teacher gap. The real-data panel,
critic evidence, and residual corrections remain those of the frozen
parent generation; no new real batch is collected between the two
actor updates. After the target update, an independent copy of
$\pi_{g+1}$ is committed as $\mu_{g+1}$. This intervention jointly
changes the receiver objective, transmission timing, and persistent
separation of the policies. It is therefore an alternative coupling
architecture rather than a pure delay-only intervention, and it is
not an implementation of GPS.

\subsection{Emphatic-weight interventions}
\label{app:emphatic-interventions}

The three emphatic interventions retain the H, A, J, and P objectives
but change which learned state-weight branches are consumed:
\begin{center}
\begin{tabular}{lcc}
\toprule
Variant & J emphatic branch & P emphatic branch\\
\midrule
Full & enabled & enabled\\
No-E & disabled & disabled\\
J-only E & enabled & disabled\\
P-only E & disabled & enabled\\
\bottomrule
\end{tabular}
\end{center}
A disabled branch uses a neutral mass network with zero output head
and unit root mass, forces both its raw mass and consumed weight to
one, and skips that branch's learner update. These unit weights are
applied before the residual-root sampling, heuristic construction,
critic evidence, actor objectives, and Fisher readers consume them.
Disabling P emphatic additionally sets its root and successor
interest to one. J-only E and P-only E refer to weighting branches;
both J and P policy objectives remain present in either case.
The endpoint effects include all subsequent changes in sampling and
optimization, so they do not independently identify a distribution
correction mechanism.

\subsection{Transmission diagnostics and their timing}
\label{app:transmission-diagnostics}

The transmission analysis follows the actual committed policy sequence,
\(\pi_g\to q_{H,g}\to\mu_{g+1}\to\pi_{g+2}\).
Both actors in an ordinary generation read frozen parents: the target update
that produces \(\pi_{g+1}\) uses \(\mu_g\), not the simultaneously produced
\(\mu_{g+1}\). We therefore pair the heuristic--behavior measurements from
update \(g\to g+1\) with the behavior--target measurements from update
\(g+1\to g+2\). The latter use the residual evidence, weights, and advantage
of generation \(g+1\). Parent and child checkpoint identities are checked
before joining the two updates; no link crosses runs or training seeds.

\paragraph{Evidence population.}
The complete diagnostic population contains both environments, all nine
warm sources and Cold, and five seeds. Warm runs have
60 updates and Cold runs have 80, giving 6,200 generation packets,
6,300 checkpoints, and 6,100 complete two-generation links. Each generation retains the original 1,152 heuristic fitting occurrences, the 128 residual-root draws, all 81 actions, and all 32 residual tapes per draw. Offline contraction of these arrays
introduces no further environment interactions or model rollouts.

\paragraph{Occurrence measures.}
The two links are measured on different panels. Write
\(\{s_h^H,\omega_h^H\}_{h=1}^{1152}\) for the original heuristic fitting
occurrences and their training weights. For residual draw \(k\), let
\(i_k\) be its sampled batch index and
\(\varpi_i=(1+w_i^J+w_i^P)/(3n)\) its proposal probability.
The retained importance factors are
\(\iota_k^J=w_{i_k}^J/(n\varpi_{i_k})\) and
\(\iota_k^P=w_{i_k}^P/(n\varpi_{i_k})\).
Define the diagnostic contractions
\begin{align}
 \langle x,y\rangle_H
 &=\frac1{1152}\sum_h\omega_h^H\sum_a x_h(a)y_h(a),\\
 \langle x,y\rangle_J
 &=\frac1{128}\sum_k\iota_k^J\sum_a x_k(a)y_k(a),
 &\langle z\rangle_P
 &=\frac1{128}\sum_k\iota_k^P\sum_a z_k(a).
 \label{eq:protocol-diagnostic-measures}
\end{align}
The weights are those supplied to the corresponding training packet.
In particular, the finite residual sample is not self-normalized by its
realized sum of importance weights. These diagnostics are occurrence-based
Euclidean contractions in action-probability space, not the parameter-space
Fisher norm used by the actor optimizer. Repeated states and repeated root
draws retain their original multiplicities.

\paragraph{Step size and direction.}
On the heuristic panel of generation \(g\), set
\(d_H=q_{H,g}-\mu_g\), \(u_H=\mu_{g+1}-\mu_g\).
On the residual panel of generation \(g+1\), set
\(d_J=\mu_{g+1}-\pi_{g+1}\), \(u_J=\pi_{g+2}-\pi_{g+1}\).
For either link \(L\in\{H,J\}\), compute
\begin{equation}
 \varrho_L=\frac{\sqrt{\langle u_L,u_L\rangle_L}}
                    {\sqrt{\langle d_L,d_L\rangle_L}},\qquad
 c_L=\frac{\langle d_L,u_L\rangle_L}
           {\sqrt{\langle d_L,d_L\rangle_L\langle u_L,u_L\rangle_L}}.
 \label{eq:protocol-step-direction}
\end{equation}
The original projection statistic is
\(\langle d_L,u_L\rangle_L/\langle d_L,d_L\rangle_L
=\varrho_Lc_L\), whenever both factors are defined. We show the factors
separately to distinguish update magnitude from directional agreement.
A zero guidance gap makes both displayed statistics undefined; a zero
update with a nonzero gap gives \(\varrho_L=0\) and an undefined cosine.
No small-gap threshold or zero-filling is introduced. Stable, scaled
contractions preserve very small nonzero gaps. Larger relative steps and
positive cosines describe geometric following, not necessarily better
policy value.

\paragraph{Adoption and reflection.}
At each residual draw of generation \(g+1\), reconstruct the full frozen
sampled advantage from the dense base and every saved residual tape:
\begin{equation}
 \widehat a_k(a)=a_k^0(a)
       +\frac1{32}\sum_{b=1}^{32}\Delta a_{kb}(a).
 \label{eq:protocol-diagnostic-advantage}
\end{equation}
Here \(\Delta a_{kb}\) is the stored action-centered residual sum over a geometrically stopped
continuation; discounting is represented by that stopping law.
No tape selection, refitting, clipping,
or additional centering is applied during postprocessing. With
\(d=d_J\), \(u=u_J\), define actionwise masks
\(S_+=\mathbf1\{d\widehat a>0\}\) and
\(S_-=\mathbf1\{d\widehat a<0\}\). Opportunities and responses are
\begin{align}
 B_+&=\langle[d\widehat a]_+\rangle_P,
 & B_-&=\langle[-d\widehat a]_+\rangle_P,\\
 G_+&=\langle u\widehat a S_+\rangle_P,
 & G_-&=\langle u\widehat a S_-\rangle_P.
 \label{eq:protocol-adoption-reflection}
\end{align}
Positive \(G_+\) denotes favorable target movement on advantageous
behavioral deviations; positive \(G_-\) denotes favorable correction of
adverse deviations. Negative responses are retained. In each region,
the plotted favorable and adverse components are
\begin{equation}
 F_\pm=\langle[u\widehat a S_\pm]_+\rangle_P,\qquad
 L_\pm=\langle[-u\widehat a S_\pm]_+\rangle_P,\qquad
 G_\pm=F_\pm-L_\pm.
 \label{eq:protocol-response-components}
\end{equation}
The shaded areas show \(F_\pm\) above zero and \(-L_\pm\) below zero.
The neutral region \(d\widehat a=0\) is retained in the audit, so its
response together with \(G_++G_-\) recovers the total sampled target
advantage change. When \(B_\pm=0\), that opportunity is inactive; no
response-to-opportunity ratio is imputed. These are signed sums under a
frozen sampled advantage, without division by \(\tau\). They are neither
raw-cost improvements nor estimates of a separately identified causal
mediation effect.

\paragraph{Costs and presentation.}
The \(\mu\) and \(\pi\) curves use the standard stochastic 500-tape,
1,000-period evaluator, at every saved checkpoint. In contrast,
\begin{equation}
 C_H(g)=\frac1{1152}\sum_h\omega_h^H c_h
 \label{eq:protocol-h-cost}
\end{equation}
is the realized immediate raw cost on the original controller occupancy
panel: 16 anchors, eight paths per anchor, and nine stages per path.
For DSIS-RB these paths use the generation-frozen fitted model; for DSIS
they use the public model. Thus \(C_H\) is a short-rollout reference from
training, not a standard-reset, long-run evaluation of a third policy.
It uses neither a new rollout nor an additional bootstrap value, and has
no fabricated terminal-generation point.

Costs are indexed by their own checkpoint generations. Link geometry and
responses are indexed by the origin \(g\) of the complete delayed chain;
the last heuristic update has no subsequent target update and is omitted
from these linked curves. Warm axes add 20 generation equivalents for the
500,000-interaction native prefix; Cold axes begin at zero. All statistics
are computed per seed before aggregation. The main figure shows arithmetic
means over all five seeds, pointwise 25th--75th percentile bands for the
link and response curves (linear interpolation), and individual seed cost
curves. These bands describe seed dispersion, not confidence intervals.
Undefined seed-level values leave a gap in the five-seed mean; they do not
reduce its denominator. There is no smoothing or temporal subsampling.
Logarithmic cost axes and inverse-hyperbolic-sine step/response axes alter
only display coordinates, not the underlying means or signed components.

\end{document}